\documentclass{article}
\usepackage[a4paper, top=1.1in, bottom=1.1in, left=1in, right=1in]{geometry}
\usepackage[utf8]{inputenc}
\usepackage[english]{babel}
\usepackage[T1]{fontenc}
\usepackage{rotating}
\usepackage{amsmath}
\usepackage{booktabs}
\usepackage{longtable}
\usepackage{subcaption}
\usepackage{multirow}
\usepackage{float}
\usepackage{makecell}
\usepackage{tikz}
\usepackage{lipsum}
\usepackage[semicolon, numbers,sort&compress]{natbib}
\usepackage{adjustbox}
\usepackage{amsfonts}
\usepackage{color}
\usepackage{graphicx}
\usepackage{hyperref}

\usepackage{setspace}
\usepackage[linesnumbered,ruled,vlined]{algorithm2e}
\usepackage{mathrsfs}
\usepackage{xr-hyper}
\usepackage{subcaption}

\usepackage{diagbox}
\usepackage{tabularx}

\def\Csc{\mathcal{C}}
\def\Isc{\mathcal{I}}
\def\PPMI{\mathbb{PPMI}}
\def\Rscr{\mathscr{R}}
\def\Esc{\mathcal{E}}
\def\Ssc{\mathcal{S}}
\def\Rsc{\mathcal{R}}
\def\Lscr{\mathscr{L}}

\def\suphalf{^{\scriptscriptstyle \half}}
\def\half{\frac{1}{2}}

\def\V{\mathbf{V}}
\def\X{\mathbf{X}}
\def\Y{\mathbf{Y}}
\def\Z{\mathbf{Z}}
\def\W{\mathbf{W}}
\def\U{\mathbf{U}}

\def\trans{^{\scriptscriptstyle \sf T}}

\definecolor{purple}{RGB}{200,000,200}
\definecolor{blue}{RGB}{050,000,200}
\definecolor{green}{RGB}{000,150,100}

\SetCommentSty{mycommfont}
\SetKwInput{KwInput}{Input}                
\SetKwInput{KwOutput}{Output}              

\begin{document}
\title{Representation learning to advance multi-institutional studies with electronic health record data from US and France}
\date{}

\author{Doudou Zhou$^{1,2}$, Han Tong$^{3}$, Linshanshan Wang$^{2}$, Suqi Liu$^{4}$, Xin Xiong$^2$,\\
Ziming Gan$^5$, Romain Griffier$^{6,7}$,  Boris P. Hejblum$^{6,8}$, Yun-Chung Liu$^9$, \\
Chuan Hong$^9$, Clara-Lea Bonzel$^{2,4}$, Tianrun Cai$^{10,11}$, 
Kevin Pan$^{12}$, Yuk-Lam Ho$^{10}$, \\
Lauren Costa$^{10}$, Vidul A. Panickan$^{4,10}$, J. Michael Gaziano$^{4,10,11}$,  Kenneth D. Mandl$^{13}$, \\
Vianney Jouhet$^{6,7}$, Rodolphe Thiebaut$^{6,7,8}$, Zongqi Xia$^{14}$, Kelly Cho$^{4,10,11}$,
\\ Katherine Liao$^{4, 10, 11*}$,  Tianxi Cai$^{2,4,10*}$ \\
\textsuperscript{1} Department of Statistics and Data Science, National University of Singapore, SG \\
\textsuperscript{2} Harvard T.H. Chan School of Public Health, MA, USA \\ 
\textsuperscript{3} Department of Statistics, Columbia University, NY, USA \\
\textsuperscript{4} Harvard Medical School, MA, USA \\
\textsuperscript{5} Department of Statistics, University of Chicago, Chicago, IL, USA  \\
\textsuperscript{6} Univ. Bordeaux, INSERM, Bordeaux Population Health Research Center, Bordeaux, France \\
\textsuperscript{7} CHU de Bordeaux, Service d'Information Médicale,  Bordeaux, France \\
\textsuperscript{8} Inria SISTM Team, Talence, France \\
\textsuperscript{9} 
Duke University, Durham, NC, USA \\
\textsuperscript{10} VA Boston Healthcare System, Boston, MA, USA\\
\textsuperscript{11} Brigham and Women's Hospital, Boston, MA, USA \\ 
\textsuperscript{12} 
Brown University,  Providence, RI, USA \\
\textsuperscript{13} Computational Health Informatics Program, Boston Children's Hospital, Boston, MA, USA\\
\textsuperscript{14} Department of Neurology, University of Pittsburgh, Pittsburgh, PA, USA
}

\maketitle

\noindent\textbf{These authors contributed equally:} Doudou Zhou, Han Tong, Linshanshan Wang, Suqi Liu.\\
\noindent\textbf{These authors jointly supervised this work:} Katherine Liao, Tianxi Cai.\\
\noindent\textbf{Correspondence:} \href{mailto:kliao@bwh.harvard.edu}{kliao@bwh.harvard.edu}; \href{mailto:tcai@hsph.harvard.edu}{tcai@hsph.harvard.edu}.

\begin{abstract}
The widespread adoption of electronic health records has created new opportunities for translational clinical research, yet this promise remains constrained by fragmented data across privacy-siloed institutions and substantial heterogeneity in local coding practices. While privacy-preserving collaborative learning allows institutions to work together without sharing patient-level data, it does not address inconsistencies in how clinical concepts are represented across sites. We introduce a graph-based framework that addresses this gap by treating data harmonization as a scalable representation learning problem. Rather than relying on fixed standards or manual mappings, the framework integrates institution-specific summary statistics from health records, curated biomedical knowledge graphs, and semantic information derived from large language models to learn a shared semantic space. This joint learning approach aligns diverse, site-specific vocabularies while preserving patient privacy. Evaluated across seven institutions and two languages, the framework provides a robust, data-centric foundation for training and deploying clinical models across heterogeneous healthcare systems.
\end{abstract}

 
\section*{Introduction}

Electronic health record (EHR) data play a central role in contemporary clinical and translational research. By capturing longitudinal information on diagnoses, medications, laboratory results, procedures, and detailed clinical narratives for millions of patients, EHR data support a wide range of studies, including cohort construction, disease phenotyping, risk prediction, real world evidence generation, and subgroup discovery \citep{liao2015development,Wangstrat,doshi2014comorbidity,sheu2023efficient}. Additionally, these data have facilitated the creation of clinical decision support tools 
\citep{federico2015gnaeus}, epidemiological surveillance \citep{ferte2022benefit}, and drug discovery \cite{wen2023multimodal}. A promise of EHR-based research is the potential for multicenter studies, which can include broader patient populations, improve generalizablility, and uncover heterogeneity in associations across subgroups \citep{cai2022consensus,pmlr-v174-hur22a,molaei2024federated,thakur2024knowledge}.  However, realizing this potential at scale remains challenging due to substantial heterogeneity across EHR and healthcare systems, where clinically important concepts can be coded differently, highlighting the need for effective data harmonization methods.

Standardized coding systems such as the International Classification of Diseases (ICD) \citep{world1988international} and the Logical Observation Identifiers Names and Codes (LOINC) for laboratory tests \citep{mcdonald2003loinc} provide an essential foundation for constructing common datasets across healthcare systems. However, the extent of adoption of these ontologies varies substantially across institutions: while ICD is used globally, LOINC is not universally implemented. As a result, a laboratory test included in a study may be represented by a standard LOINC code at one institution, but by one or several institution-specific identifiers, referred to here as local codes, in another healthcare system. Depending on the healthcare system, the number of such local codes can range from hundreds to tens of thousands. 

As artificial intelligence (AI) becomes increasingly integrated into healthcare research, recently developed algorithms often rely on high-dimensional EHR features. This growing complexity amplifies the need for scalable approaches to harmonize EHR data across healthcare systems, so that AI algorithms developed in one institution can be reliably deployed in others despite differences in local coding practices. Manual mapping is infeasible at scale, and ontology-based alignment alone cannot resolve the proliferation of local, legacy, or language-specific codes that differ across systems. These challenges underscore the need for automated, data-driven, and adaptive methods to achieve consistent and interoperable data integration at scale.

Another major challenge in multi-institutional EHR research is the need to preserve patient privacy. Traditional collaborative frameworks often require institutions to share patient-level data with a centralized site for model training, an approach that has become increasingly impractical due to regulatory, resource, and logistical constraints \citep{chen2016privacy}. Federated learning (FL) methods \citep{Federated_learning, molaei2024federated} address this limitation by enabling model co-training without transferring individual-level data. In FL, institutions perform local computation and share intermediate parameter updates or summary statistics with a central coordinator, without exchanging individual-level data. However, standard FL protocols typically assume that participating institutions operate in comparable feature spaces, such as shared laboratory or diagnostic codes mapped to a common ontology.

In practice, institutions differ widely in vocabulary coverage, coding conventions, and data scale, and many rely heavily on local codes that are not mapped to standardized systems. These discrepancies make cross-site feature alignment a critical challenge that traditional FL methods do not address. While recent approaches \citep[e.g.,][]{thakur2024knowledge} introduce knowledge abstraction or filtering to mitigate heterogeneity, they remain task-specific and communication-intensive. As a result, although FL provides a scalable and privacy-preserving framework, its success in biomedical research hinges on effective feature harmonization across institutions, a problem that must be addressed prior to, or alongside, federated model training.

Traditional efforts to align heterogeneous medical code systems have relied on rule-based or ontology-driven mapping, which are difficult to scale and often fail to capture the contextual relationships that define how codes are used in practice. Semantic embedding approaches emerged as a more data-driven and scalable alternative \citep{mikolov2013distributed,pennington2014glove,pmlr-v174-hur22a}, where each medical code is represented as a numeric vector encoding semantic similarity and relational structure. Among these, knowledge graph (KG) embeddings \citep{wang2014knowledge,balavzevic2019tucker,yuan2022coder,Yucong_disease} leverage ontological hierarchies, such as those in the Unified Medical Language System (UMLS) \citep{UMLS}, to capture structured biomedical relationships. Models like SAPBERT \citep{liu-2021-sapbert} and CODER \citep{yuan2022coder} use contrastive learning to enhance alignment among UMLS concepts and achieve good performance on entity normalization tasks \citep{maldonado2019adversarial,michalopoulos-etal-2021-umlsbert}. In parallel, graph neural network (GNN)-based methods \citep{piya2024healthgat} propagate information along UMLS or other ontology edges, leveraging the explicit relational structure of KGs to enrich embeddings with graph-level context. Together, these two approaches, text-enhanced fine-tuning and graph-structure propagation, have advanced semantic modeling of clinical concepts. However, they remain bounded by the coverage and completeness of existing ontologies and do not fully capture the real-world coding variation and usage patterns.

To address these limitations, a complementary line of work has focused on learning EHR-derived embeddings directly from co-occurrence relationships among medical codes in patient records \citep{choi2016multi,kartchner2017code2vec,hong2021clinical,MIKGI,gan2023arch}. These methods ground representations in observed clinical practice, allowing them to reflect institution-specific coding behaviors and latent associations that are not encoded in curated ontologies. However, embeddings trained within a single institution can generalize poorly to others due to differences in vocabularies, coding conventions, and data distributions \citep{joint_learning,MIKGI}. Thus, ontology-based and EHR-derived methods address complementary aspects of semantic modeling---global structure versus local semantics---but neither alone is sufficient to achieve robust interoperability across health systems.

Building on these foundations, recent studies have explored contrastive learning and large language models (LLMs) to integrate information from clinical text, biomedical ontological, and EHR data \citep{ying2024cortex,gao2025leveraging,cai2024contrastive}. These methods highlight the potential of enriching embeddings with linguistic and multimodal context beyond what any single data source can provide. However, most existing methods are developed within a single institution or rely on fixed ontologies, and thus lack effective mechanisms for integrating EHR information across multiple healthcare systems in the presence of heterogeneous coding practices and patient populations. While some approaches combine two modalities, such as ontology knowledge with language-based representations, they typically do not jointly leverage all available information. As a result, current models fall short of producing embeddings that simultaneously capture standardized biomedical knowledge, linguistic semantics, and real-world clinical usage in a manner that is robust and transferable across diverse healthcare systems, particularly when ontologies are incomplete or inconsistently adopted.

To address these limitations, we develop GAME (Graph Alignment for Multi-institutional EHR Data), a scalable framework that jointly learns representations for the union of EHR codes across multiple institutions. GAME integrates biomedical ontologies, pretrained language models (PLMs), and EHR co-occurrence patterns within a privacy-preserving federated architecture to generate unified embeddings for heterogeneous vocabularies. By applying contrastive learning with hard negatives curated by large language models (LLMs), the framework refines semantic distinctions and improves alignment accuracy across institutions. 

\begin{figure}[t]
    \centering
    \includegraphics[width=0.99\textwidth]{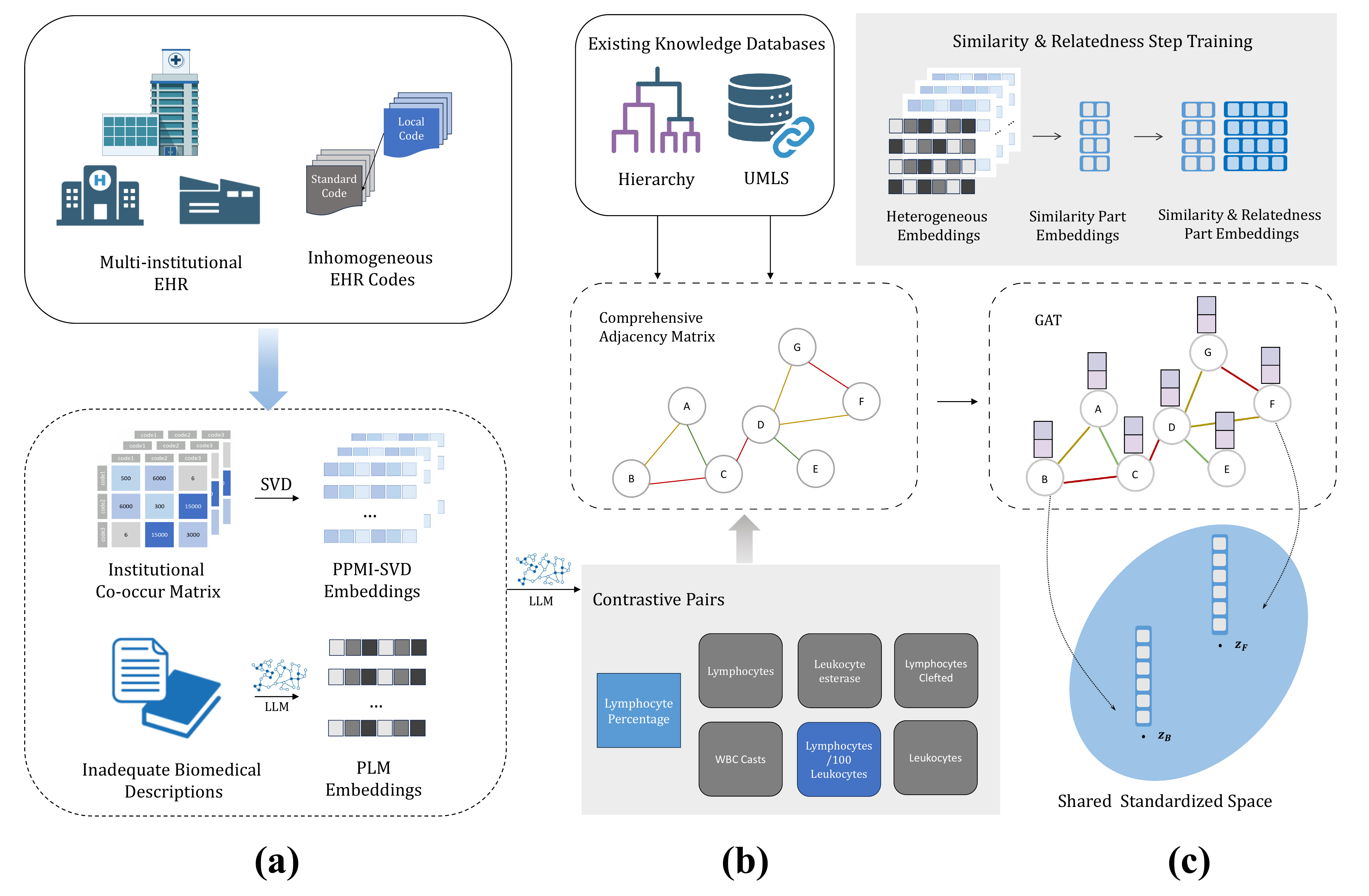} 
    \caption{Overview of the GAME (Graph Alignment for Multi-institutional EHR Data) algorithm. (a) Multi-institutional electronic health record (EHR) datasets and preprocessing to obtain initial embeddings. (b) Integration of existing biomedical knowledge graphs (KGs) and GPT-4-guided semantic relationships to curate contrastive pairs and construct graph edges. (c) Two-stage training of a graph attention network (GAT) to align heterogeneous EHR codes into a shared embedding space, yielding unified representations across institutions. Created using BioRender.com.} 
    \label{fig:outline}
\end{figure}

Unlike traditional FL approaches that focus on aggregating model parameters, GAME targets a fundamental source of EHR heterogeneity---variation in coding systems and feature representations---by applying FL directly at the representation level. Using only summary-level EHR data, GAME aligns code embeddings across sites before downstream modeling, eliminating the need for repeated inter-institutional communication or retraining. This design enables scalable, secure, and semantically consistent harmonization, allowing AI models trained in one healthcare system to be reliably deployed in others despite differences in local coding practices. To highlight the utility of GAME embeddings, we applied them to federated patient stratification for two clinical conditions using EHR data from six institutions in the United States and one in France. The overall framework is illustrated in Fig.~\ref{fig:outline}.

\section*{Results}

\subsection*{Validation of the GAME algorithm}

We performed a wide range of validation studies to evaluate the quality and clinical utility of GAME embeddings and compare them to benchmark embedding models. The validation tasks included: (1) detecting similar and related clinical relationships; (2) cross-institutional local code mapping,  which evaluates the quality of the harmonized embeddings; (3) feature selection;  (4) federated patient risk profiling using trained embeddings for (a) suicide-related behaviors and (b) Alzheimer's disease (AD) progression across multiple institutions, as representative downstream tasks.

For benchmark models, we included the original PPMI-SVD (Positive Pointwise Mutual Information -Singular Value Decomposition) embeddings \citep{levy2014neural,mikolov2013distributed}
 from individual institutions, PubMedBERT (PBERT) \citep{gu2021pubbert}, BioBERT (BBERT) \citep{lee2020biobert},  SAPBERT (SBERT), CODER, BAAI General Embedding (BGE; 768 dimensions) \cite{chen2024bge}, and OpenAI text-embedding-3-small (OpenAI; $1536$ dimensions; \url{https://platform.openai.com/docs/guides/embeddings/embedding-models}). Unless specified otherwise, all models were used off the shelf without fine-tuning, consistent with their original implementation protocols. In addition, we conducted ablation studies to assess the contributions of edge construction, the two-step training framework, and hard negatives. Details of the GAME algorithm and validation analyses are provided in the Methods.

\subsection*{Detecting similarity and relatedness between clinical concepts}

\begin{figure}[t]
\includegraphics[width=0.95\linewidth]{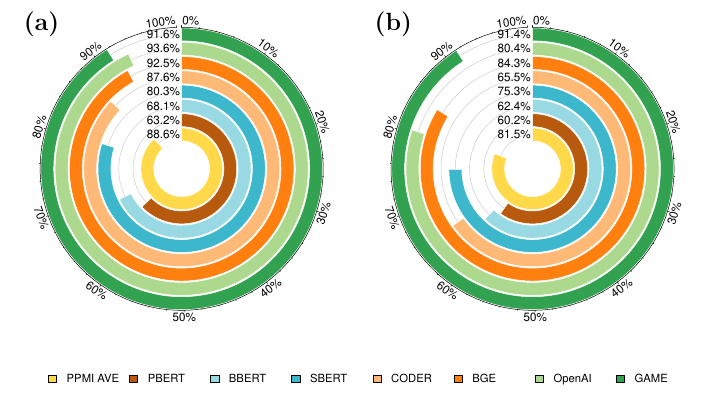}
\caption{Performance comparison of embeddings for similarity and relatedness detection. (a) Area under the receiver operating characteristic curve (AUC) for detecting similarity relationships and (b) AUC for detecting relatedness relationships, computed using embeddings derived from different methods or pretrained language models (PLMs). Each colored ring represents the mean AUC achieved by a specific embedding type. PPMI AVE denotes the average AUC obtained from institutional positive pointwise mutual information (PPMI) embeddings. All values are expressed as percentages. Source Data are provided in the Source Data file.}
\label{P2_AUC}
\end{figure}

We evaluated the GAME embeddings based on how well they detect similar and related clinical concepts through proximity in the embedding space. Fig.~\ref{P2_AUC} presents the area under the receiver operating characteristic curves (AUCs) for detecting similar and related clinical concept pairs against randomly sampled pairs. GAME achieved AUCs of $0.916$ for similarity relationships and $0.914$ for relatedness relationships. A detailed description of the data, along with confidence intervals and bootstrap $p$-values, is provided in Supplementary Table~\ref{R2}. The number of validation pairs is detailed in Supplementary Table~\ref{num_table}, and AUCs for specific relationship types are reported in Supplementary Table~\ref{R_sim}.

BGE (768 dimensions) and OpenAI (1536 dimensions) performed well for similarity detection, while GAME (256 dimensions) achieved comparable performance with lower-dimensional representations. In this validation, rather than directly using individual pairs, we used a more stringent approach by splitting hierarchical similarity pairs by branches; details are provided in Supplementary Section~\ref{sec:data_sup}. GAME achieved the highest relatedness AUC among the evaluated methods. To provide a strong baseline, we computed AUCs using institutional PPMI-SVD embeddings by evaluating pairs within each institution and then averaging AUCs across institutions. The PPMI-SVD embeddings largely preserved within-institution relationships but failed to detect relationships across institutions, whereas GAME performed well for both within- and cross-institution relationships.

Institution-specific analyses are presented in Supplementary Section~\ref{result_AUC_insti}. For relatedness AUC, GAME achieved the best performance across institutions (Supplementary Table~\ref{tab:rela_inst_AUC}). For similarity AUC, GAME trailed OpenAI and BGE by an average of 4.2\%, which is consistent with the use of 256-dimensional embeddings in GAME versus 1536-dimensional embeddings in OpenAI. When replacing the input pretrained embeddings from SAPBERT with BGE or CODER (Supplementary Section~\ref{sec:S2}), the gap narrowed, with GAME trailing OpenAI by 2.7\% and 2.0\%, respectively. These trends are consistent with the overall evaluation (Supplementary Table~\ref{R2}) and with institution-specific results (Supplementary Table~\ref{tab:simi_inst_AUC}).

\subsection*{Mapping codes between EHR systems}

\begin{table}[H]
\centering
\setlength{\tabcolsep}{4pt}
\begin{tabular}{c|lllllll}
\toprule
Measure (\%) & PBERT & BBERT & SBERT & CODER & OpenAI & BGE & GAME \\ \hline
TOP1
& $11.2^{(0.7)}_{***}$ 
& $20.8^{(0.9)}_{***}$ 
& $59.7^{(1.1)}_{***}$ 
& $55.6^{(1.1)}_{***}$ 
& $62.2^{(1.1)}_{***}$ 
& $61.9^{(1.1)}_{***}$ 
& $\mathbf{74.2^{(1.0)}}$ \\

TOP5  
& $18.3^{(0.9)}_{***}$ 
& $32.1^{(1.0)}_{***}$ 
& $79.0^{(0.9)}_{***}$ 
& $75.9^{(0.9)}_{***}$ 
& $86.4^{(0.8)}_{**}$ 
& $79.0^{(0.9)}_{***}$ 
& $\mathbf{88.0^{(0.7)}}$ \\

TOP10
& $21.9^{(0.9)}_{***}$ 
& $36.7^{(1.1)}_{***}$ 
& $82.9^{(0.8)}_{***}$ 
& $85.0^{(0.8)}_{***}$ 
& $89.6^{(0.7)}$ 
& $84.3^{(0.8)}_{***}$ 
& $\mathbf{90.7^{(0.6)}}$ \\
TOP20
& $27.0^{(1.0)}_{***}$ 
& $42.5^{(1.1)}_{***}$ 
& $85.7^{(0.8)}_{***}$ 
& $89.7^{(0.7)}_{***}$ 
& $92.6^{(0.6)}$ 
& $89.7^{(0.7)}_{***}$ 
& $\mathbf{92.7^{(0.6)}}$ \\
\bottomrule
\end{tabular}
\begin{flushleft}
\footnotesize
\textbf{Bold} indicates the best performance in each row. Values are reported as mean$^{(\mathrm{standard\ deviation})}$, with standard deviation measured in percentage points. Statistical significance relative to GAME was assessed using a two-sided McNemar's test ($\mathrm{df}=1$) and is indicated by superscripts: ** $p<0.005$, *** $p<0.0005$ (no adjustment for multiple comparisons).
\end{flushleft}
\caption{Accuracy (in \%) of mapping VA local laboratory codes to LOINC/LP codes using different methods. TOP$k$ accuracy indicates that the correct standard code is within the top $k$ closest codes ranked by cosine similarity. PBERT, PubMedBERT; BBERT, BioBERT; SBERT, SapBERT; BGE, BAAI General Embedding; GAME, Graph Alignment for Multi-institutional EHR Data; VA, U.S.\ Department of Veterans Affairs healthcare system; LOINC, Logical Observation Identifiers Names and Codes; LP, LOINC Part; $k$, number of top-ranked candidate codes considered.}
\label{R1}
\end{table}

We evaluated the capability of GAME to recover mappings from Veteran Affairs healthcare (VA) local laboratory codes to LOINC and LOINC Part (LP) codes. The ground truth was created through manual mapping by domain experts. Table~\ref{R1} presents the accuracy of recovering the correct mappings using embeddings from various PLMs. GAME achieved a TOP1 accuracy of $74.2\%$, whereas no other approach exceeded $62.2\%$. GAME also achieved a TOP10 accuracy of $90.7\%$. These results were obtained using only the similarity component of GAME with a dimensionality of $256$, whereas the other methods used higher-dimensional embeddings (Methods). We additionally compared lower-dimensional BGE and OpenAI embeddings, with results shown in Supplementary Table~\ref{R1_GPT}.

\begin{figure}[t]
    \centering
   \includegraphics[width=1\linewidth]{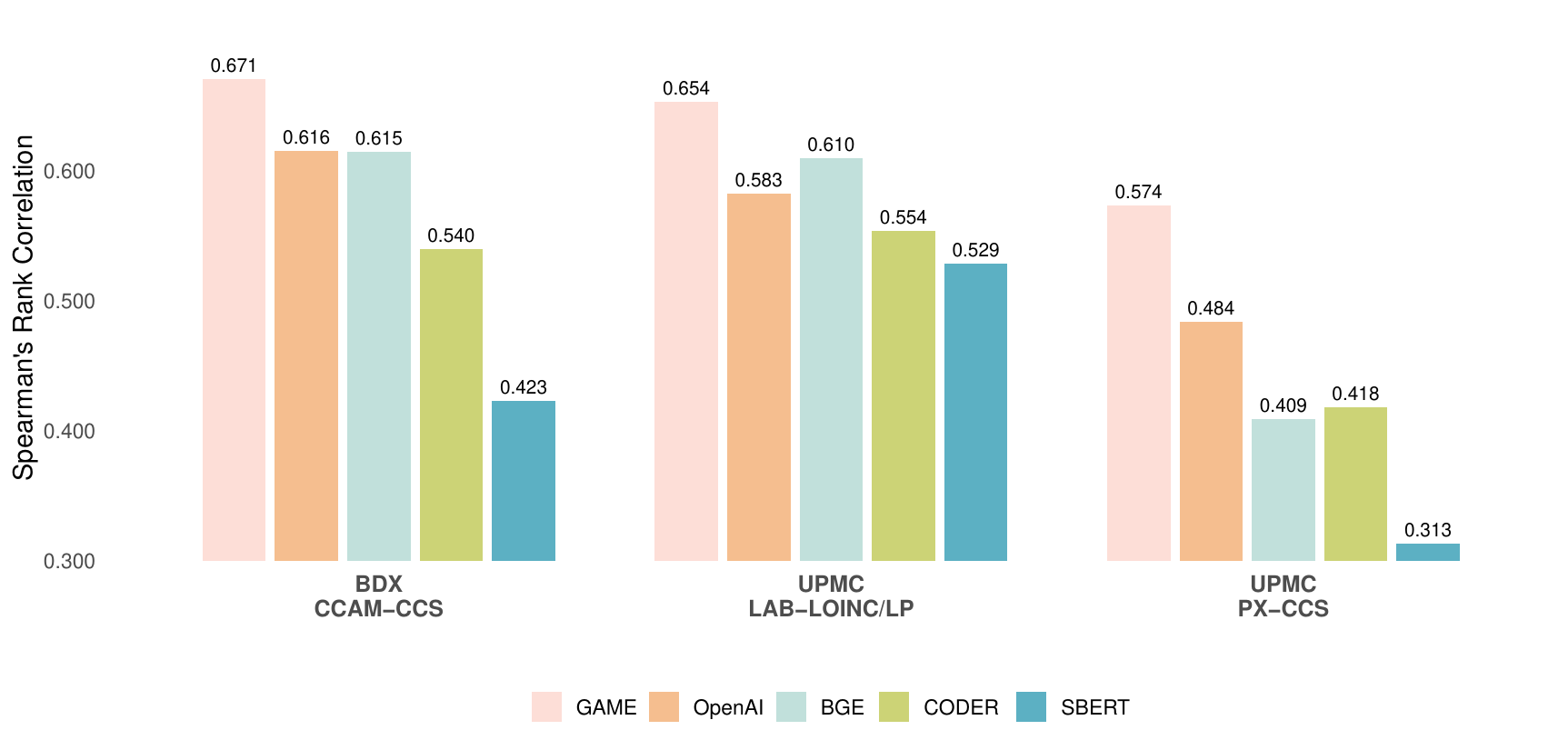}
    \caption{Rank correlations of GAME (Graph Alignment for Multi-institutional EHR Data) and pretrained language model similarities with human code mappings. Cosine similarities from GAME and other pretrained language models were compared with human-verified mappings of local codes to standardized codes at two institutions: Bordeaux University Hospital (BDX; Classification Commune des Actes Médicaux, CCAM, to Clinical Classifications Software, CCS) and University of Pittsburgh Medical Center (UPMC; local laboratory codes to Logical Observation Identifiers Names and Codes (LOINC) and LOINC Part (LOINC LP), and local procedure codes to CCS). Source Data are provided in the Source Data file.}
    \label{fig:code_map_corr}
\end{figure}

Fig.~\ref{fig:code_map_corr} shows Spearman rank correlations between cosine similarities and human annotations for mapping University of Pittsburgh Medical Center (UPMC) local procedures and Bordeaux University Hospital (BDX) Classification Commune des Actes Médicaux (CCAM) to Clinical Classification Software categories (CCS), and mapping UPMC local laboratory codes to LOINC/LP. In all three tasks, GAME achieved the highest correlation. Besides GAME, OpenAI and BGE performed well in parts of the tasks. Results for PPMI-SVD, BioBERT, and PubMedBERT are not shown in Fig.~\ref{fig:code_map_corr} because correlations were near zero; the full data table is provided in Supplementary Table~\ref{R2_code_map_2}.

\subsection*{Feature selection}

\begin{figure}[t]
\centering
\includegraphics[width=0.95\linewidth]{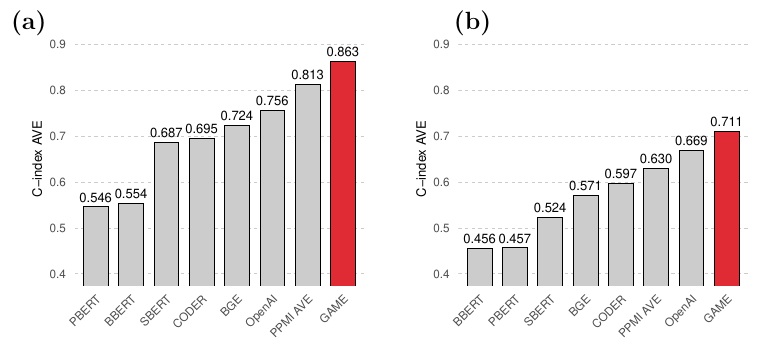}
\caption{Feature selection performance. (a) Concordance index (C-index) between expert-labeled relevance scores and cosine similarities derived from embeddings, averaged across six diseases. (b) Corresponding analysis using GPT-4-assigned scores as silver-standard labels. Red denotes the GAME method for visual emphasis only. Source Data are provided in the Source Data file.}
\label{P30}
\end{figure}

To assess whether GAME embeddings capture clinically meaningful relationships, we evaluated feature relevance for six diseases (Type 1 Diabetes, Epilepsy, Pulmonary Hypertension, Asthma, Crohn's Disease, and Ulcerative Colitis) using expert annotations and GPT-4-generated relevance scores. For each disease, we compared cosine similarity-based feature rankings across embedding methods and measured concordance with expert or GPT-4 relevance ratings. A higher concordance indicates better alignment between embedding-derived similarities and clinically validated feature importance. Fig.~\ref{P30} summarizes results using expert annotations and GPT-4 scores, with detailed results in Supplementary Tables~\ref{tab:fea_sel_human} and~\ref{R3_}. In Fig.~\ref{P30}a, average performance across diseases is shown using expert annotations; for clarity, we report the average over institutional PPMI embeddings (PPMI AVE). GAME outperformed other representations. In Fig.~\ref{P30}b, analyses using GPT-4 scores as silver-standard labels showed similar patterns. These findings, together with the correlation between GPT-4 scores and expert annotations (Supplementary Table~\ref{tab:fea_sel_human} and Supplementary Section~\ref{sec:gpt_make_sense}), support the use of GAME embeddings for selecting clinically meaningful features.

\subsection*{Joint patient profiling across institutions}

\begin{figure}[t!]
    \centering
    \includegraphics[width=0.9\textwidth]{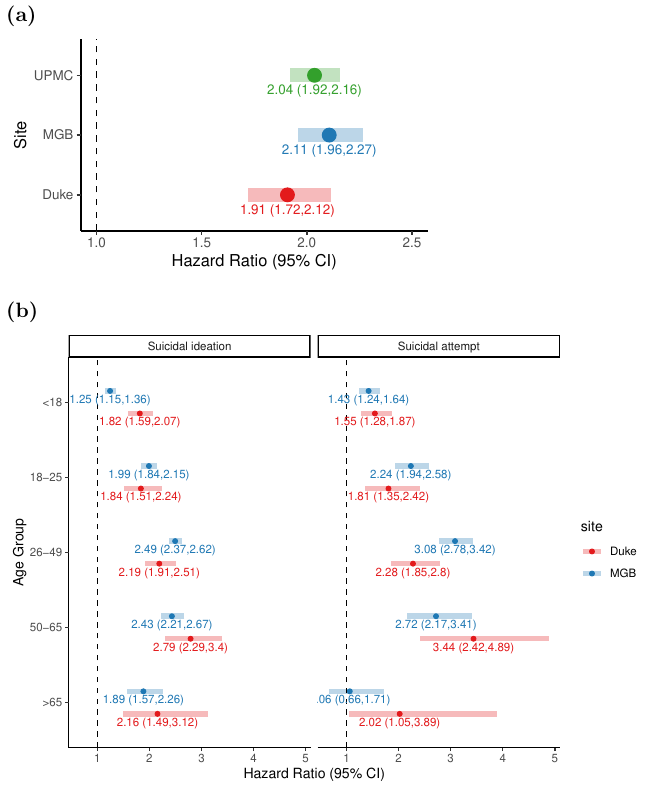}
    \caption{Risk across clinical subgroups identified by GAME (Graph Alignment for Multi-institutional EHR Data) embeddings. (a) Hazard ratios for future nursing home admission across Alzheimer's disease (AD) subgroups at University of Pittsburgh Medical Center (UPMC; $n=16{,}411$), Mass General Brigham (MGB; $n=17{,}770$), and Duke Clinical Research Datamart (Duke; $n=10{,}660$). (b) Hazard ratios for suicidal ideation and suicidal attempt across mental health subgroups, stratified by age group at MGB (age $<18$, $n=24{,}048$; age $18$--$25$, $n=36{,}666$; age $26$--$49$, $n=123{,}034$; age $50$--$65$, $n=72{,}482$; age $>65$, $n=39{,}509$) and Duke (age $<18$, $n=31{,}576$; age $18$--$25$, $n=23{,}271$; age $26$--$49$, $n=94{,}846$; age $50$--$65$, $n=70{,}651$; age $>65$, $n=51{,}867$). Here, $n$ represents the number of patients. Cox proportional hazards models were adjusted for age, sex, and race/ethnicity. Hazard ratios were estimated by comparing each subgroup to the reference subgroup within the same cohort. Results are presented as adjusted hazard ratios with two-sided $95\%$ confidence intervals (CI). Source Data are provided in the Source Data file.}

    \label{fig:HR_combined}
\end{figure}

We applied GAME embeddings to perform federated, multi-institutional patient profiling for disease progression analyses, examining risk of nursing home admission among patients with Alzheimer's disease (AD) and risk of suicide behaviors among patients with mental health disorders. Harmonized embeddings enabled federated joint clustering to identify latent patient subgroups with distinct trajectories. Subgroup membership was then incorporated into survival models to evaluate subgroup-specific differences in progression risk. For suicide risk profiling, patients were additionally stratified into age groups to account for age-related differences in clinical patterns. Patient-level data were obtained from UPMC, Mass General Brigham (MGB), and Duke Clinical Research Datamart (Duke) for AD and from MGB and Duke for suicide behaviors.

For AD progression, the AD cohorts included $16,411$ patients from UPMC, $17,770$ from MGB, and $13,438$ from Duke, with $64-65\%$  females and an average age of approximately $80$ years at first AD diagnosis. Additional demographic information is provided in Supplementary Table~\ref{tab:AD_demo}. Median follow-up was $78$ months at UPMC, $81$ months at MGB, and $36$ months at Duke. AD patient embeddings from all three institutions are visualized in Supplementary Fig.~\ref{fig:AD_tsne}. As expected, by representing EHR codes in each institution with GAME embeddings, patients from different institutions can be projected to the same embedding space, allowing for co-clustering of patient across multiple institutions. Federated $k$-means clustering of GAME-based patient embeddings at the time of AD diagnosis identified two subgroups: a fast-decline group (group 1), comprising 34.3\% to 59.3\% of patients across institutions, and a slow-decline group (group 2), comprising the remaining patients.

Cluster membership was associated with future risk of nursing home admission after adjustment for age, sex, and race/ethnicity (Fig.~\ref{fig:HR_combined}a). Kaplan--Meier (KM) curves for time to nursing home admission are shown in Supplementary Fig.~\ref{fig:AD_KM}. The fast-decline group had a higher risk of nursing home admission than the slow-decline group (median time to nursing home admission at UPMC: fast $37.1$ months, slow $86.3$ months; at MGB: fast $43.9$ months, slow $106.2$ months; at Duke: fast $94.0$ months, slow not reached).

Supplementary Fig.~\ref{fig:AD_phewas} highlights features associated with differences between the fast- and slow-decline groups within each institution. Fast-decline groups showed higher intensity of mental disorders and more frequent prescriptions of antipsychotics such as quetiapine and olanzapine \cite{cipriani2011comparative, arvanitis1997multiple} across institutions. This pattern suggests that cognitive and psychiatric comorbidities may have contributed to faster cognitive deterioration and a more challenging clinical trajectory \cite{ismail2022psychosis}. Abnormal movement, essential hypertension, and malaise and fatigue were also associated with the fast-decline group across institutions. At UPMC and Duke, patients in the fast-decline group were more frequently prescribed AD-related medications (for example, memantine, an N-Methyl-D-Aspartate (NMDA) receptor antagonist) \cite{liu2019role, tariot2004memantine} before diagnosis than patients in the slow-decline group, which may indicate cognitive and functional decline prior to a definitive AD diagnosis or diagnostic delay. This pattern was not observed at MGB, potentially due to differences in clinical or coding practice.

Prediction performance for nursing home admission models is presented in Supplementary Fig.~\ref{fig:AD_AUC}. Compared with models trained on raw EHR features, models using GAME-derived embeddings achieved higher predictive accuracy within sites and improved transportability across institutions.

For suicide risk assessment, GAME stratified mental health patients within each age group into two subgroups with distinct future risks of suicidal ideation and suicidal attempts. Demographic information is provided in Supplementary Table~\ref{tab:mental_demo_consolidated}, and patient embeddings are shown in Supplementary Fig.~\ref{fig:suicide_tsne}. Subgroups showed differential risks of suicidal ideation across age groups without adjustment and after adjustment for age, sex, and race/ethnicity (Fig.~\ref{fig:HR_combined}b). Risk of suicidal attempt was also associated with subgroup membership across age groups, except in the oldest age group (age $>65$; Fig.~\ref{fig:HR_combined}b). KM curves are shown in Supplementary Fig.~\ref{fig:KM_rpdr}. Key codes differentiating the clusters are shown in Supplementary Figs.~\ref{fig:suicide_phewas_phe} and~\ref{fig:suicide_phewas_oth}. Differences were primarily characterized by mental health conditions, such as mood disorders, major depressive disorder, and generalized anxiety disorder. In older age groups, conditions such as insomnia, essential hypertension, GERD, and symptoms like joint/back pain, nausea, vomiting, and fatigue also emerged as important differentiators between clusters.

Model performance of predicting suicide attempt is shown in Supplementary Fig.~\ref{fig:suicide_AUC} and Table~\ref{fig:suicide_AUC_table}. Compared to the existing clinical model, models trained with GAME-derived embeddings achieved comparable performance at MGB and outperformed it when evaluated at Duke. Additionally, the GAME-based models also outperformed those trained on raw EHR features.

\subsection*{Ablation study results}

We performed ablation studies to assess the contributions of design components in GAME, including edge construction, training schemes, and federated configurations. Configurations are summarized in Table~\ref{tab:ablation_studies}; implementation details are provided in the Methods.

\begin{table}[ht]
\centering
\begin{tabular}{ccc}
\toprule
Study \# & Training Type & Edge/Traning Configuration \\ \midrule
(a) & One-step & Hierarchical edges only \\ 
(b) & One-step & Hierarchical + UMLS edges \\ 
(c) & One-step & Hierarchical + UMLS edges + PPMI binomial edges \\
(d) & One-step & Hierarchical + UMLS edges + PPMI GPT-4 edges \\ 
(e) & One-step & Hierarchical + UMLS edges + PPMI GPT-4 + code mapping edges \\ 
(f) & One-step & PPMI GPT-4 + code mapping edges only \\ 
(g) & Two-step & No hard negatives (random same-type sampling for $\mathcal N_i$) \\ 
{ (h)} &{ Two-step} & { Centralized pooling of PPMI without  alignment} 
\\\bottomrule
\end{tabular}
\caption{Configurations used in the ablation studies to isolate the effects of edge types and training strategies in the GAME framework. GAME, Graph Alignment for Multi-institutional EHR Data; UMLS, Unified Medical Language System; PPMI, positive pointwise mutual information; GPT, generative pretrained transformer.}
\label{tab:ablation_studies}
\end{table}

As shown in Supplementary Table~\ref{edge_count}, a graph constructed solely from curated hierarchical and UMLS-derived edges is sparse. Incorporating GPT-4-generated edges increases the total edge count by approximately threefold and reduces sparsity. Supplementary Section~\ref{sec:S2} summarizes the impact of edge types and training strategies on similarity and relatedness detection, local code mapping, and feature selection.

For similarity and relatedness detection, GPT-4-selected PPMI edges (ablation d) outperformed binomial PPMI edges (ablation c). Curated UMLS and hierarchical edges (ablation f) also contributed. Hard negatives improved detection accuracy and supported $256$-dimensional embeddings, as indicated by the difference between ablation (g) and the full GAME model. Although ablation (h) performed comparably to GAME for relatedness detection, it performed worse for similarity detection, likely because it does not support similarity training effectively in low-dimensional spaces.

For local code mapping, binomial PPMI edges (ablation c) reduced performance, whereas GPT-4-selected PPMI edges (ablation d) provided gains. Adding GPT-4-selected local mapping edges (ablation e) yielded substantial improvement. The full GAME model matched or exceeded ablation (e), while ablation (g), which used random negatives, performed worse. Results from the VA local laboratory mapping analysis (Supplementary Table~\ref{tab:inst_ablation_code_mapping}) show that while graph neural networks capture intra-institutional structure, they do not align code semantics across sites, supporting the need for cross-institutional integration. Ablation (h) also lagged behind GAME by $3-4\%$ in TOP5, TOP10, and TOP20.

For feature selection, ablation (g) and the full GAME model outperformed other configurations, consistent with benefits of two-step training that separates similarity and relatedness signals. GAME further improved upon ablation (g) through hard negatives. Ablation (f), which used only GPT-4-generated edges, also performed competitively. Ablations (h) and (c) performed similarly, possibly because both introduce additional PPMI information in ways that are functionally similar for this task.

\section*{Discussion}

In this paper, we presented GAME, a framework that successfully harmonizes heterogeneous EHR codes by jointly integrating institution-specific co-occurrence statistics, ontological structures, and LLM-derived semantics. By relying on summary-level PPMI matrices rather than patient-level data, the framework ensures privacy and scalability. Our evaluations demonstrate that GAME embeddings outperform existing methods in code mapping, concept relatedness detection, and feature selection. Ablation studies confirm that these performance gains are driven by the synergistic integration of all framework components; consistent results are achieved only when multi-edge graph construction, cross-institutional alignment, GNN-based fusion, two-step training, and hard negative sampling are utilized together. Furthermore, the framework exhibits robust performance across different underlying models. Substituting SAPBERT with models such as BGE yields comparable results (Supplementary Section~\ref{sec:S2}), and concept expansion using GPT-4 produces results consistent with those from Claude-3~\citep{anthropic2024claude3} and LLaMA-4~\citep{meta2025llama4}, suggesting minimal model-induced bias (Supplementary Table~\ref{tab:llm_similarity}).

While GAME utilizes a GNN framework for relational learning, its primary contribution lies in the formulation of the harmonization problem and the systematic curation of heterogeneous relational information. By systematically combining KG edges, EHR derived summary statistics, and LLM-refined semantic relations, \textsc{GAME} provides a general, data-centric framework for  cross-institutional joint representation learning that is scalable  and privacy-preserving. Importantly, this curated relational structure is not tied to a single modeling backbone. In principle, it can support alternative architectures, including PLMs or hybrid graph–text approaches, provided task-specific objectives balance textual and relational supervision. To explore this, we fine-tuned a biomedical PLM using the same curated relations (Supplementary Section~\ref{sec:PLM}). However, its performance remained below that of GAME, highlighting that standard contrastive fine-tuning does not yet fully exploit the structured, heterogeneous supervision arising from EHR statistics and KG relations. Realizing the full potential of PLMs in this context will likely require customized loss formulations and explicit relational constraints to mitigate optimization instability, which remains an important direction for future research.

A fundamental advantage of GAME is its ability to handle the ``long tail'' of local coding practices without relying on rigid, predefined mappings. As shown in Table~\ref{stat}, approximately $69\%$ of codes in our dataset are local to a single institution (e.g., $60\%$ in MIMIC and $78\%$ in Bordeaux). GAME leverages shared standardized codes as anchors to align these local vocabularies within a unified semantic space. This distinguishes GAME from common data models like Observational Medical Outcomes Partnership (OMOP)~\citep{OMOP} or standards like Fast Healthcare Interoperability Resources (FHIR)  (\url{https://docs.smarthealthit.org/}); whereas those frameworks focus on schema normalization, GAME focuses on semantic harmonization. This design ensures adaptability to evolving systems—such as the introduction of 395 new ICD-10 codes in 2024 alone or the transition to ICD-11—without manual intervention. Furthermore, the framework produces general-purpose representations that transfer effectively to downstream tasks, such as AD stratification and suicide risk prediction, facilitating data-driven insights across diverse patient populations.

While the current implementation focuses on structured codes, GAME can be extended to incorporate unstructured clinical text and temporal dynamics. To assess the feasibility of integrating unstructured data, we extended our experiments to include Concept Unique Identifiers (CUIs) extracted from clinical notes in the MIMIC and VA datasets (Supplementary Section~\ref{sec:NLP}). After incorporating approximately $20{,}000$ CUIs, the framework maintained strong performance, with AUCs for CUI similarity and relatedness reaching $0.980$ and $0.956$, respectively, and it continued to perform well in code mapping and feature selection tasks (Supplementary Tables~\ref{R2-CUI}--\ref{R3_-CUI}). This confirms that GAME can robustly harmonize both structured and text-derived entities. Regarding temporal dynamics, the current PPMI formulation aggregates co-occurrences within a window to maintain scalability. However, medical sequences contain valuable directional information. Future methodological extensions could incorporate directional PPMI matrices or causal message passing—similar to approaches like \cite{wen2025dome}—to capture disease trajectories. Additionally, future work could explore unsupervised alignment methods (e.g., optimal transport~\citep{chen2020graph}) to further support settings with minimal shared ontologies, and expand human annotation efforts to validate the heuristic guidance provided by LLMs.

In summary, GAME provides a robust and modular infrastructure for harmonizing EHR concepts across distributed healthcare systems. By unifying co-occurrence patterns, curated knowledge, and real-world semantics into a privacy-preserving architecture, it lays the groundwork for broader interoperability in biomedical research. Its emphasis on transparency and semantic consistency ensures that EHR-driven machine learning can evolve equitably, supporting scalable and interpretable models across the heterogeneous landscape of global healthcare data.

\section*{Methods}

This research was conducted in accordance with all relevant ethical regulations. 
The study protocol was reviewed and approved by the institutional review boards (IRBs) or ethics committees of the participating institutions, including Boston Children’s Hospital (IRB-P00031305, ``Large-scale computable phenotyping for pediatric cohorts''; IRB-P00037726, ``Longitudinal Molecular Genetics Study of Lung Function Growth''), Mass General Brigham (IRB protocol 2018P000609, ``Resource Core: Bioinformatics--Subproject for Institution \#2016D008375'', active, data analysis only), Duke University (Protocol ID: Pro00110219, ``Development of novel statistical and machine learning methods for predictive modeling using electronic medical records data''), University of Pittsburgh Medical Center (Protocol ID: STUDY21020153, ``Integrating Electronic Health Records for Alzheimer’s Disease Research''), the U.S. Department of Veterans Affairs Central IRB (cIRB 18-38, ``MVP-CHAMPION: Innovative Analytics for Big Data in the VA''), and the Centre d’Éthique et de Recherche en Santé Bordeaux (CERS; record number CER-BDX 2024--84). 
For institutions providing non-public electronic health record (EHR) data, the requirement for written informed consent was waived by the approving IRBs or ethics committees because this was a retrospective analysis of existing clinical data and posed minimal risk to participants, in accordance with local regulations. 
For the publicly available MIMIC-IV dataset, data access and use complied with the corresponding data use agreement and ethics approvals associated with the resource. 
No participant compensation was provided, as this study involved secondary use of existing clinical data.

The GAME algorithm is built upon the Graph Attention Network (GAT) \cite{GAT}, a variant of GNNs \cite{gori2005new}, which serves as the core architecture for harmonizing multi-institutional EHR code embeddings. In this framework, EHR codes from multiple institutions are represented as nodes in a unified graph, 
with edges capturing diverse relationships derived from three sources: (i) co-occurrence statistics within EHR data, (ii) biomedical knowledge bases including UMLS and ontologies, and (iii) GPT-4-informed semantic similarity and clinical relatedness. GAT is particularly well-suited for this setting, where the strength of the relationship between a target node (e.g., a disease) and its neighboring nodes (e.g., medications, comorbidities) can vary widely. Its attention mechanism dynamically weighs the influence of neighboring nodes during message passing, allowing the model to focus on the most relevant, clinically meaningful signals. Unlike traditional GNNs that aggregate all neighbor information uniformly, GAT selectively amplifies important connections, enabling context-aware learning across heterogenous data sources.

 While GAME leverages established components such as GNNs, PLMs, and KGs, it introduces several key innovations tailored to the challenges of multi-institutional EHR representation learning:
\begin{itemize}
    \item Multi-source Graph Construction: We integrate structured biomedical knowledge with institution-specific EHR statistics and GPT-4-guided similarity judgments to construct a richly connected, heterogeneous graph that captures both standardized and local usage patterns beyond those represented in UMLS.
    \item Federated Representation Harmonization: Embeddings are trained in a privacy-preserving manner using only summary-level inputs, enabling cross-institution learning without sharing patient-level data or requiring identical EHR entities across sites. This design achieves federated representation learning in the embedding space---an often assumed but rarely addressed step in prior FL frameworks.
    
    \item Contrastive Learning with Curated Hard Negatives: We construct high-quality positive and negative training pairs using both PPMI statistics and GPT-4 informed predictions. Ablation studies demonstrate that this targeted selection of hard negatives substantially enhances embedding quality, improving cross-institution code mapping and feature selection.
\end{itemize}

Building on these innovations, we next describe the EHR datasets from seven institutions and their preprocessing steps, followed by the initialization of embeddings derived from EHR co-occurrence patterns and PLMs. We then detail the construction of KG edges used for supervision, present the full GAME training pipeline, and conclude with the validation strategy and the ablation study design. 

\subsection*{EHR data sources and preprocessing}

We utilized EHR data from seven hospital systems: Boston Children's Hospital (BCH), Bordeaux University Hospital (BDX), Duke Clinical Research Datamart (Duke), Mass General Brigham (MGB), Medical Information Mart for Intensive Care IV (MIMIC)  \citep{mimiciV}, University of Pittsburgh Medical Center (UPMC), and Veteran Affairs healthcare (VA). In this study, BCH contains $251$K patients from $2009$ to $2022$; BDX EHR data covers $2.5$ million patients from $2010$ to $2023$; Duke includes data on over $6$ million patients from $2014$ to $2023$; MGB EHR data contains $2.5$ million patients from $1998$ to $2018$; and the VA Corporate Data Warehouse (CDW) aggregates data from $150$ VA facilities into a single data warehouse, with records from $1999$ to $2019$ covering $12.6$ million patients. BCH, BDX, Duke, and VA include inpatient and outpatient codified data from patients with at least one visit, while MGB includes only patients with at least three visits spanning more than $30$ days. The MIMIC dataset contains data on over $65$K ICU admissions and over $200$K emergency department admissions at Beth Israel Deaconess Medical Center in Boston, Massachusetts, spanning $2008$ to $2019$. The UPMC EHR data includes $95$K patients from $2004$ to $2022$, focusing on individuals with at least one occurrence of ICD codes related to Alzheimer's disease and dementia  or multiple sclerosis.

We map ICD codes to PheCodes using the PheWAS catalog (\url{https://phewascatalog.org/phecodes}).  Diagnostic codes that cannot be mapped to PheCodes are retained as individual codes. All
Current Procedural Terminology (CPT), Healthcare Common Procedure Coding System (HCPCS), and ICD procedure codes are grouped into Clinical Classification Software categories (CCS)  using the CCS mapping (\url{https://www.hcup-us.ahrq.gov/toolssoftware/ccs_svcsproc/ccssvcproc.jsp}). BDX uses the Classification Commune des Actes Médicaux (CCAM) \cite{bousquet2010evaluation}, totaling $5246$ distinct codes, for procedures.  For laboratory tests, all laboratory codes from MGB and BCH are mapped to LOINC, whereas laboratory codes from VA, UPMC, Duke, MIMIC, and BDX are primarily local. Similarly, for medication codes, we group them into ingredient-level RxNorm codes when possible and retain others, primarily from UPMC and BDX, as local codes. In this study, we refer to PheCode, CCS, LOINC, and RxNorm as standard codes, while other codes that occur in local institutions are referred to as local codes. The goal of the algorithm is to accurately map local codes to the appropriate
standard codes (Fig.~\ref{fig:map-codes-gpt4}).

\def\Vsc{\mathcal{V}}

When constructing the PPMI matrix, co-occurrences fewer than $10$ times are set to $0$, and any code that co-occurs with other codes fewer than $10$ times is removed to reduce noise from rare codes. Since laboratory tests are frequently analyzed at the first level of LOINC PART (LP) codes—parent codes representing groups of individual LOINC codes—we further aim to create embeddings for these LP codes. Including LP codes, we have $5,745$ codes at BCH, $26,632$ codes at BDX, $3,125$ codes at Duke, $6,969$ codes at MGB, $4,341$ codes at MIMIC, $17,271$ codes at UPMC, and 6,660 codes at VA. In total, we obtained $70,743$ codes, of which \( N = 50,738 \) were unique. Our goal is to create unified embeddings for these \( N \) unique EHR codes, denoted by \(\mathcal{V}\).

A detailed summary of the codes used is provided in Table~\ref{stat}. The table shows substantial cross-site variation in the proportion of non-standard (local) codes, ranging from 0\% at MGB and BCH to nearly 80\% at BDX, with an overall average close to 70\%. 
It also documents large differences in vocabulary size (roughly $3{,}000$ to over $26{,}000$ unique codes per site) and patient counts (from $95$K to $12.6$M). Taken together, these observations indicate that harmonization must handle both heterogeneous vocabularies and highly imbalanced data scales, motivating our use of federated alignment rather than simple centralized pooling.

\begin{table}[H]
\footnotesize
    \centering
    \resizebox{\textwidth}{!}{
 \begin{tabular}{ccc|cccccc} \toprule
 \multirow{2}{*}{ {Institution}} & \multirow{2}{*}{ {Location}} & \multirow{2}{*}{ {Patients}} & \multicolumn{6}{c}{{Unique codes}} \\
 & & & {PheCode} & {CCS} & {LOINC} & {RxNorm} & {Non-standard} & {Total} \\ \midrule 
 BCH  & {\scriptsize NE US} & $251$K  & 1405  &  199  & 3024 & 1117 & 0 (0) & 5745 \\ 
BDX & {\scriptsize France}  &  $2.5$M &  1656 &  0&   2935 & 1103  &  20938 (78\%)& 26632 \\ 
Duke & {\scriptsize SE US} & $6.0$M & 1439 &  0 &   554 & 278     & 854 (27\%)  & 3125 \\ 
MGB & {\scriptsize NE US}  & $2.5$M & 1772  & 243 & 3719 & 1235 & 0  (0)  & 6969 \\ 
MIMIC & {\scriptsize NE US} & $265$K & 637 &  129 & 0  & 959    & 2616 (60\%) & 4341\\ 
UPMC & {\scriptsize NE US} & $95$K & 1841 & 245 & 5127 & 1987  & 8071 (47\%) & 17271\\  
VA   & {\scriptsize US} & $12.6$M & 1776 & 224 & 730 & 1469   & 2461 (37\%) & 6660 \\ \midrule
 Total & - &  24.2M & 1869 & 248 & 9410 & 4271 & 34940 (69\%) & 50738\\ \bottomrule
\end{tabular} }
    \caption{Types of electronic health record (EHR) codes studied across the seven institutions. CCS, Clinical Classifications Software; LOINC, Logical Observation Identifiers Names and Codes. NE = northeast and SE = southeast. BCH = Boston Children's Hospital, BDX = Bordeaux University Hospital, Duke = Duke Clinical Research Datamart, MGB = Mass General Brigham, MIMIC = Medical Information Mart for Intensive Care IV, UPMC = University of Pittsburgh Medical Center, and VA = U.S.\ Department of Veterans Affairs healthcare system. The proportion of non-standard codes within each institution is shown in parentheses.}
    \label{stat}
\end{table}

\subsection*{Generating multi-source initial embeddings}

We next outline the generation of two sets of initial embeddings for the $N$ unique EHR codes: institutional PPMI-SVD embeddings and textual description embeddings, essential for capturing clinical and semantic information of medical codes.  The left part of Fig.~\ref{fig:outline} provides an intuitive illustration of the initial embedding generation process.

\paragraph{Institution-specific PPMI-SVD embeddings from co-occurrence patterns.}
We construct a co-occurrence matrix to derive PPMI-SVD embeddings for each institution to capture the interactions between EHR codes based on their co-occurrence within patient records, following the methodology described in \cite{beam2018clinical} and \cite{hong2021clinical}. For each institution $m \in \{1, \ldots, M\}$, the matrix $\Csc_m = \big[\Csc_m(i,j)\big]$ records the frequency of co-occurrences between the $i$-th and $j$-th codes within $30$-day windows. We adopt a 30-day co-occurrence window as proposed in prior research \cite{hong2021clinical, MIKGI, gan2023arch}.

We generate PPMI-SVD embeddings for all codes in the $m$th institution by applying SVD to the PPMI matrix. To create initial embeddings for LP codes, we initialize the embedding of each LP code as a weighted average of the embeddings of its associated child LOINC codes. Detailed steps for generating the PPMI-SVD embeddings for both the EHR base codes and LP codes in the $m$th institution are provided in Supplementary Section~\ref{supp:ppmi}. The resulting $d$-dimensional PPMI-SVD embeddings are denoted by $\V_{m}$. For ease of implementation, we transform $\V_m$ into an $N\times d$ matrix to represent the embeddings of the $N$ unique EHR codes across all $M$ institutions by padding with zeros for the codes that do not appear in the $m$th institution.

\paragraph{Embedding textual description using PLMs.} 
PLMs have proven to be highly effective in identifying biomedical relationships and generating high-quality semantic embeddings \cite{lee2020biobert, shin2020biomegatron, gu2021pubbert,liu-2021-sapbert,yuan2022coder,chen2024bge}. 
We use SAPBERT embeddings \cite{liu-2021-sapbert}, a PLM fine-tuned on UMLS synonymous relationships, to extract semantic information from the textual descriptions of EHR codes. However, the effectiveness of these embeddings depends significantly on the quality and specificity of the underlying descriptions. Many local codes, unfortunately, have abbreviated or vague descriptions. For example, the VA local lab code 1000023750 is described merely as ``TIBC,'' whereas its more detailed and informative description is ``Iron binding capacity [Mass/volume] in Serum or Plasma.'' To address this limitation, we employ GPT-4 to generate more detailed and informative descriptions for local codes. We first expand laboratory test acronyms into full names using resources such as the Laboratory Alliance's list of test abbreviations (\url{https://www.laboratoryalliance.com/healthcare-providers/laboratory-services/test-abbreviations/}). Next, we prompt GPT-4 to produce comprehensive descriptions for the local codes,  with the corresponding prompts provided in Supplementary Table~\ref{tab:desc_prompt}. For French descriptions from BDX, we first use GPT-4 to translate the text into English, ensuring compatibility with SAPBERT. Based on these GPT-expanded descriptions, we create SAPBERT embeddings for all codes in $\Vsc$, denoted by $\X$.

While SAPBERT embeddings are used as input to the GAME algorithm, we also generate other PLM embeddings, including BioBERT, PubMedBERT, CODER, BGE, and OpenAI text-embedding-3-small (OpenAI). These embeddings help identify candidate code pairs, which are subsequently evaluated by GPT-4 to generate similarity labels used during the contrastive learning step of GAME training, as detailed below. 

\subsection*{Curation of the adjacency matrix}

We constructed a comprehensive adjacency matrix from multiple knowledge sources (Fig.~\ref{fig:outline}), which serves as input to the graph attention network (GAT) in the GAME algorithm.  The matrix integrates edges from three primary sources: (1) hierarchical code structures, including PheCode, LOINC, RxNorm, and CCAM; (2) relationships provided by UMLS; and (3) additional edges generated using GPT-4.  We categorize relation pairs into two groups: similar pairs and related pairs. The process for constructing edges between different EHR codes is described below.  
A summary of the number and types of edges is provided in Supplementary Table~\ref{edge_count}. The final set of edges, representing the curated KG, is denoted as $\mathcal{E}$.

\paragraph{Hierarchical information from common ontologies.} We derive edges from the hierarchical structures of PheCode, RxNorm, LOINC, and CCAM to capture relationships between similar codes. In PheCode, more digits indicate greater specificity (e.g., PheCode 296 for mood disorders is the parent of PheCode 296.1 for bipolar disorder and PheCode 296.2 for depression, with PheCode 296.22 for major depressive disorder as a further refinement). We connect PheCodes sharing the same integer part. In LOINC, edges link codes to their parent LP codes and between codes with the same LP parent. For RxNorm, where all codes we use are leaf codes, we create edges between those with common parents. For CCAM, codes sharing the first four characters (e.g., ``GLLD015'' and ``GLLD008'') are connected.

\paragraph{UMLS.} 

UMLS includes a broad range of annotations, capturing similarity and relatedness in medical relationships between entity pairs. For similarity, in addition to the hierarchical relationships from common ontologies, UMLS includes non-hierarchical relationships that do not follow a strict parent-descendant structure. For relatedness, we consider relationships like ``associated with'', ``may treat'' and ``co-occurs with'', as detailed in Table~\ref{num_table} in Supplementary Section~\ref{sec:data_sup}. However, since the UMLS concepts are mostly encoded as CUIs, their relationships cannot be directly translated to relationships for EHR codes. We map CUIs to EHR codes using both existing mappings (RxNorm, CCS, LOINC to CUIs) (\url{https://bioportal.bioontology.org/ontologies}) and GPT-4 (PheCode to CUIs). Specifically, for PheCode to CUIs, we use SAPBERT to choose the most similar PheCode for each CUI whose semantic type is ``Disease or Syndrome.'' We then prompt GPT-4 to annotate whether the CUI can be mapped to the selected PheCode and retain only the positive pairs as the final CUI-PheCode mappings.

\paragraph{Graph expansion derived using GPT-4.}

While ontology hierarchies and UMLS provide valuable insights into medical relationships, they are sparse and insufficient for diverse downstream tasks. First, they lack connections between EHR codes from different institutions, such as relationships involving local codes. Second, they indicate whether two codes are related but do not quantify the closeness of these relationships, limiting their precision.

To address these gaps, we leverage GPT-4 as a lightweight \emph{denoising and labeling module} to curate edges for code mapping and the identification of related codes as additional training labels. All GPT-4 calls are restricted to small, pre-filtered candidate sets obtained through embedding-based similarity search, ensuring both computational feasibility and bounded context length. All duplicate code–code edges ($\approx 7\%$ overlap across sources) were deterministically removed by exact code matching before training, ensuring a unique, non-inflated graph without affecting efficiency or performance.

\begin{figure}[t!]
    \centering
    \includegraphics[width=1\textwidth]{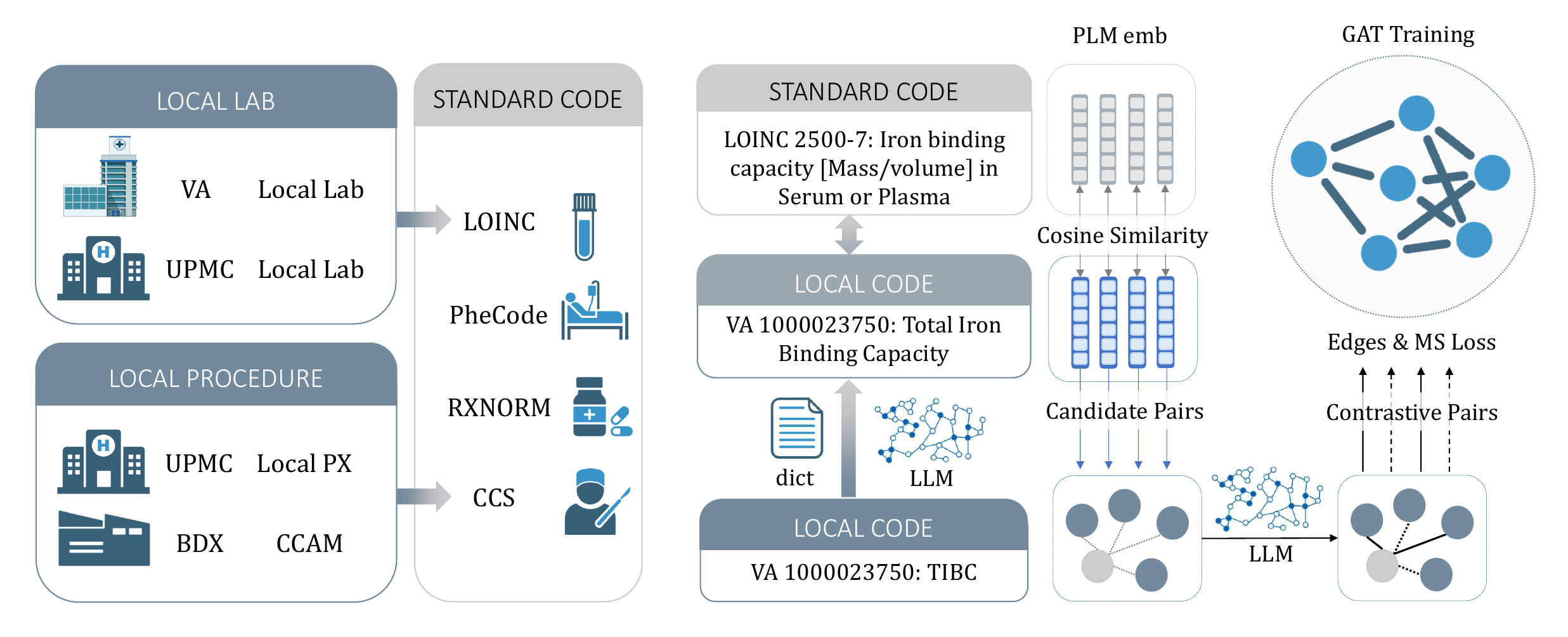} 
    \caption{Mapping local electronic health record (EHR) codes to standardized vocabularies using GPT-4. Local laboratory and procedure codes from multiple institutions are aligned to standardized vocabularies, including Logical Observation Identifiers Names and Codes (LOINC) and Clinical Classifications Software (CCS). GPT-4 evaluates candidate pairs selected by pretrained language-model (PLM) embeddings based on cosine similarity, using code descriptions to generate positive and negative relationships for contrastive learning. These GPT-4-curated pairs are incorporated as additional edges during graph attention network (GAT) training, enhancing cross-institutional representation alignment within the GAME (Graph Alignment for Multi-institutional EHR Data)  framework. Created using BioRender.com.}
    \label{fig:map-codes-gpt4}
\end{figure}

\paragraph{GPT-4-guided mapping pairs.} 
To facilitate the training of embeddings for local codes, we generate edges that map these codes to common ontologies. For each local code, we calculate its cosine similarity with standardized codes using different PLM embeddings (SAPBERT, CODER, BGE, and OpenAI), as shown in Fig.~\ref{fig:map-codes-gpt4}. For example, we map local VA lab codes to LOINC, and local BDX procedure codes to CCS.  For every local code, we retain the top $20$ standardized codes with the highest cosine similarity from each embedding model and take their union (typically $\leq 60$ candidates per code). Only candidates with high-similarity scores are sent to GPT-4o for binary classification as \emph{correct} or \emph{incorrect} mappings. For each local code, all its candidate pairs are asked in one prompt (about $2\,000$ tokens per call). Given that there are about $2\,000$ local labs in VA, this corresponds to roughly $2\times10^3$ GPT-4 evaluations and results in a total cost of approximately US\$5 for this institution. The GPT-4-labeled correct and incorrect mappings are used as training data in the contrastive learning step.  Pairs that are textually similar but labeled incorrectly by GPT-4 serve as ``hard negatives,'' which are especially valuable for refining the embedding geometry.

\paragraph{GPT-4-confirmed relevance pairs.} A key advantage of PPMI-SVD embeddings derived from EHR data is their ability to capture positive and negative pair information. We further leverage GPT-4 to create related edges and ``hard-negative'' related pairs. { To keep the scale tractable, we pre-filter pairs before prompting. For each combination of code types (e.g., RxNorm–PheCode, LOINC–LOINC), we select only the top $0.1\%$ of pairs with the highest cosine similarity across any institutional embedding $\V_m$, reducing billions of potential pairs to about $6\times10^5$. 
GPT-4o-mini then labels these candidate pairs as ``clinically related'' or ``unrelated'' in batches of size $50{,}000$. 
Batch prompting streamlines the process and reduces the overall cost by roughly half, enabling scalable inference. Pairs labeled as unrelated yet exhibiting high embedding similarity are retained as additional hard negatives.}
This process yields $130{,}801$ positively related pairs and $467{,}580$ negative pairs, as summarized in Table~\ref{edge_count} in Supplementary Section~\ref{sec:data_sup}.   Overall, GPT-4 is applied only to sizable, high-confidence candidate sets, ensuring both computational efficiency and practical feasibility at the scale of our merged EHR concept graph.

\begin{figure}[t]
    \centering
   \includegraphics[width=\textwidth]{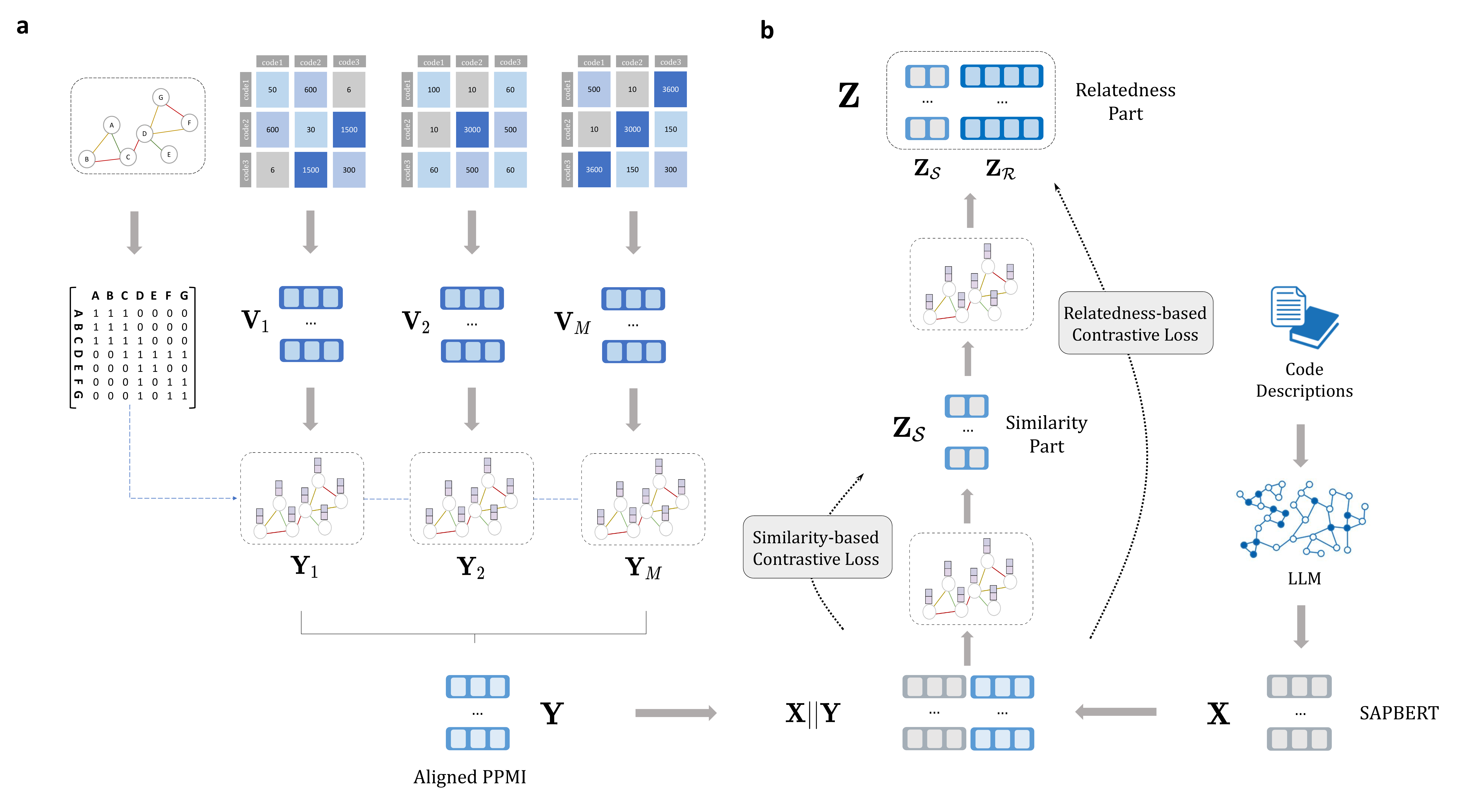}
    \caption{Overview of key steps in the GAME algorithm. (a) Alignment of embeddings from multiple institutions into a shared representation space. Institution-specific positive pointwise mutual information (PPMI) matrices are converted into embeddings and harmonized through graph alignment to produce the aligned embedding $\mathbf{Y}$. (b) Sequential learning of similarity and relatedness using contrastive objectives. The two-step graph attention network (GAT) training yields embeddings $\mathbf{Z}_{\Ssc}$ and $\mathbf{Z}_{\Rsc}$ that jointly capture semantic similarity and clinical relatedness, forming the unified representation space for GAME. Created using BioRender.com.}
    \label{fig:alg-key}
\end{figure}

\subsection*{The GAME algorithm}

As shown in Fig.~\ref{fig:alg-key}, the training of the GAME algorithm comprises two key steps: (1) learning initial embeddings by aligning $M$ sets of PPMI-SVD embeddings into a shared representation space, enhanced with KG information, using GAT; and (2) sequentially learning similarity and relatedness embeddings by integrating these initial embeddings with PLM embeddings through GAT combined with contrastive learning. This process utilizes positive and hard-negative similarity and relatedness labels from multiple sources, as described above.

\paragraph{KG-enhanced alignment of PPMI-SVD embeddings through GAT.} 

We first train an initial set of harmonized embeddings for all EHR codes $\Vsc$ by aligning PPMI-SVD embeddings $\{\mathbf{V}_{m}\}_{m=1,...,M}$, incorporating curated KG $\mathcal{E}$ using GAT. As detailed in Algorithm~\ref{alg:align_sppmi} in Supplementary Section~\ref{loss}, a GAT is trained for each institution, taking 
$\mathbf{V}_{m}$ and $\mathcal{E}$ as inputs, and outputting embeddings  $\mathbf{Y}_m$:
\begin{align}
    \mathbf{Y}_m &=  \text{Linear}^{(m)}\left\{\textbf{GAT}^{(m)}\left(\V_m,\mathcal{E}\right)\right\},  \label{Align}
\end{align}
where the linear layer, $\text{Linear}^{(m)}$, further aligns the institutional embedding into a shared space. The GAT training is regularized with the alignment loss   
\begin{equation}
    \sum_{m_1, m_2 \in \{1,2,\ldots,M\}}\|\mathbf{Y}_{m_1}[\Isc_{m_1}, : ] -  \mathbf{Y}_{m_2}[\Isc_{m_1}, : ] \|_{\rm F}^2 ,
\label{align_loss}
\end{equation}
where $\Isc_m$ indexes codes appearing in the $m$th institution. This loss minimizes the difference between the embeddings of corresponding medical codes across institutions. The initial harmonized embeddings are obtained as
$\mathbf{Y} = \Rscr\left\{\sum_{m=1}^M\mathbf{Y}_m\right\}$, where $\Rscr(\mathbf{V})$ normalizes a given vector $\mathbf{V}$ to unit norm.

\paragraph{Sequential GAME embedding training with contrastive learning.}

We concatenate the PLM embeddings for code descriptions, 
$\X$, with the initial harmonized embeddings, 
$\Y$, to serve as input for sequentially training similarity and relatedness GAME embeddings. The goal is to generate a set of unified GAME embeddings $\Z = [\Z_{\Ssc}, \Z_{\Rsc}]$ that integrate descriptive code information and EHR knowledge across institutions. Here, $\Z_{\Ssc}$ is specifically optimized to capture semantic similarity, while the full embedding $\Z$ is designed to support more complex downstream tasks. Since similarity represents a stronger and more direct relationship, low-dimensional embeddings are sufficient for its representation. In contrast, tasks involving relatedness demand richer representations to capture broader and more nuanced semantic relationships. To achieve this, training is performed using GAT with contrastive learning, leveraging the edge set $\Esc$. Various edge types contribute to different components of the contrastive loss, incorporating hard-negatives derived from ontology hierarchies as well as those generated by GPT-4.

We start with learning similarity embeddings using GAT, formulated as: 
\begin{equation}
    \Z_{\mathcal{S}} =  \Rscr \left\{
\text{Linear}_{\mathcal{S}}\left\{\text{GAT}_{\mathcal{S}}\left([\X,\Y],\mathcal{E} \right)\right\} 
    \right\}.
    \label{SIMI} 
\end{equation}
This is optimized with a similarity contrastive loss $\Lscr_{\Ssc}$, where positive and negative pairs are derived from UMLS similarity relations, ontology hierarchies, and GPT-enhanced edges. The inclusion of hard-negatives from the ontology and GPT further strengthens the training process.   
To train $\Z=[\Z_{\Ssc}, \Z_{\Rsc}]$, we fix $\Z_{\mathcal{S}}$ from the first step and learn $\Z_{\mathcal{R}}$ using another GAT:
\begin{equation}
    \Z_{\mathcal{R}} =  \Rscr\left\{
\text{Linear}_{\mathcal{R}}\left\{\textbf{GAT}_{\mathcal{R}}\left([\X,\Y], \mathcal{E} \right)\right\} 
    \right\}.
    \label{RELA} 
\end{equation}
This is optimized using a relatedness contrastive loss $\Lscr_{\Rsc}$ that integrates UMLS relatedness edges with GPT-enhanced edges derived from EHR PPMI matrices, enabling robust learning of more nuanced and complex semantic relationships.

The contrastive losses $\Lscr_{\Ssc}$ and $\Lscr_{\Rsc}$ are constructed based on the Multi-Similarity (MS) loss \citep{wang2019multi}, which effectively handles tasks with multiple semantic relationships by dynamically balancing the pulling of positive pairs and pushing apart negative pairs. For the $i$th anchor code with a set of positive pairs $\mathcal{P}_i$ and a set of negative pairs $\mathcal{N}_i$, the MS loss is defined as:
\begin{equation} \mathcal{L}^{(i)} (\mathbf{Z}) = \frac{1}{\alpha} \log \left( 1 + \frac{1}{|\mathcal{P}_i|} \sum_{j \in \mathcal{P}_i} e^{-\alpha (\mathbf{Z}_i^\top \mathbf{Z}_j - \lambda)} \right) + \frac{1}{\beta}\log \left( 1 + \frac{1}{|\mathcal{N}_i|} \sum_{j \in \mathcal{N}_i} e^{\beta (\mathbf{Z}_i^\top \mathbf{Z}_j - \lambda)} \right). \label{M_S} \end{equation}
Here, $\alpha$, $\beta$, and $\lambda$ are hyperparameters that control the strength of the loss. See Supplementary~\ref{loss} for details on the losses and Algorithm~\ref{alg:encoder} in Supplementary~\ref{loss} for details on the sequential GAME embedding training.

Since SAPBERT embeddings use a feature dimension of \(d = 768\), a common choice for pre-trained language models, we retain this dimension for the SAPBERT embeddings \(\X\), PPMI-SVD embeddings \(\V_{m}\)'s, aligned embeddings \(\Y\), and final embeddings \(\Z\). For similarity embeddings \(\Z_{\mathcal{S}}\), we reduce the dimension to $256$,  enabling relatedness embeddings \(\Z_{\mathcal{R}}\) to use  $512$ dimensions.  Supplementary Section~\ref{Tune} provides details on hyperparameter tuning, and Supplementary Section~\ref{supp:gpu_time} summarizes the GPU specifications and training times.

\subsection*{Detecting similarity and relatedness between codes}

We first evaluated the quality of embeddings with respect to their ability to detect semantic similarity and clinical relatedness between EHR codes. We split the known similarity and relatedness pairs from PheCode, RxNorm, LOINC, CCAM, and UMLS into training and validation sets, as detailed in Supplementary Section~\ref{sec:data_sup}.
 For each relationship type, we computed cosine similarities for positive pairs and type-matched negative pairs. For example, negatives for ``may treat'' or ``may prevent'' were sampled from disease-drug pairs, and for LOINC hierarchies, from LOINC-LOINC pairs. This matching controls for confounding due to semantic type differences, ensuring that performance reflects true relational discrimination rather than trivial mismatches.
Additionally, all relationships—both similar and related—were grouped by code type (e.g., ``PheCode–RxNorm'', ``LOINC–LOINC'') to enable clear stratified analysis and produce stable AUC estimates. Within each group, we computed the AUC using cosine similarity scores to assess the model’s ability to distinguish true positive pairs from semantically plausible but incorrect ones.

Due to variation across individual institutions, we also conducted institution-specific analyses to evaluate GAME’s performance in a more localized context, thereby enhancing the interpretability and robustness of our findings. Specifically, we identified code pairs that appeared in only one or two institutions and referred to them as institution-specific pairs. We then evaluated similarity and relatedness detection using institutional PPMI, various PLM-based embeddings, and GAME, reporting AUC scores separately for each institution. Details are provided in Supplementary Section~\ref{result_AUC_insti}.

\subsection*{Translating and mapping codes across EHR systems}

We evaluated the accuracy of mapping local codes to a common ontology using embeddings, against gold standard labels assembled via human curation. We considered four sets of mappings: 1) local VA lab codes to the first level LP codes, with $11,808$ curated mappings; 2) BDX CCAM procedure codes to CCS, with $537$ curated mappings; 3) UPMC local procedure codes to CCS, with $199$ curated mappings; and 4) UPMC local lab codes to LP codes, with $1,814$ curated mappings. 

Of these four sets of mappings, only the VA mappings were previously curated at scale, with detailed background knowledge about the codes in the Observational Medical Outcomes Partnership (OMOP) \cite{OMOP}, which allows us to examine the top $k$ accuracy of the codes for each set of embeddings. The top $k$ accuracy is defined as the proportion of test cases in which the correct mapping for a given code appears among the top 
$k$ predictions generated by the embeddings. 

The remaining three sets were curated only for a subset of pairs sampled according to the embedding-based cosine similarities, as detailed in Supplementary Section~\ref{supp: upmc bdx mapping}. Because of potential ambiguity in the code descriptions for these three sets, the annotators assigned one of the following labels: ``Highly likely to be mapped'' (confirming that the mapping is correct), ``Maybe mappable'' (suggesting that the mapping is potentially correct but could be more precise), and ``Not likely to be mapped'' (indicating that the mapping is incorrect). We evaluated Spearman's rank correlation between the embedding-assigned cosine similarities from each method and the annotated labels.

\subsection*{Feature selection}

Feature selection is a critical step in many downstream predictive modeling tasks, as it directly impacts the quality and interpretability of the results. GAME embeddings aim to enhance this process by improving the identification and selection of relevant features. To evaluate the effectiveness of GAME embeddings, we collected expert annotations on the relevance of $785$ features, randomly selected across a range of cosine similarity levels, for six diseases: Type 1 Diabetes (T1D), Epilepsy, Pulmonary Hypertension (PH), Asthma, Crohn’s Disease (CD), and Ulcerative Colitis (UC). Each feature is labeled by domain experts as “Not related” (0), “Maybe related” (1), or “Strongly related” (2). We then compute the concordance index (C-index) between cosine similarity scores from different embeddings and these expert labels. A higher C-index indicates better alignment between the embedding and expert-defined feature relevance.

Given the limited availability of expert annotations, we further supplement this evaluation by leveraging GPT-4 to generate feature relevance scores (ranging from 0 to 1) for each disease. For consistency, we focus on the same six diseases as above.
For each disease, we applied all aforementioned embedding methods to identify the top $100$ features with the highest cosine similarity to the disease's PheCode. Additionally, we included $100$ randomly sampled features as negative controls. The union of these selected features formed the feature set for assessment against each disease. For each disease, we computed the cosine similarity between every feature in the feature set and the target PheCode across all embedding methods. We then evaluated the relevance of each feature to the target disease on a scale from $0$ to $1$, as determined by GPT-4,  with the prompt given in Supplementary Section~\ref{sec:fea_result_all}. To compare embedding methods, we assessed the concordance between their cosine similarity scores and GPT-4 relevance ratings using the C-index, treating GPT-4 assessments as a high-quality reference standard.  The validity of using GPT-4 scores as a reference is supported by Supplementary Table~\ref{tab:fea_sel_human}, which shows a strong correlation between GPT-4 scores and human-annotated labels.

Since PPMI-SVD embeddings are only available for features within each institution, we evaluate them using expert-provided labels or available features at each institution corresponding to two methods described above.

\subsection*{Joint patient stratification and risk prediction across institutions}

We used GAME embeddings to support two key analyses: (1) unsupervised clustering to identify patient subgroups with distinct clinical trajectories, and (2) personalized risk prediction via embedding-informed survival models. These analyses were applied to two clinical tasks: progression of Alzheimer’s disease (AD) and suicide risk among patients with mental health disorders.
Prior work on patient clustering has largely focused on single-institution EHRs using aggregated feature counts \cite{doshi2014comorbidity, li2015identification} or embedding-based methods \cite{landi2020deep}. Extending such analyses across institutions is promising for improving generalizability but is hindered by coding heterogeneity. Similarly, traditional risk prediction models often rely on raw EHR features or institution-specific codes, limiting their transportability across diverse health systems.
GAME overcomes these challenges by learning harmonized code embeddings in a shared representation space, enabling joint modeling of patient profiles across institutions. This unified representation allows for seamless integration of patient-level data despite coding differences, facilitating robust clustering, risk stratification, and applications such as identifying “patients like me” in multi-site settings.

\paragraph{Patient embedding for clustering and risk prediction.} For each condition, we defined a baseline period to extract relevant feature counts and computed patient embeddings as the weighted sum of feature embeddings. The weights were determined by multiplying the standard TF-IDF score by the cosine similarity between each feature embedding and the embedding of the target disease's PheCode (290.11 for AD and 297 for suicide), as detailed in Supplementary Section~\ref{supp: patident embedding}. 
To ensure relevance, we included only features with cosine similarity exceeding the $99$th percentile of random pair similarities. Patient embeddings were constructed independently for each institution. 
For clustering, embeddings were first reduced to three dimensions using t-Distributed Stochastic Neighbor Embedding (t-SNE)  approximated via a variational autoencoder \citep{van2008visualizing}. We then applied a federated $k$-means clustering method that iteratively shares local centroids across institutions to enable global clustering while preserving data privacy  \citep{garst2024fed}. Clustering quality was evaluated by examining the association between cluster membership and future clinical outcomes using Cox proportional hazards models. To interpret the clusters, we computed odds ratios and $p$-values for each EHR code. For personalized risk prediction, we projected the patient embeddings into $10$-dimensional Uniform Manifold Approximation and Projection (UMAP)  space and trained Cox models using these embeddings alongside demographic features. We evaluated model performance using the time-dependent AUC for 1-year risk.

\paragraph{AD progression.} Although AD often presents as an amnestic syndrome, clinical progression varies widely \cite{armstrong2022predictors, zheng2024predictors, abdelnour2022perspectives}. Stratifying patients at diagnosis may improve prognostication and management \cite{Wangstrat}. We profiled AD patients at UPMC, MGB, and Duke using EHR data from the two years preceding their first AD diagnosis (time$_0$). The target outcome was nursing home admission, defined as the presence of at least one diagnosis code for admission to a residential facility.
We assessed associations between cluster membership and risk of nursing home admission using Cox models, adjusting for age, sex, and race/ethnicity. We constructed personalized risk prediction models using the projected 10-dimensional GAME embeddings along with demographic features. Risk prediction models were also compared to baselines using (i) demographics only and (ii) demographics plus shared EHR features. Since existing clinical models often rely on sociodemographic data not captured in EHRs \citep{han2023identifying}, we focused on routinely available variables. To assess generalizability, we trained institution-specific models and tested model transportability from UPMC to MGB and Duke.

\paragraph{Suicide risk assessment.} Mental health conditions such as depression, sleep disorders, and anxiety are known risk factors for suicidal ideation and attempts \cite{favril2022risk, sutar2023suicide}. We clustered patients with mental health disorders based on their EHR profiles from the two years following first diagnosis (see Supplementary Table~\ref{tab:mental_phecode}), using data from MGB and Duke. Given that suicide risk varies significantly by age \cite{fazel2020suicide}, we stratified patients into five age groups: $\text{age} < 18$, $18 \le \text{age} \le 25$, $26 \le \text{age} \le 49$, $50 \le \text{age} \le 65$, and $\text{age} > 65$, and performed clustering and analysis separately for each group.
We assessed the relationship between cluster membership and future risk of suicidal ideation or attempts using Cox models adjusted for age, sex, and race/ethnicity.  Personalized risk models were trained using GAME embeddings and demographics, and compared against (i) demographics-only models, (ii) models using demographics plus shared EHR features, and (iii) a validated MGB-developed risk model \citep{sheu2023efficient}, currently under clinical evaluation (NCT05671133). Model transportability was assessed by training at MGB and evaluating at Duke. We conducted stratified analysis of model performance across different age groups as well as across institutions.

 \subsection*{Ablation studies}

To better understand the architectural design and critical components of GAME, we conducted a comprehensive series of ablation studies, as summarized in Table \ref{tab:ablation_studies}. These experiments isolate and evaluate the impact of: (1) Edge configurations, including traditional and GPT-4-enhanced edge types; (2) Contrastive loss construction, with and without hard negative sampling; and (3) Training strategy, comparing one-step and two-step optimization. All experiments used SAPBERT embeddings and aligned PPMI-SVD embeddings as input. Depending on the setting, we applied either one-step training, where all edge types and loss components are optimized jointly, or two-step training, which separately learns similarity embeddings ($256$-D) and relatedness embeddings ($768$-D).
Specifically, studies (a) through (f) progressively incorporate different edge types under a one-step training regime, allowing us to assess the incremental value of adding UMLS edges, traditional PPMI-based edges, GPT-4-enhanced PPMI edges, and GPT-4-based code mapping edges. The main difference between (c) and (d) lies in the construction of PPMI edges for feature selection. In (c), after generating PPMI matrices for each institution, we established edges between codes $i$ and $j$ if $\mathbb{PPMI}_m(i,j)$ exceeded the $99$th percentile of non-zero values for any institution $m$, following the standard thresholding approach in \cite{lee2020harmonized}. The positive set $\mathcal{P}_i$ included codes connected to $i$ via these edges, and the negative set $\mathcal{N}_i$ was constructed by randomly sampling from its complement. In contrast, (d) replaces threshold-based PPMI edges with GPT-4-curated feature selection edges, while retaining the same loss construction and sampling strategies as in GAME. Study (g) focuses on isolating the effect of the negative sampling strategy. While keeping the edge structure and other components identical to the full GAME implementation, we modified the construction of $\mathcal{N}_i$ by replacing hard negatives with randomly sampled codes of the same type and size. The key methodological differences between GAME and (g) are as follows. In the full GAME model, the code mapping loss $\mathcal{L}_{\rm map}^{(i)}$ uses hard negative candidates—standard codes that were not selected by GPT-4. Similarly, the feature selection loss $\mathcal{L}_{\rm fea}^{(i)}$ incorporates negative codes with high PPMI similarity that were excluded by GPT-4. In contrast, (g) replaces both of these curated hard negative sets with random negatives sampled from the complement of $\mathcal{P}_i$, maintaining the same sample size and code types to ensure controlled comparisons. 

To evaluate whether federated alignment is essential, study (h) removes the alignment stage and instead performs centralized pooling of PPMI statistics across all institutions. Specifically, each institutional PPMI matrix is first dimensionally aligned by zero-padding missing codes, after which we compute the element-wise mean across institutions to obtain a pooled PPMI matrix. We then perform SVD to derive pooled PPMI–SVD embeddings, which are used directly in the two-step training without any federated synchronization. This setup thus represents a ``no-FL'' baseline that preserves all other modeling components of GAME.

Furthermore, while GAME integrates both GNN and FL, we also investigate whether GNN alone is sufficient for downstream tasks such as code mapping. To this end, we train GNN models independently within each institution and perform cross-institutional code mapping using task-specific embeddings and GPT-4–generated alignment pairs. The details are presented in Supplementary Section~\ref{sec:code_map_VA} and Supplementary Table~\ref{tab:inst_ablation_code_mapping}.

\subsection*{Statistics \& Reproducibility}

This study is a retrospective observational analysis based on existing EHR data and a public critical care dataset (MIMIC-IV). No statistical method was used to predetermine sample size; sample sizes were determined by the availability of eligible records after applying predefined criteria. No data were excluded from the analyses beyond the predefined inclusion criteria described in the Methods. All data processing, cohort definitions, and analyses are described in the Methods and Supplementary sections. Model performance was summarized using AUC, Spearman correlation, and C-index.

Sex and/or gender information (where available) was obtained from clinical records. Sex and/or gender was not a variable of interest in the study design because the primary evaluations are code- and embedding-level rather than patient-level effects. Accordingly, sex and/or gender was used for covariate adjustment in patient-level models where applicable, but no sex- or gender-stratified or interaction analyses were performed, and sex/gender definitions are not consistently harmonized across institutions.

All data preprocessing, modeling, and statistical analyses were conducted using Python (v3.9.23) and R (v4.4.2). 
Graph-based models were implemented using PyTorch and PyTorch Geometric, and textual embeddings were generated using transformer-based language models implemented with the Hugging Face Transformers library. 
Statistical analyses and evaluation metrics were computed using standard Python and R packages, as detailed in the Reporting Summary.

\subsection*{Data Availability}

Source Data are provided in the source data file. 
The minimum dataset required to interpret and reproduce the main findings of this study consists of the pairwise cosine similarity data derived from the GAME embeddings, which are publicly available at 
\url{https://shiny.parse-health.org/GAME/}.

Institution-level summary data are available under restricted access due to data use agreements (DUAs) with the participating healthcare institutions. 
Access to these data may be obtained by submitting a request and establishing a DUA with the relevant institution(s); 
requests should be directed to the corresponding author.

Individual-level clinical data used for downstream analyses are not publicly available due to patient privacy protections, ethical approval constraints, and institutional regulations. 
These data were accessed under institution-specific DUAs and IRB approvals and cannot be shared beyond those agreements.

Requests for restricted data are reviewed by the relevant institution(s), with an expected response time of approximately 4--8 weeks. 
The duration of access and permitted use are governed by the terms of the corresponding DUA.

\section*{Code Availability}

The code used in this study is publicly available at \url{https://github.com/celehs/GAME}. 
The specific version used in this study has been archived on Zenodo at \url{https://doi.org/10.5281/zenodo.18222787}, ensuring long-term accessibility and reproducibility \cite{GAME2026}. 
An interactive visualization of the knowledge graph derived from the GAME embeddings is available at \url{https://shiny.parse-health.org/GAME/}.

\bibliographystyle{naturemag}
\bibliography{ref}

\section*{Acknowledgements}

This research was supported by the Office of Research and Development, Veterans Health Administration, under award MVP000. This work also used resources of the Knowledge Discovery Infrastructure (KDI) at Oak Ridge National Laboratory, supported by the Office of Science of the U.S. Department of Energy under Contract No.\ DE-AC05-00OR22725. The contents of this publication do not represent the views of the U.S. Department of Veterans Affairs or the United States Government.

D.\ Z.\ was supported by the MOE AcRF Tier 1 Grant A-8003569-00-00 and the NUS Start-up Grant A-0009985-00-00. Z.\ X.\ was supported by NIH grant 5R01NS098023. K.\ L.\ was supported by NIH grants P30 AR072577 and K24 AR085342. T.\ C.\ was supported by NIH grants R01 LM013614, R01 HL089778, P30 AR072577, and P50 MH129699.


\section*{Author Contributions Statement}

D.\ Z., H.\ T., L.\ W., and S.\ L.\ conceived the study.
D.\ Z.\ and H.\ T.\ contributed to the methodology and model design. D.\ Z.\ conceptualized the study and H.T. led the implementation.
H.\ T.\ and L.\ W.\ conducted data analysis and validation experiments.
S.\ L., X.\ X., and Z.\ G.\ contributed to data preprocessing and experimental evaluation.
R.\ G., B.\ H., V.\ J., and R.\ T.\ contributed to the French institutional data and clinical interpretation.
Y.-C.\ L.\ and C.\ H.\ contributed to data extraction and site-specific implementation at Duke University.
C.-L.\ B., V.\ A.\ P., K.\ P., and Z.\ X.\ assisted with data harmonization and result interpretation.
T.\ R.\ C., Y.-L.\ H., L.\ C., J.\ M.\ G., and K.\ C.\ contributed to study design, clinical interpretation, and application development.
K.\ M.\ contributed to informatics design and integration strategy.
K.\ L.\ and T.\ C.\ jointly supervised the study.
All authors contributed to manuscript writing and approved the final version.

\section*{Competing Interests Statement}
The authors declare no competing financial or non-financial interests.

\clearpage

{\centering
\Large
Supplementary Material to\\
Representation Learning to Advance Multi-institutional Studies with Electronic Health Record Data from US and France
\par}

\setcounter{section}{0}
\setcounter{figure}{0}
\setcounter{table}{0}
\setcounter{equation}{0}

\renewcommand{\thesection}{S\arabic{section}}
\renewcommand{\thefigure}{S\arabic{figure}}
\renewcommand{\thetable}{S\arabic{table}}
\renewcommand{\theequation}{S\arabic{equation}}

\makeatletter
\renewcommand{\fnum@figure}{\textbf{Supplementary Figure~\thefigure}}
\renewcommand{\fnum@table}{\textbf{Supplementary Table~\thetable}}
\makeatother

\vspace{3em}

We begin by describing the data in Section~\ref{sec:data_sup}, outlining the construction of the training and validation datasets and summarizing the sources and quantities of edge pairs used in model training. Next, we provide additional methodological details. Section~\ref{supp:ppmi} describes the construction of institutional PPMI-SVD embeddings. Section~\ref{sec:desc_LLM} explains how we refine local code descriptions using various large language models, along with the specific prompts employed. Section~\ref{loss} outlines the loss functions used in our two-step training procedure, with the corresponding algorithms presented. Section~\ref{Tune} discusses the selection of tuning parameters, while Section~\ref{supp:gpu_time} reports training time and computational resources. Section~\ref{supp: upmc bdx mapping} describes how we curated the labels for local code mapping. Finally, Section~\ref{supp: patident embedding} provides implementation details for the joint patient stratification task.
Finally, we present additional validations and results, including ablation studies. Section~\ref{supp:AUC_ALL} shows the results of detecting clinical similarity and relatedness, including institutional AUC comparisons. Section~\ref{sec:code_map_supp} presents the results of local code mapping. Section~\ref{sec:fea_sel_supp} provides feature selection results using both human annotations and GPT-based scores. Section~\ref{supp:patient_stratification} presents the results for joint patient stratification and risk prediction. Section~\ref{sec:PLM} compares PLM fine-tuning and GAME, and lastly, Section~\ref{sec:NLP} shows the results of GAME using NLP-derived Concept Unique Identifiers (CUIs).

\section{Construction of training and validation sets}
\label{sec:data_sup}

We describe the strategy for partitioning similarity and relatedness pairs into training and validation sets. For non-hierarchical similarity and relatedness pairs, we randomly split them with a 7:3 ratio between training and validation sets.

For hierarchical similarity pairs, we split the data at the branch level instead of the pair level to avoid information leakage. Importantly, we maintain the overall 7:3 training-to-validation ratio with respect to the number of pairs, by assigning branches in a way that approximates this ratio. This ensures that all pairs involving codes from the same branch are assigned to the same set.

For example, if the pairs (PheCode 296.22, 296.2) and (296.2, 296.1) are placed in the training set while the pair (296.22, 296.1) is placed in the validation set, then the model could indirectly learn the validation relation from the training data. This would result in unfair information leakage.

To prevent this, we define branches as follows:
\begin{itemize}
    \item PheCode: Codes with the same integer part are grouped into one branch.
    \item RxNorm: Since all RxNorm codes used are leaf nodes, we define a branch as the set of codes sharing the same grandparent.
    \item LOINC: We identify the highest-level parent codes present in our dataset as the original ancestors. All descendant codes of the same ancestor are grouped into a single branch.
    \item CCAM: Codes sharing the first four characters are grouped into the same branch.
\end{itemize}

This branch-based partitioning ensures that structurally related codes do not appear in both the training and validation sets. Returning to the earlier example, PheCodes 296.22, 296.2, and 296.1 belong to the same branch, so all edges between them are assigned to the same split.

Statistics for training set edges and contrastive pairs are reported in Supplementary Table~\ref{edge_count}, and validation set relation pairs are summarized in Supplementary Table~\ref{num_table}, which together illustrate the sparsity of the generated graph. Specifically, prior to incorporating GPT-4-generated edges, the training graph included $69,605$ hierarchical edges, $4,844$ UMLS-derived similar (non-hierarchical) edges, and $6,542$ UMLS-related edges, resulting in a total of $78,991$ curated edges. Given $50,738$ nodes in the graph, this corresponds to only $0.0061\%$ of all possible edges. 
Following GPT-4 augmentation, we added $36,913$ edges from GPT-4–assisted local code mapping and $130,801$ edges from GPT-4–guided PPMI-based feature selection relationships. These additions increased the total number of edges to $246,705$—approximately $0.0191\%$ of all possible edges. Although the graph remains sparse (as is typical for medical knowledge graphs), these enhancements substantially improve connectivity and support more effective representation learning.

\begin{table}[ht]
\centering
\begin{tabular}{c | c c r}
\toprule
{Category} & {Source} & {Detail} & {\# Pairs} \\
\midrule
\multirow{15}{*}{Similar}  & \multirow{5}{*}{Similar Pairs from code hierarchies and UMLS} & LOINC Hierarchy & 20917 \\
 & & PheCode Hierarchy & 3131 \\
 & & RxNorm Hierarchy & 21841 \\
 & & CCAM Hierarchy & 23716 \\
 & & Non-Hierarchical & 4844 \\
\cline{2-4}
 & \multirow{10}{*}{GPT-4 Local-Code mapping} & VA local-lab positive & 12402 \\
 & & VA local-lab negative & 162657 \\
 & & UPMC local-lab positive & 8643 \\
 & & UPMC local-lab negative & 142816 \\
 & & UPMC medication positive & 4578 \\
 & & UPMC medication negative & 129097 \\
 & & UPMC procedure positive & 2430 \\
 & & UPMC procedure negative & 21029 \\
 & & BDX procedure positive & 8860 \\
 & & BDX procedure negative & 242208 \\
\midrule
\multirow{3}{*}{Related} & Related Pairs from UMLS & - & 6542 \\
\cline{2-4}
 & \multirow{2}{*}{GPT-4 PPMI feature selection} & Positive pairs & 130801 \\
 & & Negative pairs & 467580 \\
\bottomrule
\end{tabular}
\caption{Summary of edges and pairs in the training set. All relationships, except negative pairs, are used as edges; positive and negative pairs are utilized in the contrastive loss function.}
\label{edge_count}
\end{table}


\begin{table}[ht]
\centering
\setlength{\tabcolsep}{2pt}
\begin{tabular}{c|c|ccccccc|c}
\toprule
\multirow{2}{*}{{Type}} & \multirow{2}{*}{{Source}} & \multicolumn{7}{c|}{{PPMI-SVD}} & \multirow{2}{*}{{PLM}}\\
 & & {BCH} & {BDX} & {Duke} & {MGB} & {MIMIC} & {UPMC} & {VA} &  \\
\midrule
\multirow{9}{*}{{Similar}}  
 & LOINC Hierarchy & 655 & 975 & 55 & 1062 &  & 2625 & 313 & 6192\\
 & RxNorm Hierarchy & 403 & 383 & 91 & 503 & 458 & 1469 & 489 &4753\\
 & CCAM Hierarchy & & 4301 & & & & & &4301\\
 & PheCode Hierarchy & 603 & 1072 & 618 & 1199 & 221 & 1258 & 1245 & 1300\\
 & Is a & 548 & 629 & 542 & 686 & 111 & 773 & 702 & 856\\
 & CHD & 444 & 520 & 418 & 574 & 71 & 617 & 571 & 677\\
 & classifies & 108 & 126 & 109 & 115 & 20 & 134 & 131 & 137\\
 & part of &  &  &  &  &  & 88 &  & 123\\
 & mapped to & 77 & 91 & 76 & 95 & 19 & 102 & 98 & 104\\
\midrule
\multirow{16}{*}{{Related}} 
 & may treat & 365 & 477 & 26 & 556 &  & 627 & 516 & 755\\
 & clinically associated with & 305 & 396 & 307 & 457 & 270 & 478 & 453 & 482\\
 & has contraindicated drug & 217 & 249 & 10 & 335 &  & 365 & 283 & 449\\
 & manifestation of & 175 & 222 & 181 & 236 & 10 & 238 & 222 & 243\\
 & RO & 103 & 133 & 98 & 151 & 18 & 171 & 149 & 198\\
 & may prevent & 85 & 102 &  & 116 &  & 124 & 107 & 149\\
 & co-occurs with & 78 & 82 & 67 & 114 & 41 & 117 & 113 & 118\\
 & associated morphology of & 66 & 72 & 66 & 74 & 10 & 87 & 85 &87\\
 & RB & 48 & 58 & 41 & 57 & 11 & 61 & 62 & 81\\
 & associated with & 21 & 29 & 18 & 38 &  & 38 & 29 & 44\\
 & active ingredient of &  &  &  &  &  & 14 &  & 40\\
 & ssc & 31 & 28 & 31 & 34 & 25 & 36 & 36 & 36\\
 & related to & 15 & 18 & 15 & 20 & 18 & 20 & 20 & 20\\
 & disease may have finding & 13 & 17 & 13 & 18 & 18 & 18 & 18 & 18\\
 & RQ & 14 & 12 & 11 & 16 & 15 & 16 & 15 & 17\\
 & causative agent of & 13 & 13 &  & 11 &  & 13 &  & 16\\
\bottomrule
\end{tabular}
\caption{Number of relation pairs used to evaluate each source in detecting known similar and related relationships. {PLM} refers to non-PPMI-SVD embeddings, as described in Supplementary Table~\ref{R_sim}. All relation pairs used for evaluating PLM embeddings are shared across institutions. In contrast, for {PPMI-SVD} embeddings, only the relations present within each respective institution are used for evaluation. Relation type abbreviations: {CHD}: ``child of''; {RO}: ``has relationship other than synonymous, narrower, or broader''; {RB}: ``has a broader relationship''; {RQ}: ``related and possibly synonymous.''}
\label{num_table}
\end{table}

\section{Supplementary methodological details}

\subsection{Construction of PPMI-SVD embeddings}

\label{supp:ppmi}
\begin{figure}[ht]
    \centering
    \includegraphics[width=0.92\textwidth]{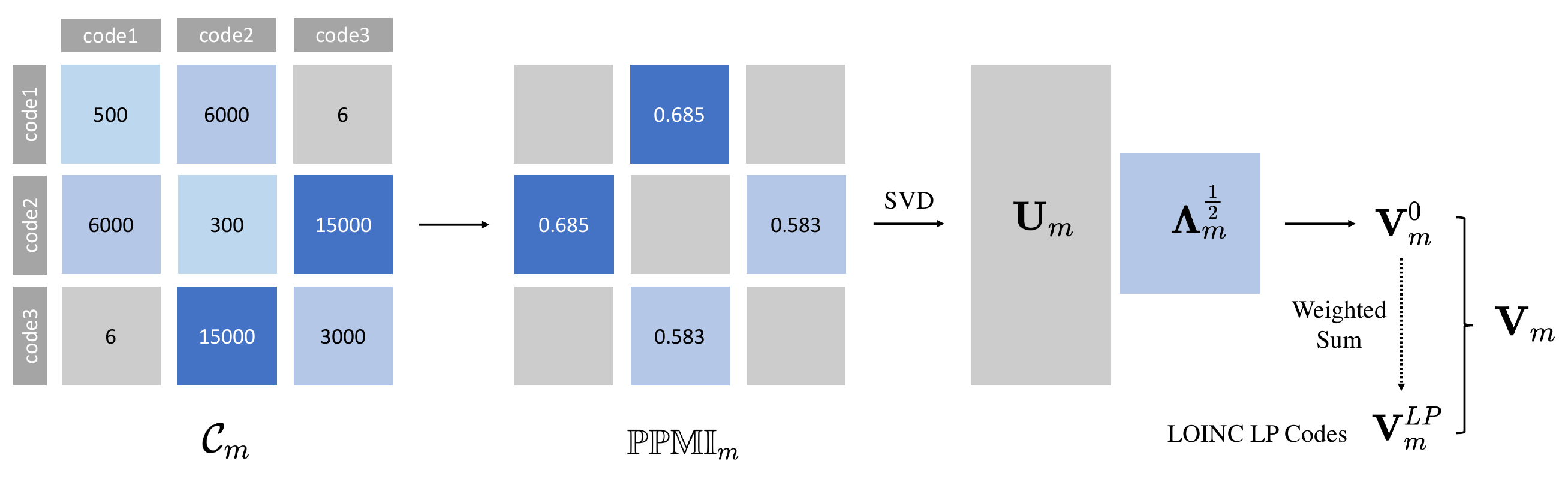}
    \caption{Generation of institutional PPMI-SVD embeddings.}
    \label{fig:PPMI}
\end{figure}

We construct institutional PPMI-SVD embeddings by first computing the Positive Pointwise Mutual Information (PPMI) matrix for each institution, as outlined in the \emph{Methods} section of the main text  and illustrated in Supplementary Fig.~\ref{fig:PPMI}. Specifically, for the $m$th institution, we compute:
\begin{equation}
\mathbb{PPMI}_m(i,j) = \max\left\{0, \log\frac{\Csc_m(i,j)\Csc_m(\cdot,\cdot)}{\Csc_m(i,\cdot) \Csc_m(j,\cdot)}\right\},
\label{PPMI_1}
\end{equation}
where $\Csc_m(i,\cdot)$ represents the sum of all occurrences for codes $i$, and $\Csc_m(\cdot,\cdot)$ is the total sum of the co-occurrence. We then apply SVD to the PPMI matrix as $\PPMI_m = \U_m\rm{diag}(\Lambda_{m,1}, ...,\Lambda_{m,n_m})\U_m\trans$ and generate the embeddings
\begin{equation}
    \V_{m}^0 = \U^{(d)}_{m} \rm{diag} \left(\Lambda_{m,1}\suphalf, ..., \Lambda_{m,d}\suphalf \right)\,,
    \label{PPMI_2}
\end{equation}
where $\U^{(d)}_{m}$ represents the first $d$ singular vectors with positive eigenvalues.

The initial co-occurrence matrix does not include LP codes. To incorporate these, we compute LP code embeddings as a weighted average of their child embeddings, using the occurrence frequency as weights. We then concatenate the base and LP embeddings:
\begin{equation}
    \V_{m} = \Rscr\left( \begin{bmatrix} \V_{m}^0 \\ \mathbf{V}_{m}^{LP} \end{bmatrix} \right),
    \label{PPMI_5}
\end{equation}
where $\Rscr(\cdot)$ denotes row-wise $\ell_2$ normalization that standardizes each row of the matrix to unit length. 

Let $\Isc_m$ represent the set of EHR codes (both base and LP codes) for the $m$th institution and $\Isc = \cup_{m} \Isc_m$ denote the unified code set across all institutions, with $N = |\Isc|$ unique codes. To handle codes missing in a single institution, we pad the corresponding entries in each institution's embedding matrix with zeros, ensuring that $\mathbf{V}_{m} \in \mathbb{R}^{N \times d}$ is consistent across institutions, with each row representing the same EHR code.

\subsection{Evaluation of LLM-generated descriptions}
\label{sec:desc_LLM}

As part of the preprocessing step described in in the \emph{Methods} section of the main text, we used GPT-4 to expand abbreviated descriptions of VA local codes, following an initial dictionary-based mapping from laboratory test acronyms to their full names. To evaluate whether GAME performance is robust to the method used for expanding abbreviations, we compared GPT-4 with two additional large language models (LLMs): Claude-3~\citep{anthropic2024claude3} and LLaMA-4~\citep{meta2025llama4}. Each model was used to generate full code descriptions, and the similarity of their outputs was quantified using the cosine similarity of SAPBERT, CODER, and BGE embeddings. Since these embeddings are used in GAME, GAME (CODER), and GAME (BGE), high cosine similarity across the LLM-generated descriptions—measured by these embeddings—indicates that the input embeddings remain consistent, thereby supporting the robustness of our method.

\begin{table}[ht]
\centering
\begin{tabular}{c|c|ccc}
\toprule
{Embedding} & {Model} & {GPT-4} & {Claude-3} & {LLaMA-4} \\
\midrule
\multirow{3}{*}{{SAPBERT}} 
& GPT-4     & 1.000 & 0.953 & 0.950 \\ 
& Claude-3  & 0.953 & 1.000 & 0.934 \\ 
& LLaMA-4   & 0.950 & 0.934 & 1.000 \\ 
\midrule
\multirow{3}{*}{{CODER}} 
& GPT-4     & 1.000 & 0.914 & 0.909 \\ 
& Claude-3  & 0.914 & 1.000 & 0.907 \\ 
& LLaMA-4   & 0.909 & 0.907 & 1.000 \\ 
\midrule
\multirow{3}{*}{{BGE}} 
& GPT-4     & 1.000 & 0.928 & 0.926 \\ 
& Claude-3  & 0.928 & 1.000 & 0.918 \\ 
& LLaMA-4   & 0.926 & 0.918 & 1.000 \\ 
\bottomrule
\end{tabular}
\caption{Average pairwise cosine similarity scores between LLM-generated local code descriptions, computed using SAPBERT, CODER, and BGE embeddings.}
\label{tab:llm_similarity}
\end{table}

The pairwise similarity scores are reported in Supplementary Table~\ref{tab:llm_similarity}. Each entry reflects the average cosine similarity across all local codes for a given model pair. The results indicate a high degree of consistency among the models, with all pairwise similarity scores exceeding $0.909$. This suggests that GPT-4’s descriptions are representative of those produced by other state-of-the-art LLMs. The prompts used to generate the code descriptions are summarized in Supplementary Table~\ref{tab:desc_prompt}, with placeholders dynamically replaced by each code’s description and unit at runtime.

\begin{table}[H]
\centering
\begin{tabular}{p{3cm} | p{11cm}}
\toprule
{Prompt Type} & {Text} \\
\midrule
System Prompt & 
As a supportive assistant to a skilled biostatistician, your task is to infer the complete names of various laboratory test labels from their abbreviated descriptions. You should avoid returning ``NA'' unless no reasonable alternative exists. The biostatistician values comprehensive and detailed names for laboratory tests. Please disregard any time-related information not directly associated with the lab test label itself, such as content in parentheses (e.g., ``(dc 1–12)''). \\
\midrule
User Prompt & 
What is the complete name of \texttt{[desc]} in laboratory tests, given that its unit is \texttt{[unit]}? You may disregard any information within parentheses. Please respond in the format: \texttt{[full name]}; \texttt{[reasons]}. Avoid responding with ``NA'' unless no plausible alternative exists. \\
\bottomrule
\end{tabular}
\caption{System and user prompts used for LLM-generated local code descriptions. Placeholders \texttt{[desc]} and \texttt{[unit]} were replaced with actual code values.}
\label{tab:desc_prompt}
\end{table}

\subsection{Detailed algorithms and loss functions}
\label{loss}

We first present the detailed algorithm (Algorithm~\ref{alg:align_sppmi}) for aligning PPMI-SVD embeddings across institutions using GAT, as introduced in the \emph{Methods} section of the main text.

\begin{algorithm}[H]
\DontPrintSemicolon
  \KwInput{Set of institutions $\Isc = \cup_{m=1}^M \Isc_m$, with union of all $|\mathcal{I}| = N$ EHR codes, institutional co-occurrence matrix $\mathcal{C}_m$ for each institution  $m$, edge set $\mathcal{E}$, $\textit{InitLoss} = \infty$, and maximum epochs $T$}
  \KwOutput{Aligned PPMI-SVD embedding $\mathbf{Y}$ across different institutions}
  \ForEach{$m \in \{1,2,\dots,M\}$}{
   \tcc{Build initial PPMI-SVD embedding for each institution}
      Compute $\mathbb{PPMI}^{(m)}$ using \eqref{PPMI_1}
      
      Obtain $\mathbf{V}^{0}_m$ using the PPMI-SVD algorithm in \eqref{PPMI_2} 
      
      Pad the LOINC LP code rows in $\mathbf{V}^{LP}_m$ in institution $m$; normalize rows and pad zeros for missing codes to obtain $\mathbf{V}_m$ using \eqref{PPMI_5}
  }
  
  \For{$\textit{epoch} = 1$ \KwTo $T$}{
    \tcc{Train GAT using initial embeddings, edges, and alignment loss}

    \ForEach{$m \in \{1,2,\dots,M\}$}{
        Use $\mathbf{V}_m$ for $m$th institution and all edges in $\mathcal{E}$ to compute $\mathbf{Y}_m$ using \eqref{Align}
    }
    
    Compute $\textit{Loss}$ using the loss function defined in \eqref{align_loss}.
    
    \If{$\textit{Loss} < \textit{InitLoss}$}{
        $\textit{InitLoss} = \textit{Loss}$ 
        
        \textit{epoch} $+= 1$
    }
    \Else{
        \text{break}
    }
    Apply transformation in \eqref{Align} to compute $\mathbf{Y}$
  }
  \Return{$\mathbf{Y}$} \tcc{Return the aligned PPMI-SVD embedding}
\caption{Aligning institutional PPMI-SVD embeddings.}
\label{alg:align_sppmi}
\end{algorithm}

We next detail the loss functions. In the \emph{Methods} section of the main text, we introduced the various loss components used in the similarity and relatedness training stages. These losses are defined based on distinct sets of positive and negative relations. Here, we provide a more detailed explanation of how these sets are constructed.

In the similarity training step, we apply the similarity-based loss \eqref{sim_loss} to train $\Z_{\mathcal{S}}$. In the second step, the relatedness-based loss \eqref{rel_loss} is used to optimize $\Z_{\mathcal{R}}$ to capture more complex associations.
\begin{align}
    \mathcal{L}_{\mathcal{S}} &= c_{\rm sim,h}\sum_{i} \mathcal{L}^{(i)}_{\rm sim,h} 
    + c_{\rm sim,nh}\sum_{i} \mathcal{L}^{(i)}_{\rm sim,nh} 
    + c_{\rm map} \sum_{i}\mathcal{L}^{(i)}_{\rm map}, 
    \label{sim_loss} \\
    \mathcal{L}_{\mathcal{R}} &=  c_{\rm rel}\sum_{i}\mathcal{L}^{(i)}_{\rm rel} +  c_{\rm fea}\sum_{i}\mathcal{L}_{\rm fea} ^{(i)}.
    \label{rel_loss} 
\end{align}

Here, $\mathcal{L}^{(i)}_{\rm sim,h}$, $\mathcal{L}^{(i)}_{\rm sim,nh}$, $\mathcal{L}^{(i)}_{\rm map}$, $\mathcal{L}^{(i)}_{\rm rel}$, and $\mathcal{L}^{(i)}_{\rm fea}$ denote the loss terms for hierarchical similarity pairs, non-hierarchical similarity pairs, local code mapping, relatedness, and feature selection, respectively, for code $i$.

As a complement to the \emph{Curation of the adjacency matrix} section in the \emph{Methods} of the main text, we define the positive set \(\mathcal{P}_i\) and the negative set \(\mathcal{N}_i\) for each loss as follows:
\begin{itemize}
    \item For $\mathcal{L}^{(i)}_{\rm sim,h}$, $\mathcal{P}_i$ consists of sibling codes of $i$, while $\mathcal{N}_i$ includes cousin codes sharing the same grandparent but having different parent codes.

    \item For $\mathcal{L}^{(i)}_{\rm sim,nh}$ and $\mathcal{L}^{(i)}_{\rm rel}$, $\mathcal{P}_i$ includes non-hierarchically similar or related codes according to UMLS, and $\mathcal{N}_i$ includes randomly sampled codes of the same type.

    \item For $\mathcal{L}^{(i)}_{\rm map}$, $\mathcal{P}_i$ refers to standard codes mapped to a local code $i$ by GPT-4; $\mathcal{N}_i$ consists of alternative candidates not selected.

    \item  For $\mathcal{L}^{(i)}_{\rm fea}$, $\mathcal{P}_i$ and $\mathcal{N}_i$ denote the feature selection positive and negative sets.
\end{itemize}

Although the definition of \(\mathcal{P}_i\) and \(\mathcal{N}_i\) varies across loss types, they consistently separate semantically relevant from irrelevant codes. This design improves the GAT's ability to capture nuanced relationships between medical codes.

To ensure that each part of the embedding focuses on specific knowledge, we update $\Z_{\mathcal{S}}$ using the similarity loss $\mathcal{L}_{\mathcal{S}}$ in \eqref{sim_loss}, and $\Z_{\mathcal{R}}$ using the relatedness loss $\mathcal{L}_{\mathcal{R}}$ in \eqref{rel_loss}. In the second step, we compute the loss using the combined embedding $\mathbf{Z} = [\mathbf{Z}_\mathcal{S}, \mathbf{Z}_\mathcal{R}]$. Although $\Z_{\mathcal{S}}$ remains fixed during the second step, this approach enhances the integration between $\Z_{\mathcal{S}}$ and $\Z_{\mathcal{R}}$. The complete training procedure is summarized in Algorithm~\ref{alg:encoder}.

\begin{algorithm}[H]
\DontPrintSemicolon

   \KwInput{Set of institutions $\Isc = \cup_{m=1}^M \Isc_m$, with union of EHR codes  $|\mathcal{I}| = N$, institutional co-occurrence matrix $\mathcal{C}_m$ for each institution $m$, edge set $\mathcal{E}$, descriptions of medical codes $\mathcal{D}$, $\textit{InitAcc} = 0$, $\textit{InitCorr} = -\infty$, and maximum epochs $T$}
   
  \KwOutput{Final embedding ${\Z}$ across all institutions}

  Use Algorithm~\ref{alg:align_sppmi} to obtain the aligned PPMI-SVD embedding $\Y$

  Use descriptions of medical codes $\mathcal{D}$ to acquire the SAPBERT embedding $\X$
  
  \For{$\textit{epoch} = 1$ \KwTo $T$}{
    \tcc{Train the Graph Attention Network using $[\X,\Y]$, edges, and the loss function (Similarity step).}
    
    Use $[\X,\Y]$ as the initial input and all edges in $\mathcal{E}$ to compute $\Z_{\mathcal{S}}^{epoch}$ using  \eqref{SIMI}

    Compute $\textit{Loss}^{epoch}$ using the embedding $\Z_{\mathcal{S}}^{epoch}$ defined in  \eqref{M_S} and \eqref{sim_loss}

    Compute accuracy $\textit{Acc}$ of Code Mapping Using $\Z_{\mathcal{S}}^{epoch}$ 

    \If{$\textit{Acc} > \textit{InitAcc}$}{
        $\textit{InitAcc} = \textit{Acc}$ 
        
        $\Z_{\mathcal{S}} = \Z_{\mathcal{S}}^{epoch}$ 
    }
    }
    \tcc{Output the similarity part of the embedding $ \Z_{\mathcal{S}}$.}

    \For{$\textit{epoch} = 1$ \KwTo $T$}{
        \tcc{Train the Graph Attention Network using $[\X,\Y]$, edges, and the loss function (Relatedness step).}
        
        Use $[\X,\Y]$ as the initial input and all edges in $\mathcal{E}$ to compute $\Z_{\mathcal{R}}^{epoch}$ using \eqref{RELA}
    
        Compute $\textit{Loss}^{epoch}$ defined in \eqref{M_S} and \eqref{rel_loss} using the embedding $[\Z_{\mathcal{S}}, \Z_{\mathcal{R}}^{epoch}]$

        Compute average correlation $\textit{Corr}$ of Feature Selection Using $\Z_{\mathcal{R}}^{epoch}$ 
        
        \If{$\textit{Corr} > \textit{InitCorr}$}{
            $\textit{InitCorr} = \textit{Corr}$ 
    
             $\Z_{\mathcal{R}} = \Z_{\mathcal{R}}^{epoch}$ 
        }
    }
    \tcc{Output the relatedness part of the embedding $\Z_{\mathcal{R}}$.}
    
    Obtain the complete embedding $\Z=[\Z_{\mathcal{S}}, \Z_{\mathcal{R}}]$.
    
  \Return{${\Z}$} 
  \tcc{Return the final embedding.}
\caption{Two-step training procedure.}
\label{alg:encoder}
\end{algorithm}

\subsection{Tuning parameters}
\label{Tune}

In our training process, we use the Stochastic Gradient Descent (SGD) optimizer, setting the learning rate to \(1 \times 10^{-4}\) for the aligned PPMI-SVD embedding step (Algorithm~\ref{alg:align_sppmi}) and \(1 \times 10^{-6}\) for the two-step training procedure (Algorithm~\ref{alg:encoder}). An exponential decay scheduler is applied with a decay rate of \(0.99\) and a minimum learning rate of \(5 \times 10^{-7}\). A drop edge rate of 0.5 is used in every step of GAT training.

The aligned PPMI-SVD training is terminated when the loss begins to increase, as described in Algorithm~\ref{alg:align_sppmi}. In the similarity training step, we retain the embedding with the highest code mapping accuracy; in the relatedness training step, we retain the embedding with the highest feature selection correlation (Algorithm~\ref{alg:encoder}).

When splitting the training and validation sets, we divide the hierarchical similarity pairs according to their branches, as described in Section~\ref{sec:data_sup}, using a 7:3 ratio. For non-hierarchical pairs (which include part of the similar pairs and all of the related pairs), we perform a random split with the same 7:3 ratio. These pairs are used in both edge construction and loss computation, as further detailed in the \emph{Methods} section of the main text .

For the loss weights in \eqref{sim_loss} and \eqref{rel_loss}, we set \(c_{\rm sim,h} = 1\), \(c_{\rm sim,nh} = 1\), \(c_{\rm map} = 30\), \(c_{\rm rel} = 5\), and \(c_{\rm fea} = 0.1\). For the contrastive loss in \eqref{M_S}, we use \(\alpha = 1\), \(\beta = 5\), and \(\lambda = 0.5\).

\subsection{Computational cost and efficiency}
\label{supp:gpu_time}

All training was performed on a single NVIDIA A100 GPU. The time consumed by each training stage { of GAME} is summarized in Supplementary Table~\ref{tab:gpu_time}. This computational cost is considered reasonable given the scale and complexity of the task. 
Compared to many large-scale biomedical language models, the training time of GAME is highly efficient. For example, BioBERT was pre-trained for $23$ days on $8$ NVIDIA V100 GPUs (approximately $4,416$ GPU-hours), and PubMedBERT was trained for about $5$ days on $16$ V100 GPUs (approximately $1,920$ GPU-hours). While SAPBERT used smaller GPUs (4 × RTX 2080 Ti), its training duration was not reported. In contrast, GAME completes all training stages in just { $18$ GPU-hours on a single NVIDIA L40s GPU}, offering a substantially more efficient solution for harmonizing over $50,000$ codes across seven institutions.
\begin{table}[H]
\centering
\begin{tabular}{lc}
\toprule
Training Stage & GPU Hours \\
\midrule
Aligning institutional PPMI-SVD embeddings & $0.1$ \\
Learning the similarity embedding $\Z_{\mathcal{S}}$ & $8.6$ \\
Learning the relatedness embedding $\Z_{\mathcal{R}}$ & $9.5$ \\
\midrule
Total & {$18.2$} \\
\bottomrule
\end{tabular}
\caption{GPU hours used per training stage.}
\label{tab:gpu_time}
\end{table}

{ In our GAME training, we performed $80$ steps for similarity learning and $40$ steps for relatedness learning according to our stopping criteria. To ensure a fair comparison of computational efficiency across all ablation studies, we trained all baselines for $120$ steps. 

As shown in Supplementary Table~\ref{tab:all_time}, the Ablation Study (e) (see definitions in the \emph{Ablation studies} section in the main text—where similarity and relatedness losses are optimized jointly within a single GAT rather than sequentially—requires approximately $48$ GPU-hours, which is significantly slower than GAME’s $18$ GPU-hours. This slowdown likely occurs because, in the two-step training scheme, we perform backward propagation with respect to only one loss (similarity or relatedness) at a time, whereas in the joint one-step training, both losses are computed and differentiated simultaneously, increasing computational load. If we use the relatedness AUC as the stopping criterion for training (e), the optimal stopping point occurs at around $60$ epochs, corresponding to approximately $24$ GPU-hours, which is still longer than GAME’s two-step training. In contrast, Ablation Studies (g) and (h), which also adopt the two-step design, exhibit training times comparable to GAME, further confirming the efficiency of the two-step approach.

Comparing other baselines in Supplementary Table~\ref{tab:all_time}, we observe that as the number of edges and losses increases, the total training time also rises. Notably, baseline (c), which adds raw PPMI edges without GPT refinement, contains a larger number of edges and losses than other settings. This leads to longer training time and inferior performance due to the additional noise introduced by unrefined edges.

We also report the typical GPU memory usage during training. GAME typically consumes about $6$~GB of GPU memory in both the similarity and relatedness phases, which is quite affordable considering that the model is trained on $50{,}000$ codes across $7$ institutions. Across ablation settings, models that incorporate more edges or losses tend to use slightly more memory, but the variation is minor. All models have peak memory usage around $9$–$10$~GB, remaining within a practical range.

In conclusion, GAME achieves strong computational efficiency primarily due to its two-step training design and the careful construction of edge sets and loss components.}

\begin{table}[H]
\centering
\begin{tabular}{cc|ccc}
\toprule
\multicolumn{2}{c|}{Method} & Edge Count & Typical Memory (GB) & GPU Hours (One/Two-step) \\  
\midrule
\multicolumn{2}{c|}{GAME} & 230,993 & 5.9 / 6.6 & 18.1 (8.6 + 9.5) \\ 
\hline
\multirow{8}{*}{Ablation} 
 & (a) & 69,605  & 4.8 & 9.9 \\ 
 & (b) & 80,991  & 5.1 & 18.0 \\ 
 & (c) & 260,002 & 6.8 & 51.8 \\
 & (d) & 194,473 & 6.2 & 44.0 \\
 & (e) & 230,993 & 6.3 & 48.2 \\ 
 & (f) & 157,106 & 5.7 & 30.3 \\ 
 & (g) & 230,993 & 5.9 / 6.6 & 18.3 (8.7 + 9.6) \\ 
 & (h) & 230,993 & 6.0 / 6.6 & 18.2 (8.6 + 9.6) \\ 
\bottomrule
\end{tabular}
\caption{Training time and memory usage for GAME and ablation baselines. Typical Memory reports average GPU usage during similarity / relatedness phases.}
\label{tab:all_time}
\end{table}

\subsection{Manual label curation}
\label{supp: upmc bdx mapping}

Since UPMC local laboratory codes and both UPMC and BDX local procedure codes lacked predefined standard mappings, we manually curated a subset of mappings to evaluate model performance. Specifically, we focused on mapping UPMC local lab codes to LOINC and UPMC and BDX local procedure codes to CCS.

To construct the candidate mapping set, we used embeddings from  SAPBERT, CODER, BGE, and OpenAI. For each local code, we selected the top $20$ most similar standard codes (either LOINC or CCS) based on cosine similarity. To ensure a balanced evaluation, we also included randomly sampled negative codes—codes unlikely to be true matches—as controls. 

 We developed a custom web-based  annotation tool (\href{https://celehs.connect.hms.harvard.edu/GAME_QC/}{https://celehs.connect.hms.harvard.edu/GAME\_QC/}) to support the curation process. The interface guided annotators through the following steps:
\begin{enumerate}
    \item Select a Target Standard Code – either a CCS (for procedure mappings) or LOINC (for lab mappings).
    \item Review Candidate Local Codes – a randomized list of potential local code matches, including both highly ranked candidates and negative controls, was presented.
    \item Assign Mapping Likelihood – annotators were asked to classify each local code into one of three categories: a) should not be mapped to the target code with a numeric score of 0; b) can probably be mapped with a numeric score of 0.5; and c) should be mapped with a score of 1. 
\end{enumerate}
Final scores for each candidate were computed by averaging across annotators, who were clinicians based in both the United States and India. This curation process resulted in a total of  $199$ UPMC procedure code mappings to CCS, $537$ BDX (CCAM) procedure mappings to CCS, and $1,814$ UPMC lab code mappings to LOINC.

For the six diseases evaluated in the \emph{Feature selection} section of the main text, the same system was employed by seven clinicians to rate the relevance of various clinical features for each disease. Each feature was annotated as ``Strongly related'', ``Maybe related'', or ``Not related'', based on clinical judgment.

\subsection{Details of patient stratification procedure}
\label{supp: patident embedding}

We provide additional details for the procedure described in the \emph{Methods} section of the main text. The joint patient stratification consists of four steps: (1) feature selection, (2) generation of patient embeddings, (3) dimension reduction, and (4) federated clustering. This section outlines the detailed implementation of each step.

\paragraph{Feature selection}
To ensure the relevance of clustering, we included only features with high relevance to the disease or outcome of interest, based on the trained GAME embeddings. Specifically, we randomly sampled pairs of codes and computed the cosine similarity between their embeddings. We then set the $99$th percentile of these similarities as a threshold and selected features whose cosine similarity with the target codes (PheCode 290.11 for Alzheimer's disease and PheCode 297 for suicide) exceeded this value.

\paragraph{Generation of patient embeddings}
We computed the embedding for the $i$th patient using the following weighted TF-IDF procedure:
$$
\W_i^{[j]} = \sum_{c \in \mathcal{V}_{\text{tar}}^{[j]}} \cos(\widehat\Z_{\text{tar}}, \widehat\Z_c) \cdot \frac{\log(a_{ic}^{[j]}+1)}{\log(b_c^{[j]}+1)} \cdot \widehat\Z_c,
$$
where $\mathcal{V}_{\text{tar}}^{[j]}$ is the selected feature set present in the EHR system of institution $j$, $a_{ic}^{[j]}$ is the count of feature $c$ for patient $i$, and $b_c^{[j]}$ is the occurrence of feature $c$ in all patients from institution $j$.

\paragraph{Encoder for dimension reduction of patient embeddings}
We used t-SNE to reduce the dimensionality of patient embeddings at the central site (UPMC for Alzheimer's disease; MGB for mental health disorders). To learn a mapping from the high-dimensional embeddings to the corresponding t-SNE representation with $3$ dimensions, we trained a fully connected neural network with three hidden layers ($30$, $20$, and $10$ nodes). To prevent overfitting, a dropout layer with a rate of $0.2$ was added. The network was trained using the Adam optimizer with a learning rate of $1 \times 10^{-4}$.

\paragraph{Federated clustering}
We adopted the federated $k$-means algorithm proposed by \cite{garst2024fed}. After each institution performs one iteration of $k$-means clustering on its local data, it sends the resulting cluster means and corresponding sample counts to the central server. The server then aggregates the received means by running global $k$-means clustering on the combined local centroids until convergence, using the global $k$ parameter. The global aggregation step is weighted by the number of samples in each cluster, so that larger clusters contribute more strongly to the final cluster centers. These steps are repeated until convergence.

\section{Supplementary analyses and results}
\label{sec:S2}

This section presents extended analyses and validation results that support the main findings of the paper. We report additional evaluations on similarity and relatedness detection, cross-institutional code mapping, feature selection, and patient stratification. For completeness, we also include ablation results (expanding in the \emph{Ablation studies} section of the main text) and comparative performance across multiple baseline methods.

As discussed in the \emph{Discussion} section of the main text, to further assess the generalizability of the GAME algorithm, we substitute SAPBERT with BGE (768-dimensional) and CODER (768-dimensional) as the input $\mathbf{X}$ in Algorithm~\ref{alg:encoder} (line 2), keeping all other components unchanged. The resulting variants, referred to as GAME (BGE) and GAME (CODER), are evaluated using the same downstream tasks.

\subsection{Evaluation of similarity and relatedness detection}
\label{supp:AUC_ALL}

We report the evaluation results for detecting clinical similarity and relatedness relationships using various embedding methods. The overall performance across all relation pairs is summarized in Section~\ref{sec:AUC_result}, while institution-specific analyses are provided in Section~\ref{result_AUC_insti}.

\subsubsection{Detecting general similarity and relatedness relationships}
\label{sec:AUC_result}

\begin{table}[H]
\centering
\begin{tabular}{cc|cccc}
\toprule
\multicolumn{2}{c|}{Method} & Similarity & Similarity CI & Relatedness & Relatedness CI \\ \midrule
\multirow{7}{*}{PPMI-SVD} 
 & BCH & 0.830 & [0.817, 0.844] & 0.758 & [0.738, 0.775] \\ 
 & BDX & 0.873 & [0.861, 0.884] & 0.825 & [0.816, 0.848] \\ 
 & Duke & 0.828 & [0.809, 0.844] & 0.795 & [0.773, 0.813] \\
 & MGB & 0.941 & [0.937, 0.952] & 0.837 & [0.819, 0.844] \\ 
 & MIMIC & 0.927 & [0.914, 0.947] & 0.938 & [0.916, 0.956] \\ 
 & UPMC & 0.876 & [0.865, 0.884] & 0.756 & [0.736, 0.768] \\ 
 & VA & 0.926 & [0.917, 0.934] & 0.793 & [0.776, 0.806] \\
\multicolumn{2}{c|}{AVE} & 0.886 & [0.870, 0.902] & 0.815 & [0.793, 0.836] \\ \hline
\multicolumn{2}{c|}{BBERT} & 0.681 & [0.663, 0.686] & 0.624 & [0.609, 0.631] \\ 
\multicolumn{2}{c|}{PBERT} & 0.632 & [0.621, 0.644] & 0.602 & [0.599, 0.620] \\ 
\multicolumn{2}{c|}{SBERT} & 0.803 & [0.799, 0.817] & 0.753 & [0.745, 0.764] \\ 
\multicolumn{2}{c|}{CODER} & 0.876 & [0.867, 0.883] & 0.655 & [0.646, 0.667] \\  
\multicolumn{2}{c|}{BGE} & 0.925 & [0.916, 0.928] & 0.843 & [0.824, 0.846] \\ 
\multicolumn{2}{c|}{OpenAI} & \textbf{0.936} & [0.934, 0.943] & 0.804 & [0.803, 0.827] \\ \hline
\multicolumn{2}{c|}{GAME} & 0.916 & [0.910, 0.922] & 0.914 & [0.911, 0.923] \\ 
\multicolumn{2}{c|}{GAME (BGE)} & 0.921 & [0.916, 0.927] & 0.938 & [0.927, 0.937] \\ 
\multicolumn{2}{c|}{GAME (CODER)} & 0.931 & [0.921, 0.930] & 0.931 & [0.924, 0.935] \\ \hline
\multirow{7}{*}{Ablation} 
 & {(a)} & 0.862 & [0.854, 0.873] & 0.827 & [0.814, 0.832] \\ 
 & {(b)} & 0.875 & [0.865, 0.879] & \textbf{0.940} & [0.933, 0.943] \\ 
 & {(c)} & 0.838 & [0.828, 0.844] & 0.892 & [0.879, 0.893] \\
 & {(d)} & 0.854 & [0.849, 0.866] & 0.935 & [0.927, 0.937] \\
 & {(e)} & 0.883 & [0.876, 0.888] & 0.917 & [0.908, 0.920] \\ 
 & {(f)} & 0.882 & [0.871, 0.886] & 0.786 & [0.775, 0.795] \\ 
 & {(g)} & 0.868 & [0.863, 0.877] & 0.906 & [0.896, 0.907] \\ 
 & {(h)} & 0.817 & [0.812, 0.829] & 0.935 & [0.928, 0.939] \\ \bottomrule
\end{tabular}
\begin{flushleft}
\footnotesize
\textbf{Bold} indicates the best performance within each column.
\end{flushleft}
\caption{AUC for detecting similarity and relatedness relationships using various methods. AVE represents the average PPMI AUC across different institutions. The $95\%$ confidence intervals (CIs) are estimated via bootstrapping. The CI for AVE is estimated under the normality assumption. BBERT denotes BioBERT, PBERT denotes PubMedBERT, and SBERT denotes SAPBERT. For simplicity, the notations below remain the same and are omitted.}
\label{R2}
\end{table}

In Supplementary Table~\ref{R2}, we report the results corresponding to Fig.~\ref{P2_AUC} in the \emph{Results} section of the main text, along with ablation study results described in the \emph{Ablation study results} section of the main text. We also include performance results for GAME (BGE) and GAME (CODER), as introduced in Section~\ref{sec:S2}. We report the $95\%$ confidence intervals (CIs) for the AUCs, obtained via $100$ bootstrap iterations, and compute $p$-values using the same method. Except for methods that outperform GAME, all $p$-values testing whether GAME is significantly better are below $1 \times 10^{-3}$. Thus, we omit reporting individual $p$-values for brevity.  For the ablation study, methods (a) through (f) use one-step training, while (g) and GAME use two-step training, as detailed in the \emph{Ablation studies} section in the main text.  Accordingly, methods (a) to (f) use $768$-dimensional embeddings for all tasks, whereas (g), (h), and  GAME use $256$-dimensional embeddings for the similarity task and $768$-dimensional embeddings for the relatedness task.

From (a) to (b), the similarity AUC remains largely unchanged, while the relatedness AUC increases significantly. This is expected, as we add UMLS edges in (b), which include many useful relatedness relationships. From (b) to (c), we introduce binomial PPMI edges without GPT-4 refinement. As a result, the AUC for both similarity and relatedness drops substantially, suggesting that the added information introduces considerable noise. In contrast, from (b) to (d), where we add GPT-4–refined feature selection edges, the results remain similar. Adding code mapping edges in (e) slightly improves the similarity AUC but slightly decreases the relatedness AUC. This observation is consistent with the nature of code mapping, which contributes more to similarity than to relatedness.

Comparing (e) and (f), both of which use GPT-4-generated edges, we observe that the similarity AUC remains largely unchanged, while the relatedness AUC drops substantially. This suggests that GPT-4 alone does not capture all relevant relatedness information, and that incorporating UMLS relationships remains beneficial. We then compare (g) and GAME. The only difference is that (g) uses random negative pairs in feature selection and code mapping, as described in the \emph{Ablation studies} section in the main text, whereas GAME uses hard negatives in both stages. Although the improvement from (g) to GAME is smaller than in previous comparisons, GAME still achieves higher similarity and relatedness AUCs, indicating that hard negative sampling contributes to broader relationship detection.

Next, we contrast (e) and GAME. These two variants use the same edges and loss functions but differ in the training strategy: (e) performs joint training, while GAME adopts two-step training. The relatedness AUCs are comparable, but GAME shows a clear improvement in similarity AUC. This suggests that two-step training helps prevent interference from relatedness learning, leading to more effective low-dimensional embeddings and overall better performance. Finally, when SAPBERT is replaced by BGE or CODER, GAME achieves even higher AUCs for both similarity and relatedness. This is reasonable, as BGE and CODER generally outperform SAPBERT, especially in capturing similarity relationships, which enhances semantic representation. 

{ Finally, comparing (h) and GAME, they perform comparably on relatedness detection, but (h) performs much worse than GAME on similarity tasks, possibly because incorporating graph information during the PPMI alignment step helps support subsequent similarity training in lower-dimensional spaces.}
Additional details of similarity and relatedness AUCs by relation type are presented in Supplementary Table~\ref{R_sim}.

\begin{table}[H]
\centering
\setlength{\tabcolsep}{2pt}
\resizebox{1\textwidth}{!}{%
\begin{tabular}{c|c|ccccccc|c|ccccccc}
\toprule
\multirow{2}{*}{Type} & \multirow{2}{*}{{Source}} & \multicolumn{7}{c|}{PPMI-SVD}& \multirow{2}{*}{AVE} &
\multirow{2}{*}{BBERT} & \multirow{2}{*}{PBERT} & \multirow{2}{*}{SBERT} & \multirow{2}{*}{CODER} & \multirow{2}{*}{BGE} & \multirow{2}{*}{OpenAI} & \multirow{2}{*}{GAME} \\
 & & BCH & BDX & Duke & MGB & MIMIC & UPMC & VA &  & & & & & & & \\
\midrule
\multirow{9}{*}{Similar}
& LOINC Hierarchy         & 0.948 & 0.944 & 0.909 & 0.987 &       & 0.892 & 0.981 & 0.943 & 0.772 & 0.726 & 0.964 & 0.854 & 0.971 & 0.946 & \textbf{0.972} \\
& RxNorm Hierarchy        & 0.722 & 0.876 & 0.725 & 0.915 & 0.941 & 0.847 & 0.890 & 0.845 & 0.637 & 0.564 & 0.630 & 0.813 & 0.818 & \textbf{0.895} & 0.792 \\
& CCAM Hierarchy          &       & 0.835 &       &       &       &       &       & 0.835 & 0.662 & 0.649 & 0.819 & 0.956 & \textbf{0.977} & \textbf{0.977} & 0.964 \\
& PheCode Hierarchy       & 0.854 & 0.950 & 0.881 & 0.959 & 0.932 & 0.960 & 0.970 & 0.929 & 0.570 & 0.587 & 0.823 & \textbf{0.957} & 0.940 & 0.951 & 0.932 \\
& Is a                     & 0.754 & 0.845 & 0.787 & 0.882 & 0.908 & 0.796 & 0.872 & 0.835 & 0.534 & 0.555 & 0.751 & 0.851 & 0.862 & 0.876 & \textbf{0.905} \\
& CHD                & 0.802 & 0.895 & 0.807 & 0.912 & 0.907 & 0.855 & 0.904 & 0.869 & 0.591 & 0.560 & 0.816 & 0.893 & 0.922 & 0.924 & \textbf{0.925} \\
& classifies              & 0.819 & 0.879 & 0.759 & 0.938 & 1.000 & 0.890 & 0.889 & 0.896 & 0.570 & 0.608 & 0.736 & 0.935 & 0.916 & \textbf{0.943} & 0.940 \\
& part of                 & 0.808 &       & 0.794 &       &       & 0.790 &       & 0.797 & 0.273 & 0.172 & 0.454 & 0.928 & 0.794 & \textbf{0.974} & 0.971 \\
& mapped to               & 0.835 & 0.880 & 0.874 & 0.888 & 0.881 & 0.849 & 0.893 & 0.871 & 0.502 & 0.511 & 0.797 & 0.897 & 0.915 & 0.907 & \textbf{0.964} \\
\midrule
\multirow{16}{*}{Related} & may treat & 0.744 & 0.907 & 0.707 & 0.906 & 0.930 & 0.850 & 0.854 & 0.843 & 0.678 & 0.661 & 0.769 & 0.608 & 0.884 & 0.901 & \textbf{0.964} \\
 & clinically associated with & 0.758 & 0.816 & 0.886 & 0.781 & 0.931 & 0.700 & 0.761 & 0.805 & 0.598 & 0.589 & 0.708 & 0.659 & 0.798 & 0.653 & \textbf{0.894} \\
 & has contraindicated drug & 0.647 & 0.713 & 0.200 & 0.728 &       & 0.630 & 0.657 & 0.629 & 0.691 & 0.612 & 0.738 & 0.488 & 0.818 & 0.772 & \textbf{0.918} \\
 & manifestation of & 0.821 & 0.763 & 0.647 & 0.771 & 0.760 & 0.635 & 0.689 & 0.726 & 0.567 & 0.522 & 0.641 & 0.638 & 0.694 & 0.713 & \textbf{0.865} \\
 & RO & 0.812 & 0.881 & 0.771 & 0.886 & 0.883 & 0.804 & 0.843 & 0.840 & 0.530 & 0.560 & 0.752 & 0.820 & 0.841 & 0.869 & \textbf{0.897} \\
 & may prevent & 0.762 & 0.921 &       & 0.885 &       & 0.860 & 0.885 & 0.863 & 0.736 & 0.576 & 0.783 & 0.653 & 0.904 & 0.937 & \textbf{0.954} \\
 & co-occurs with & 0.798 & 0.948 & 0.890 & 0.895 & 0.955 & 0.795 & 0.922 & 0.886 & 0.598 & 0.664 & 0.784 & 0.783 & 0.916 & 0.868 & \textbf{0.927} \\
 & associated morphology of & 0.692 & 0.749 & 0.732 & 0.833 & 0.800 & 0.645 & 0.744 & 0.742 & 0.448 & 0.511 & 0.684 & 0.728 & 0.802 & 0.805 & \textbf{0.822} \\
 & RB & 0.763 & 0.847 & 0.860 & 0.820 & 0.818 & 0.742 & 0.891 & 0.820 & 0.521 & 0.606 & 0.814 & 0.822 & 0.869 & 0.874 & \textbf{0.933} \\
 & associated with & 0.728 & 0.731 & 0.846 & 0.780 &       & 0.751 & 0.768 & 0.767 & 0.610 & 0.551 & 0.849 & 0.864 & 0.905 & 0.886 & \textbf{0.908} \\
 & active ingredient of &       &       &       &       &       & 0.735 &       & 0.735 & 0.513 & 0.396 & 0.770 & 0.819 & 0.868 & 0.954 & \textbf{0.956} \\
 & ssc & 0.918 & 0.967 & 0.935 & 0.868 & 0.946 & 0.745 & 0.851 & 0.889 & 0.612 & 0.679 & 0.840 & 0.882 & 0.788 & 0.877 & \textbf{0.939} \\
 & related to & 0.667 & 0.877 & 0.653 & 0.963 &       & 0.798 & 0.928 & 0.814 & 0.413 & 0.563 & 0.820 & 0.930 & 0.910 & 0.873 & \textbf{0.880} \\
 & disease may have finding & 0.775 & 0.834 & 0.692 & 0.744 &       & 0.670 & 0.670 & 0.731 & 0.417 & 0.676 & 0.586 & 0.673 & 0.698 & 0.457 & \textbf{0.735} \\
 & RQ & 0.923 & 0.826 & 0.876 & 0.973 &       & 0.965 & 0.862 & 0.904 & 0.671 & 0.651 & 0.834 & 0.813 & 0.886 & 0.779 & \textbf{0.938} \\
 & causative agent of & 0.722 &       &       & 0.992 &       & 0.763 &       & 0.825 & 0.578 & 0.637 & 0.762 & 0.734 & 0.879 & 0.879 & \textbf{0.910} \\
\bottomrule
\end{tabular}
}
\begin{flushleft}
\footnotesize
\textbf{Bold} indicates the highest AUC among all methods for each source–relation pair.
\end{flushleft}
\caption{AUCs of between-vector cosine similarity for detecting known similar and related pairs using various methods. Empty entries indicate that no such source exists for the corresponding institution. The meanings of ``CHD'', ``RO'', ``RB'', and ``RQ'' are the same as in Supplementary Table~\ref{num_table}. }
\label{R_sim}
\end{table}

\subsubsection{Detecting institutional similarity and relatedness pairs}
\label{result_AUC_insti}

To account for heterogeneity across institutions, we conducted institution-specific analyses to evaluate the performance of GAME in more localized settings. These analyses enhance the credibility and interpretability of our findings in real-world, multi-institutional environments. Specifically, we focused on code pairs that appear in only one or two institutions, referring to them as \textit{institution-specific pairs}. The rationale is that code pairs occurring in three or more institutions are more likely to be common and widely shared, and thus not reflective of institution-specific characteristics. We evaluated these pairs using AUC metrics across institutions, comparing institutional PPMI-SVD embeddings and various PLM-based embeddings, including GAME.

\begin{table}[ht]
\centering
\resizebox{1\textwidth}{!}{%
\begin{tabular}{lccccccccccc}
\toprule
& PPMI & BBERT & PBERT & SBERT & CODER & BGE & OpenAI & GAME & G (BGE) & G (CODER) & \# Pairs\\
\midrule
BCH   & 0.655 & 0.692 & 0.678 & 0.858 & 0.848 & 0.924 & \textbf{0.928} & 0.881  & 0.918 & 0.924 & 1954\\
BDX   & 0.795 & 0.668 & 0.650 & 0.835 & 0.912 & 0.961 & \textbf{0.965} & 0.946 & 0.951 & 0.946 & 6526\\
Duke  & 0.532 & 0.692 & 0.697 & 0.808 & 0.841 & 0.931 & \textbf{0.945} & 0.877 & 0.921 & 0.937 & 677\\
MGB   & 0.694 & 0.734 & 0.676 & 0.871 & 0.837 & 0.936 & 0.935 & 0.915 & 0.924 & \textbf{0.937} & 2875\\
MIMIC & \textbf{0.947} & 0.671 & 0.594 & 0.494 & 0.837 & 0.769 & 0.905 & 0.839 & 0.803 & 0.816 & 516\\
UPMC  & 0.800 & 0.703 & 0.670 & 0.866 & 0.866 & 0.933 & \textbf{0.941} & 0.909 & 0.929 & 0.932 & 4836\\
VA    & 0.609 & 0.688 & 0.648 & 0.824 & 0.846 & 0.914 & \textbf{0.929} & 0.882 & 0.907 & 0.910 & 1834\\
\midrule
AVE   & 0.719 & 0.693 & 0.659 & 0.794 & 0.855 & 0.910 & \textbf{0.935} & 0.893 & 0.908 & 0.915 & \\
\bottomrule
\end{tabular}
}
\begin{flushleft}
\footnotesize
\textbf{Bold} indicates the highest AUC within each row.
\end{flushleft}
\caption{Institutional AUC scores for detecting institution-specific similarity relationships using various methods. AVE denotes the average AUC score across all institutions. \# Pairs denotes the number of evaluated pairs in each institution. G (BGE) and G (CODER) denote  GAME (BGE) and GAME (CODER), respectively.}
\label{tab:simi_inst_AUC}
\end{table}

\begin{table}[ht]
\centering
\resizebox{1\textwidth}{!}{%
\begin{tabular}{lccccccccccc}
\toprule
& PPMI & BBERT & PBERT & SBERT & CODER & BGE & OpenAI & GAME & G (BGE) & G (CODER) & \# Pairs\\
\midrule
BCH   & 0.509 & 0.633 & 0.639 & 0.746 & 0.656 & 0.840 & 0.858 & 0.934 & \textbf{0.955} & \textbf{0.955}  & 432 \\
BDX   & 0.603 & 0.572 & 0.611 & 0.727 & 0.649 & 0.816 & 0.848 & 0.938 & 0.948 &  \textbf{0.956} & 518 \\
Duke  & 0.493 & 0.603 & 0.626 & 0.751 & 0.657 & 0.833 & 0.851 & 0.946& \textbf{0.957} & 0.954 & 350 \\
MGB   & 0.582 & 0.621 & 0.646 & 0.750 & 0.669 & 0.839 & 0.851 & 0.942 & \textbf{0.950} & \textbf{0.950} & 615 \\
MIMIC & 0.469 & 0.576 & 0.637 & 0.725 & 0.643 & 0.776 & 0.872 & 0.921 & 0.956 & \textbf{0.968} & 239 \\
UPMC  & 0.605 & 0.587 & 0.617 & 0.760 & 0.681 & 0.831 & 0.859 & 0.942 & 0.947 & \textbf{0.952} &  684 \\
VA    & 0.534 & 0.590 & 0.663 & 0.723 & 0.644 & 0.830 & 0.859 & 0.941 & 0.952 & \textbf{0.959} & 573 \\
\midrule
AVE   & 0.542 & 0.597 & 0.634 & 0.740 & 0.657 & 0.824 & 0.857 & 0.938 & 0.952& \textbf{0.956} &\\
\bottomrule
\end{tabular}
}
\begin{flushleft}
\footnotesize
\textbf{Bold} indicates the highest AUC within each row.
\end{flushleft}
\caption{Institutional AUC scores for detecting institution-specific relatedness relationships using various methods. AVE denotes the average AUC score across all institutions. \# Pairs denotes the number of evaluated pairs in each institution. G (BGE) and G (CODER) denote  GAME (BGE) and GAME (CODER), respectively.}
\label{tab:rela_inst_AUC}
\end{table}

Supplementary Table~\ref{tab:simi_inst_AUC} presents the results for institution-specific similarity detection. We observe that OpenAI and BGE outperform GAME, consistent with the findings in Supplementary Table~\ref{R2}. On average, GAME's similarity AUC is $4.2\%$ lower than that of OpenAI. This performance gap is expected, given that GAME uses $256$-dimensional similarity embeddings, whereas OpenAI employs $1,536$-dimensional embeddings—six times larger in size. Notably, when replacing SAPBERT with stronger pretrained models such as BGE and CODER, GAME's performance improves significantly. The average gap in similarity AUC relative to OpenAI narrows to $2.7\%$ with BGE and $2.0\%$ with CODER, respectively. These results further highlight the efficiency of the GAME architecture: better pretrained input embeddings can lead to even stronger outcomes despite lower embedding dimensionality.

In contrast, Supplementary Table~\ref{tab:rela_inst_AUC} shows that GAME consistently achieves the highest AUC scores for relatedness detection across all institutions, outperforming all baseline methods. While PPMI-SVD embeddings perform reasonably well on common code pairs (as seen in Supplementary Table~\ref{R2}), its effectiveness deteriorates markedly for institution-specific codes. For instance, the AUC drops to $0.469$ in MIMIC, $0.493$ in Duke, and $0.509$ in BCH—barely above random guessing.

This stark contrast underscores the key strength of GAME’s federated learning framework: it enables institutions to benefit from shared semantic representations learned across sites, even for rare or institution-specific codes. Unlike PPMI-SVD embeddings, which rely solely on local co-occurrence statistics and struggles with data sparsity, GAME’s harmonized embedding enables robust generalization. As a result, GAME maintains strong performance across both global and local settings, making it particularly effective in federated environments where code overlap is inherently limited.

\subsection{Cross-institutional code mapping}
\label{sec:code_map_supp}

In this section, we present results on local code mapping. Section~\ref{sec:code_map_VA} details the validation and outcomes of VA-specific code mapping, including analyses of how embedding dimensionality and description quality affect performance. Notably, we show that federated learning is essential for effective code mapping; training GNN models within a single institution is insufficient.
In Section~\ref{Sec:3.3.2}, we generalize the VA results to a broader local code mapping task using a set of human-curated labels, further reinforcing and extending our findings.

\subsubsection{Evaluation of VA local lab code mapping accuracy}
\label{sec:code_map_VA}

As described in the \emph{Methods} section of the main text, we evaluated the accuracy of mapping VA local lab codes to LOINC codes. Specifically, we calculated top \(k\) accuracy for each local lab code by selecting the LOINC codes with the highest cosine similarity across different embeddings. Accuracy was measured by checking whether the correct mapping was included among the top \(k\) selected LOINC/LP codes (based on cosine similarity), with evaluations at \(k = 1\), \(5\), \(10\), and \(20\). Accuracy was assessed at the LP level—meaning that if a local lab code mapped to the parent or sibling of the correct LOINC leaf code, it was considered correct.

Similar to Table~\ref{R1} in the \emph{Results} section of the main text, we report VA code mapping accuracy for the ablation settings described in the \emph{Ablation study results} section in the main text, as shown in Supplementary Table~\ref{tab:R1_ablation}. We observe that using only hierarchical edges (a), or adding UMLS edges (b), results in similarly low accuracy. Adding PPMI edges without GPT-4 refinement (c) leads to a significant performance drop compared to (a) and (b), suggesting that noisy PPMI edges impair similarity-based tasks such as code mapping.

In contrast, incorporating GPT-4–refined feature selection edges (d) improves performance beyond (a) and (b), indicating the added information is both cleaner and more informative than in (c). Further improvement is observed in (e), where adding GPT-4–revised code mapping edges yields the highest TOP1 accuracy among all settings. Even when using only the GPT-4–revised feature selection and code mapping edges—as in (f)—the embedding achieves notably better performance than (a), (b), and (d), underscoring the value of GPT-4–curated information.

All results above are based on one-step training and use full $768$-dimensional embeddings without isolating the similarity component. In contrast, settings (g) and GAME apply two-step training with a $256$-dimensional similarity embedding. Interestingly, although the TOP1 accuracy of (g) and GAME is slightly lower than that of (e), which uses combined similarity and relatedness information, their TOP20 accuracy is higher. This suggests that the low-dimensional similarity embedding obtained via two-step training is not only sufficient but also more robust for code mapping.

Comparing (g) and GAME, both use GPT-4–revised edges, but (g) employs randomly sampled negatives, while GAME uses hard negatives—candidate codes rejected by GPT-4, as described in the \emph{Methods} section in the main text. We find that (g) achieves higher TOP1 and TOP20 accuracy, whereas GAME performs better at TOP5 and TOP10. Although hard negatives show limited advantage here, we will demonstrate their effectiveness in Supplementary Table~\ref{R2_code_map_2}.

{ In addition, (h) and GAME show comparable TOP1 accuracy, while (h) attains 3–4\% lower scores in the TOP5, TOP10, and TOP20 metrics compared to GAME. This may be because (h) still performs coarsely in code mapping without incorporating additional information during the PPMI step. We will further demonstrate their difference in Supplementary Table~\ref{R2_code_map_2}.}

Lastly, GAME (BGE) and GAME (CODER) slightly outperform the SAPBERT-based variant, especially in TOP1 accuracy. This aligns with the observation that BGE and CODER embeddings yield better raw code mapping performance (e.g., in TOP20 accuracy). However, the improvement is modest, suggesting that GAME can effectively integrate semantic signals from different pretrained models to achieve strong and consistent performance.

\begin{table}[H]
\centering
\begin{tabular}{cc|cccc}
\toprule
\multicolumn{2}{c|}{} & TOP1 (\%) & TOP5 (\%) & TOP10 (\%) & TOP20 (\%) \\
\midrule
\multicolumn{2}{c|}{GAME}         & 74.2 & 88.0 & 90.7 & 92.7 \\
\multicolumn{2}{c|}{GAME (BGE)}   & 76.8 & 88.3 & \textbf{91.6} & \textbf{93.2} \\
\multicolumn{2}{c|}{GAME (CODER)} & 76.6 & \textbf{89.0} & 91.1 & 92.8 \\
\midrule
\multirow{7}{*}{Ablation}
& (a) & 66.6 & 75.7 & 79.1 & 83.7 \\
& (b) & 66.0 & 75.6 & 79.8 & 83.6 \\
& (c) & 26.1 & 38.4 & 45.1 & 51.7 \\
& (d) & 69.0 & 80.6 & 83.1 & 85.7 \\
& (e) & \textbf{77.5} & 86.5 & 88.9 & 90.4 \\
& (f) & 71.2 & 87.1 & 89.6 & 91.1 \\
& (g) & 76.0 & 87.5 & 90.5 & 93.0 \\
& (h) & 75.9 &	84.2 & 87.4 & 89.5 \\
\bottomrule
\end{tabular}
\begin{flushleft}
\footnotesize
\textbf{Bold} indicates the highest accuracy within each column.
\end{flushleft}
\caption{Accuracy of mapping VA local lab codes to LOINC/LP codes using different methods.}
\label{tab:R1_ablation}
\end{table}

 \paragraph{Comparative performance of baseline embeddings.}

As shown in Supplementary Table~\ref{R1_GPT}, BGE and OpenAI embeddings with reduced dimensionality ($256$, matching the similarity component of the GAME embedding) perform noticeably worse than their higher-dimensional counterparts. This highlights the efficiency of the GAME embedding, which achieves strong performance using a compact $256$-dimensional representation.
 
\begin{table}[H]
\centering
\begin{tabular}{c|cc|ccc}
\toprule
\multirow{2}{*}{\diagbox{{Measure (\%)}}{{Dimension}}} & \multicolumn{2}{c|}{{BGE}} & \multicolumn{3}{c}{{OpenAI}} \\ 
 &{256} & {768} &{256} & {768} & {1536} \\ \midrule
{TOP1}  & 57.2 & 61.9 & 56.8 & 62.2 & 61.9 \\ 
{TOP5}  & 76.5 & 79.0 & 78.5 & 86.4 & 85.9 \\ 
{TOP10} & 83.8 & 84.3 & 84.2 & 89.6 & 89.8 \\ 
{TOP20} & 87.8 & 89.7 & 87.9 & 92.6 & 92.6 \\ 
\bottomrule
\end{tabular}
\caption{Accuracy of code mapping using BGE and OpenAI embeddings across different dimensions.}
\label{R1_GPT}
\end{table}

Additionally, we present the results of code mapping using SAPBERT and CODER embeddings based on uncorrected descriptions in Supplementary Table~\ref{tab:before_GPT}, without applying the GAME algorithm. In this table, \textit{original} refers to the accuracy using the unmodified descriptions of the VA local lab; \textit{+dict} indicates the accuracy after refining abbreviations in the original descriptions using a dictionary; and \textit{+GPT-4} represents the accuracy after further refinement using GPT-4, as detailed in the \emph{Methods} section of the main text. The results show that GPT-4–corrected descriptions significantly improve code mapping accuracy, and that the GAME framework can further enhance performance beyond these improvements.

\begin{table}[H]
\centering
\begin{tabular}{c|ccc|ccc}
\toprule
\multirow{2}{*}{\diagbox{{Measure (\%)}}{{Description}}} & \multicolumn{3}{c|}{{SAPBERT}} & \multicolumn{3}{c}{{CODER}} \\ 
 & {original} & {+dict} & {+GPT-4}& {short} & {+dict} & {+GPT-4} \\ \midrule
{TOP1}   & 52.2 &  54.7    & 59.7   & 52.7 & 53.2  & 55.6        \\ 
{TOP5}   & 67.3 &  70.8    & 79.0   & 71.7 & 72.5  & 75.9        \\ 
{TOP10}  & 71.1 &  74.6    & 82.9   & 77.8 & 78.7  & 85.0         \\ 
{TOP20}  & 74.7 &  77.7    & 85.7   & 82.4 & 83.3  & 89.7          \\ \bottomrule
\end{tabular}
\caption{Accuracy of VA local lab mapping after each step of refining descriptions.}
\label{tab:before_GPT}
\end{table}

\paragraph{{The role of GNN and federated learning in code mapping.}}

Although GAME, which integrates both  GNN and FL, outperforms all baseline methods, we further investigate whether GNN alone is sufficient for the code mapping task. To this end, we train GNN models independently within each institution and then perform cross-institutional code mapping using task-relevant embeddings and GPT-4-generated alignment pairs.

Specifically, we focus on mapping VA local lab codes to MGB LOINC codes. We first train two separate GAT models using institutional edges from MGB and VA, respectively. The input embeddings are constructed by concatenating each institution’s PPMI-SVD embeddings with SAPBERT embeddings. Within each institution, we construct graph edges and define loss functions using only hierarchical and UMLS-based relationships. This yields distinct embeddings for all codes in MGB and VA. 

We then extract only the MGB LOINC codes and VA local lab codes and apply two alignment strategies, both leveraging GPT-4-generated mapping pairs:
\begin{itemize}
\item \textbf{Additional GAT with contrastive learning:} A GAT model is trained on GPT-4-generated positive edges between MGB LOINC and VA local lab codes, optimized using contrastive loss.
\item \textbf{Procrustes rotation:} The MGB embeddings are directly aligned to the VA embeddings via Procrustes transformation \citep{smith2017offline}
using the GPT-4-aligned code pairs. 
\end{itemize}

\begin{table}[ht]
\centering
\begin{tabular}{c|ccc}
\toprule
{Measure (\%)} & {Contrastive} & {Procrustes} & {GAME} \\
\midrule
{TOP1}  & 65.5 & 63.9 & \textbf{79.6} \\
{TOP5}  & 77.1 & 76.1 & \textbf{88.9} \\
{TOP10} & 79.7 & 82.0 & \textbf{91.9} \\
{TOP20} & 82.4 & 86.9 & \textbf{94.2} \\
\bottomrule
\end{tabular}
\begin{flushleft}
\footnotesize
\textbf{Bold} indicates the highest accuracy within each row.
\end{flushleft}
\caption{Accuracy of mapping VA local lab codes to LOINC/LP codes in MGB using different methods.
}
\label{tab:inst_ablation_code_mapping}
\end{table}

As shown in Supplementary Table~\ref{tab:inst_ablation_code_mapping}, GAME significantly outperforms both alignment strategies based on GNN alone. 
This evaluation includes $1,834$ VA local lab codes (a subset of the original $2,024$) for which LOINC mappings are available in MGB. Accordingly, the reported GAME accuracy differs slightly from that in Table~\ref{R1} of the main paper.

These results demonstrate that federated learning is crucial for achieving robust and unified embeddings across institutions. While GNNs trained within isolated institutions can capture local structure, they fall short in aligning code semantics effectively across sites. The combination of GNN and FL, as implemented in GAME, enables meaningful cross-institutional generalization and achieves superior performance in code mapping tasks.

\subsubsection{Generalized local code mapping evaluation}
\label{Sec:3.3.2}

\begin{table}[H]
\centering
\begin{tabular}{cc|c|c|c}
\toprule
\multicolumn{2}{c|}{} & {UPMC PX-CCS} & {BDX CCAM-CCS} & {UPMC LAB-LOINC} \\ \midrule
\multicolumn{2}{c|}{ BBERT}        & 0.059 & 0.138 & 0.102  \\ 
\multicolumn{2}{c|}{ PBERT}         & 0.001 & 0.317 & 0.112  \\ 
\multicolumn{2}{c|}{ SBERT}         & 0.313 & 0.423 & 0.529  \\ 
\multicolumn{2}{c|}{ CODER}         & 0.418 & 0.540 & 0.554  \\ 
\multicolumn{2}{c|}{ BGE}          & 0.409 & 0.615 & 0.610  \\ 
\multicolumn{2}{c|}{ OpenAI}       & 0.484 & 0.616 & 0.583  \\ \midrule
\multicolumn{2}{c|}{{GAME}} & 0.574 & 0.671 & \textbf{0.654} \\ 
\multicolumn{2}{c|}{{GAME (BGE)}} & 0.589 & 0.696 & 0.645 \\
\multicolumn{2}{c|}{{GAME (CODER)}} & \textbf{0.612} & \textbf{0.702} & 0.648 \\\midrule
\multirow{7}{*}{Ablation} 
& (a) & 0.482 & 0.585 & 0.493 \\ 
& (b) & 0.475 & 0.584 & 0.495 \\ 
& (c) & 0.440 & 0.508 & 0.443 \\
& (d) & 0.518 & 0.627 & 0.482 \\
& (e) & 0.587 & 0.654 & 0.563 \\ 
& (f) & 0.584 & 0.659 & 0.569 \\ 
& (g) & 0.488 & 0.593 & 0.613 \\
& (h) & 0.386 & 0.515 & 0.559 \\
\midrule
\multicolumn{2}{c|}{{\# Pairs}} & 199 & 537 & 1814 \\
\multicolumn{2}{c|}{{\#  Standard Code}} & 10 & 29 & 92 \\ \bottomrule
\end{tabular}
\begin{flushleft}
\footnotesize
\textbf{Bold} indicates the highest correlation within each column.
\end{flushleft}
\caption{
Spearman's rank correlation between cosine similarity scores and human annotations. A higher rank correlation indicates that cosine similarity better reflects the relationship between local codes and standard codes. \# Pairs denotes the number of evaluated pairs; \# Standard Codes is the number of unique standard codes. \# Pairs denotes the number of evaluated pairs, while \#  Standard Code denotes the total number of unique standard codes.
}
\label{R2_code_map_2}
\end{table}

Supplementary Table~\ref{R2_code_map_2} presents the full results corresponding to Fig.~\ref{fig:code_map_corr}, including the ablation settings. Many trends are consistent with those in Supplementary Table~\ref{tab:R1_ablation}. Specifically, configurations (a) and (b) perform similarly poorly, while (c) shows a further drop, confirming that binomial PPMI edges introduce substantial noise that impairs code mapping. In contrast, (d) improves over (a) and (b), indicating that GPT-4–refined feature selection edges capture relationships relevant to mapping. Performance increases further in (e) with the addition of GPT-4–revised code mapping edges. Notably, (f), which uses only GPT-4–refined edges, performs comparably to (e), reinforcing the utility of curated information from GPT-4. These patterns largely mirror those observed in Supplementary Table~\ref{tab:R1_ablation}.

However, { two key differences emerge. Firstly,} unlike Supplementary Table~\ref{tab:R1_ablation}, where (g) and GAME perform comparably, Supplementary Table~\ref{R2_code_map_2} shows a clear advantage for GAME. As described in the \emph{Ablation studies} section in the main text, (g) uses randomly sampled negatives in \(\mathcal{N}_i\) during both feature selection and code mapping, while GAME incorporates hard negatives—codes suggested by other methods but rejected by GPT-4.  This suggests that although random negatives may suffice for binary tasks such as VA code mapping (Supplementary Table~\ref{tab:R1_ablation}), they are insufficient in more nuanced scenarios requiring ordinal labels (e.g., 1 = ``yes,'' 0.5 = ``possible,'' 0 = ``no''). Such tasks demand that the embeddings preserve fine-grained ranking, underscoring the importance of hard negatives for learning these subtleties.

{ Secondly, (h) shows a larger performance gap compared to GAME. As we can see, (h) exhibits much lower Spearman’s rank correlations than GAME, and the correlations for UPMC PX–CCS and BDX CCAM–CCS are even worse than those of (a), which uses extended graph information in the align-PPMI step but applies hierarchical information only in other steps. Although (h) leverages full knowledge in  two step training, it still performs worse than (a), which uses full graph information to align PPMI from the very beginning. This further demonstrates that aligning PPMI with graph information is much more effective than simple PPMI pooling.}

Finally, GAME achieves strong performance using only a $256$-dimensional similarity embedding, outperforming (e), which uses a $768$-dimensional joint embedding. This further supports the effectiveness of the two-step training strategy: by isolating similarity from relatedness, GAME prevents weaker relatedness signals from interfering with the optimization of similarity-specific tasks, thereby enabling more compact and robust representations. Moreover, when the input pretrained embedding is changed from SAPBERT to BGE or CODER, GAME performs even slightly better, further demonstrating the effectiveness of the GAME algorithm.

\subsection{Feature selection}
\label{sec:fea_sel_supp}

In this section, we present the results of feature selection using both human annotations (Section~\ref{sec:gpt_make_sense}) and GPT-4 scores (Section~\ref{sec:fea_result_all}). In Section~\ref{sec:gpt_make_sense}, we also demonstrate that GPT-4 scores can serve as reliable evaluations in the absence of human annotations.

\subsubsection{Validation of feature selection against human annotations}  
  
\label{sec:gpt_make_sense}  

\begin{table}[H]
\centering
\setlength{\tabcolsep}{2pt}
\begin{tabular}{cc|cccccc|c}
\toprule
\multicolumn{2}{c|}{{Method}} & {T1D} & {Epilepsy} & {PH} & {Asthma} & {CD} & {UC} & {AVE} \\
\midrule
\multirow{7}{*}{{PPMI-SVD}} 
& {BCH}   & 0.922 & 0.897 & --  & 0.830 & 0.903 & 0.852 & 0.881 \\
& {BDX}   & 0.799 & 0.938 & 0.707 & 0.793 & 0.885 & 0.780 & 0.817 \\
& {Duke}  & 0.619 & 0.860 & --  & 0.725 & 0.826 & 0.885 & 0.783 \\
& {MGB}   & 0.697 & 0.882 & 0.781 & 0.768 & 0.891 & 0.816 & 0.806 \\
& {MIMIC} & 0.713 & 0.759 & --  & --    & 0.862 & 0.918 & 0.813 \\
& {UPMC}  & 0.866 & 0.885 & 0.713 & 0.729 & 0.849 & 0.831 & 0.812 \\
& {VA}    & 0.826 & 0.899 & 0.811 & 0.735 & 0.892 & 0.765 & 0.821 \\
\multicolumn{2}{c|}{{AVE}} & 0.777 & 0.874 & 0.753 & 0.763 & 0.872 & 0.835 & 0.813 \\
\midrule
\multicolumn{2}{c|}{{BBERT}} & 0.431 & 0.607 & 0.497 & 0.639 & 0.665 & 0.486 & 0.554 \\
\multicolumn{2}{c|}{{PBERT}} & 0.524 & 0.526 & 0.417 & 0.492 & 0.477 & 0.839 & 0.546 \\
\multicolumn{2}{c|}{{SBERT}} & 0.758 & 0.804 & 0.454 & 0.662 & 0.682 & 0.764 & 0.687 \\
\multicolumn{2}{c|}{{CODER}} & 0.803 & 0.808 & 0.595 & 0.676 & 0.694 & 0.592 & 0.695 \\
\multicolumn{2}{c|}{{BGE}}   & 0.721 & 0.728 & 0.608 & 0.643 & 0.840 & 0.801 & 0.724 \\
\multicolumn{2}{c|}{{OpenAI}} & \textbf{0.856} & 0.861 & 0.628 & {0.835} & 0.586 & 0.771 & 0.756 \\ \midrule
\multicolumn{2}{c|}{{GAME}}  & 0.823 & {0.928} & \textbf{0.841} & 0.822 & 0.928 & 0.834 & \textbf{0.863} \\
\multicolumn{2}{c|}{{GAME (BGE)}}  & 0.795 & 0.923 & 0.797 & 0.835 & {0.941} & \textbf{0.853} & 0.857 \\
\multicolumn{2}{c|}{{GAME (CODER)}}  & 0.809 & \textbf{0.931} & 0.792 & \textbf{0.845} & 0.911 & 0.825 & 0.852 \\\midrule
\multirow{7}{*}{Ablation}
& (a) & 0.815 & 0.891 & 0.670 & 0.818 & 0.853 & 0.778 & 0.804 \\ 
& (b) & 0.777 & 0.889 & 0.670 & 0.802 & 0.903 & 0.794 & 0.806 \\
& (c) & 0.767 & 0.898 & 0.693 & 0.835 & {0.928} & 0.796 & 0.820 \\
& (d) & 0.750 & 0.896 & 0.786 & 0.821 & 0.915 & 0.787 & 0.826 \\
& (e) & 0.778 & 0.912 & 0.801 & 0.824 & 0.911 & {0.844} & 0.845 \\
& (f) & 0.770 & 0.916 & 0.812 & 0.829 & 0.882 & 0.839 & 0.841 \\
& (g) & 0.845 & 0.911 & 0.798 & 0.820 & 0.915 & 0.853 & 0.857 \\ 
& (h) & 0.795 & 0.919 & 0.701 & 0.799 & \textbf{0.945} & 0.834 & 0.832 \\ 
\midrule
\multicolumn{2}{c|}{{GPT-4}} &  0.885 & 0.937 & 0.828 & 0.845 & 0.962 & 0.964 &  0.903\\ 
\multicolumn{2}{c|}{{\# Pairs}} & 60 & 300 & 160 & 180 & 40 & 40 \\ 
\bottomrule
\end{tabular}
\begin{flushleft}
\footnotesize
\textbf{Bold} indicates the highest C-index within each column.
\end{flushleft}
\caption{C-index values between cosine similarity scores and expert-labeled feature relevance for six diseases across different embedding methods. Higher values indicate better alignment with expert annotations. The ``GPT-4'' row reports agreement between GPT-4-assigned scores and expert annotations, measured by the C-index. \# Pairs denotes the number of expert-labeled features evaluated for each disease. Results for PH (PheCode 415.21) are unavailable for BCH, Duke, and MIMIC because this disease is not present in those institutions, consistent with Supplementary Table~\ref{R3_}. Additionally, results for Asthma (PheCode 495) are unavailable for MIMIC due to the absence of human annotations in MIMIC.}
\label{tab:fea_sel_human}
\end{table}
In Supplementary Table~\ref{tab:fea_sel_human}, we present the complete results underlying the left part of Fig.~\ref{P30}, including C-index values computed between cosine similarity scores from various embeddings and human-annotated feature relevance labels. For institutional PPMI-SVD embeddings, the evaluation is restricted to annotations available within each corresponding institution. Among all methods, GAME achieves the highest overall C-index among all compared methods, highlighting its strong capability in feature selection. Since the result patterns are very similar to those from the GPT-4 evaluation, we will discuss them together in detail in Section~\ref{sec:fea_result_all}.

Furthermore, Supplementary Table~\ref{tab:fea_sel_human} highlights the strong alignment between GPT-generated feature scores and expert annotations. Using the prompt shown in Supplementary Table~\ref{tab:gpt_prompt}, we computed the C-index between GPT-4-assigned relevance scores and human labels, treating this value as a measure of agreement. Notably, GPT-4 achieves the highest C-index across nearly all diseases, with individual scores exceeding $0.828$ and an average of $0.903$. These results demonstrate that GPT-4-generated scores align closely with human judgment and support their use as reliable silver-standard labels for training and evaluation purposes, as discussed in the \emph{Results} section of the main text.

\subsubsection{Feature selection performance evaluated by GPT-4} 
\label{sec:fea_result_all}

\begin{table}[H]
\centering
\setlength{\tabcolsep}{2pt}
\begin{tabular}{cc|cccccc|c}
\toprule
\multicolumn{2}{c|}{{Method}} & {T1D} & {Epilepsy} & {PH} & {Asthma} & {CD} & {UC} & {AVE} \\
\midrule
\multirow{7}{*}{{PPMI-SVD}} 
& {BCH}   & 0.563 & 0.665 & --  & 0.621 & 0.626 & 0.627 & 0.620 \\
& {Bor}   & 0.668 & 0.736 & 0.650 & 0.627 & 0.639 & 0.671 & 0.665 \\
& {Duke}  & 0.491 & 0.569 & --  & 0.464 & 0.584 & 0.592 & 0.540 \\
& {MGB}   & 0.707 & 0.734 & 0.707 & 0.683 & 0.716 & 0.736 & 0.714 \\
& {MIMIC} & 0.627 & 0.622 & -- & 0.442 & 0.596 & 0.590 & 0.576 \\
& {UPMC}  & 0.639 & 0.656 & 0.615 & 0.611 & 0.582 & 0.604 & 0.618 \\
& {VA}    & 0.680 & 0.714 & 0.744 & 0.597 & 0.658 & 0.674 & 0.678 \\
\multicolumn{2}{c|}{{AVE}} & 0.625 & 0.671 & 0.679 & 0.578 & 0.629 & 0.642 & 0.630 \\
\midrule
\multicolumn{2}{c|}{{BBERT}} & 0.447 & 0.442 & 0.469 & 0.458 & 0.431 & 0.491 & 0.456 \\
\multicolumn{2}{c|}{{PBERT}} & 0.458 & 0.437 & 0.473 & 0.459 & 0.462 & 0.457 & 0.457 \\
\multicolumn{2}{c|}{{SBERT}} & 0.568 & 0.466 & 0.546 & 0.533 & 0.525 & 0.507 & 0.524 \\
\multicolumn{2}{c|}{{CODER}} & 0.644 & 0.598 & 0.630 & 0.619 & 0.534 & 0.554 & 0.597 \\
\multicolumn{2}{c|}{{BGE}}   & 0.630 & 0.542 & 0.596 & 0.568 & 0.549 & 0.542 & 0.571 \\
\multicolumn{2}{c|}{{OpenAI}} & 0.693 & 0.694 & 0.659 & \textbf{0.684} & 0.615 & 0.668 & 0.669 \\ 
\midrule
\multicolumn{2}{c|}{{GAME}}  & \textbf{0.740} & \textbf{0.786} & \textbf{0.725} & 0.668 & 0.684 & 0.665 & \textbf{0.711} \\ 
\multicolumn{2}{c|}{{GAME (BGE)}} & \textbf{0.740} & 0.767 & 0.717 & 0.653 & \textbf{0.687} & \textbf{0.676} & 0.706 \\ 
\multicolumn{2}{c|}{{GAME (CODER)}} & 0.728 & 0.770 & 0.713 & 0.659 & 0.673 & 0.668 & 0.702 \\
\midrule
\multirow{8}{*}{Ablation}
& (a) & 0.696 & 0.752 & 0.600 & 0.648 & 0.653 & 0.648 & 0.666 \\ 
& (b) & 0.683 & 0.724 & 0.605 & 0.582 & 0.636 & 0.604 & 0.639 \\
& (c) & 0.682 & 0.740 & 0.570 & 0.594 & 0.612 & 0.612 & 0.635 \\
& (d) & 0.695 & 0.747 & 0.649 & 0.603 & 0.649 & 0.625 & 0.662 \\
& (e) & 0.714 & 0.758 & 0.666 & 0.622 & 0.664 & 0.654 & 0.680 \\
& (f) & 0.712 & 0.764 & 0.651 & 0.665 & 0.666 & 0.665 & 0.687 \\ 
& (g) & \textbf{0.740} & 0.760 & 0.678 & 0.643 & 0.674 & 0.652 & 0.691 \\ 
& (h) & 0.658 & 0.716 & 0.617 & 0.617 & 0.622 & 0.579 & 0.635 \\
\bottomrule
\end{tabular}
\begin{flushleft}
\footnotesize
\textbf{Bold} indicates the highest C-index within each column.
\end{flushleft}
\caption{C-index between cosine similarity scores and GPT-4-assigned relevance scores across diseases and methods.}
\label{R3_} 
\end{table}

In Supplementary Table~\ref{R3_}, we report the underlying data corresponding to the right panel of Fig.~\ref{P30}, where feature selection performance is evaluated using GPT-4–generated relevance scores. For each method (except institutional PPMI-SVD embeddings), we compute the C-index between its cosine similarity scores and the GPT-4 scores over a unified set of features: the top $100$ features selected by any method for each disease, along with $100$ randomly sampled negative features. For PPMI-SVD embeddings, the C-index is computed using only those features available within each institution to ensure a fair comparison. These results assess the consistency of feature relevance estimation across different embedding methods.

Ablation results show that no matter using human annotations or GPT-4 scores, configurations (g) and GAME outperform all other baselines, likely due to the benefits of two-step training, which cleanly separates similarity and relatedness signals. Notably, GAME consistently outperforms (g), suggesting that hard negative sampling contributes not only to improved code mapping but also to more accurate feature relevance ranking.

The results also offer insight into how different edge constructions affect performance. Configuration (a), which uses only hierarchical edges, yields moderate results. Adding UMLS edges in (b) leads to performance drops in diseases such as T1D, Epilepsy, and Asthma—indicating that blending similarity and relatedness without proper separation can degrade embedding quality. Configuration (c), which incorporates unrefined PPMI edges, performs poorly under GPT-4 evaluation, likely due to noise. In contrast, (d), which includes GPT-refined PPMI edges, shows substantial improvements across both evaluation criteria. Further incorporating GPT-curated code mapping edges in (e) leads to additional gains.

Configuration (f), which uses only GPT-4–curated edges, performs competitively across both evaluation schemes, highlighting GPT-4’s capacity to identify meaningful relationships. { In addition, (h) also performs worse than (a) and is only comparable to (c). Both (h) and (c) incorporate more PPMI information by directly pooling PPMI instead of performing graph-alignment of PPMI and by adding additional raw PPMI pairs. This suggests that including more unfiltered PPMI information without GPT cleaning may introduce additional noise.} Ultimately, GAME achieves the highest overall performance, validating the effectiveness of separating similarity and relatedness through two-step training. Furthermore, when replacing SAPBERT with BGE or CODER, performance degrades only slightly and remains comparable—further supporting the robustness and adaptability of the proposed method.

For reference, the prompt used to generate GPT-4 feature relevance scores is provided in Supplementary Table~\ref{tab:gpt_prompt}.

\begin{table}[H]
\centering
\begin{tabular}{p{3cm} | p{11cm}}
\toprule
{Prompt Type} & {Text} \\
\midrule
System Prompt & 
As a knowledgeable assistant supporting a healthcare professional, your task is to help determine whether various medical codes are related to \texttt{[code description]} based on their provided details. The professional values concise yet precise responses. \\
\midrule
User Prompt & 
Is the following medical code related to \texttt{[code description]}?

\texttt{[feature description]}

Please provide a score between 0 and 1 indicating the likelihood that this code is related to \texttt{[code description]}. For instance, a score of 1 indicates complete relevance, whereas a score of 0 indicates no relevance. Provide your score as a decimal (e.g., 0.13, 0.75, etc.) without any additional information or explanations. \\
\bottomrule
\end{tabular}
\caption{System and user prompts used for GPT-4-based relatedness scoring. Placeholders \texttt{[code description]} and \texttt{[feature description]} were replaced with specific content at runtime.}
\label{tab:gpt_prompt}
\end{table}

\subsection{Joint patient stratification and risk prediction}
\label{supp:patient_stratification}

In this section, we present additional results for joint patient stratification and risk prediction. See the \emph{Methods} section of the main text for methodological details.

\subsubsection{Alzheimer's disease}
\begin{figure}[H]
    \centering
    \includegraphics[width=13cm]{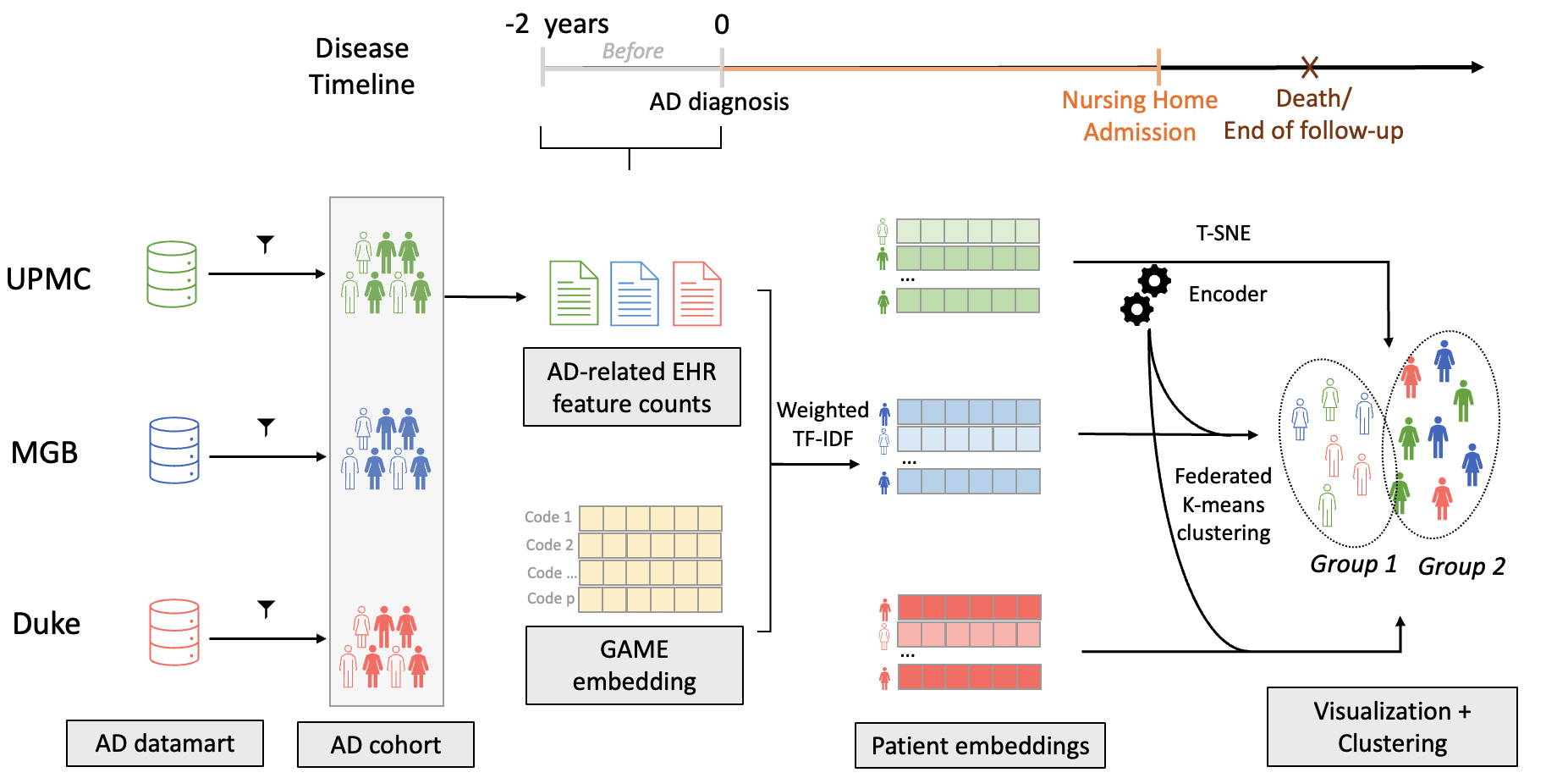}
    \caption{Schematic of joint stratification of AD patients across three instituitons (UPMC, MGB and Duke).}
    \label{fig:AD_flow}
\end{figure}

Supplementary Fig. \ref{fig:AD_flow} shows the schemetic diagram of patient stratification procedure, using AD patient stratification as an example. At each institution, we constructed an AD cohort using a previously validated phenotyping model \citep{venkatesh2025leveraging}. We defined a baseline period to extract relevant feature counts and computed patient embeddings as the weighted sum of feature embeddings. For clustering, embeddings were first reduced to three dimensions using t-SNE approximated via a variational autoencoder \citep{van2008visualizing}. We then applied a federated $k$-means clustering method that shares local centroids across institutions to enable global clustering while preserving data privacy. 

\begin{figure}[H]
    \centering
    \begin{minipage}{\textwidth}
        \begin{tikzpicture}
            \node[anchor=south west, inner sep=0] (img) at (0,0)
                {\includegraphics[width=\textwidth]{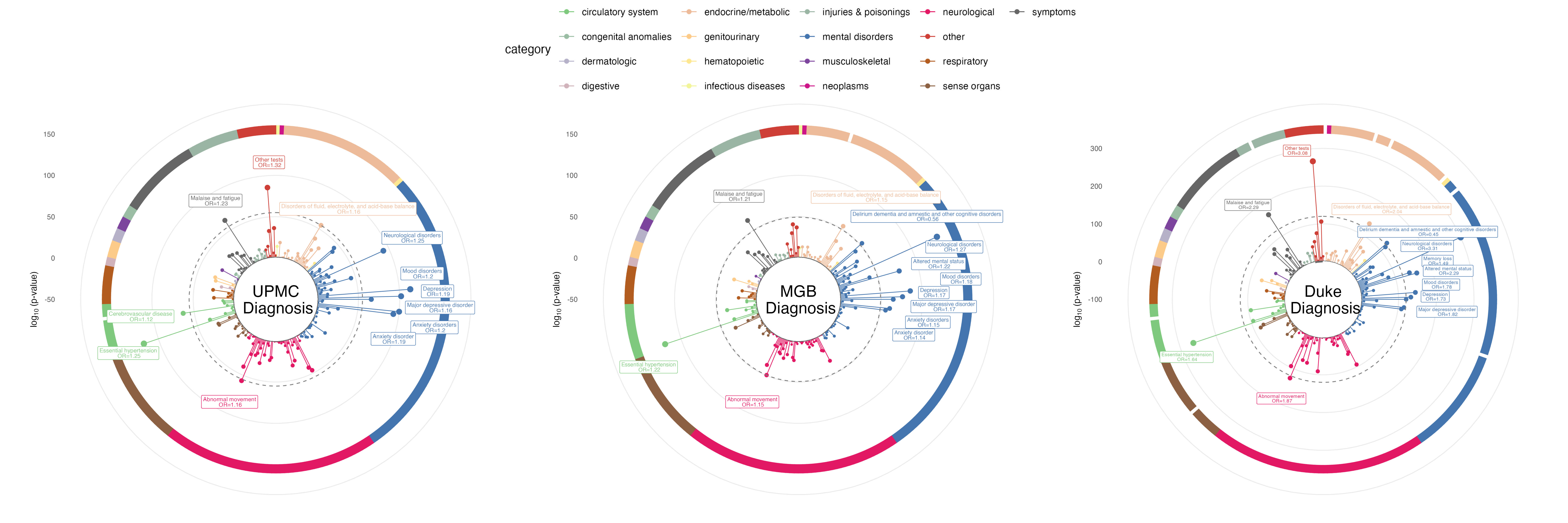}};
            \node[anchor=north west, xshift=5pt, yshift=-5pt] at (img.north west)
                {\textbf{(a)}};
        \end{tikzpicture}
    \end{minipage}

    \vspace{1em}

    \begin{minipage}{\textwidth}
        \begin{tikzpicture}
            \node[anchor=south west, inner sep=0] (img) at (0,0)
                {\includegraphics[width=\textwidth]{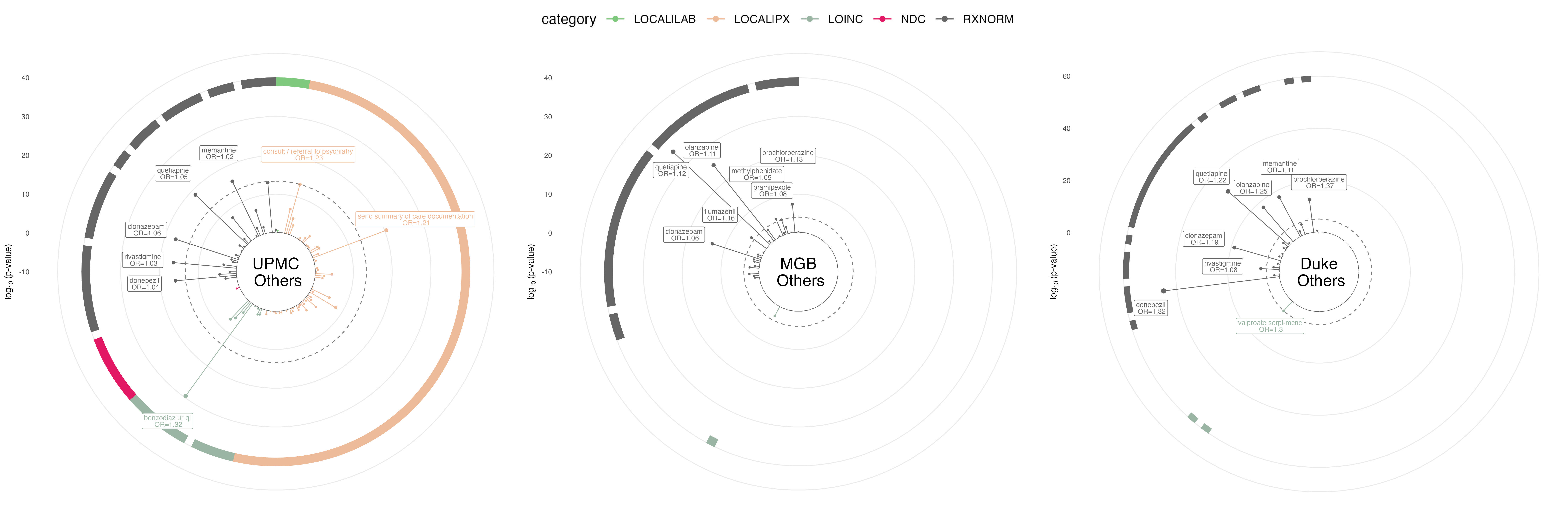}};
            \node[anchor=north west, xshift=5pt, yshift=-5pt] at (img.north west)
                {\textbf{(b)}};
        \end{tikzpicture}
    \end{minipage}

    \caption{For AD patient stratification, the association between AD-related EHR codes (top: diagnosis codes, bottom: other codes) and cluster membership (cluster 1 vs clsuter 2) in each institution (UPMC, MGB and Duke). Each dot/line represent a unqiue code, and the length shows $-\log_{10}(\text{p-value})$ of the association. 
    Panel (a) shows associations for diagnosis codes, and panel (b) shows associations for other code types. Top $12$ diagnosis features and top $8$ other features are labeled with the code description and odds ratio.  Source data are provided as a Source Data file.}
    \label{fig:AD_phewas}
\end{figure}

We report the demographics information of AD patients in each cluster at UPMC, MGB and Duke in Supplementary Table \ref{tab:AD_demo}. Visualizations of projected patient embeddings in each institution is presented in Supplementary Fig. \ref{fig:AD_tsne}. The Kaplan Meier curve of time to nusring home admission for each institution and cluster is shown in Supplementary Fig. \ref{fig:AD_KM}. Patients were effectively stratified into fast decline group (Cluster 1) and slow decline group (Cluster 2). The fast decline group had higher risk of nursing home admission compared to the slow decline group (median time to nursing home UPMC: Fast $37.1$ months, Slow $86.3$ months; MGB: Fast $43.9$ months, Slow $106.2$ months; Duke: Fast $94.0$ months, Slow not reached). Supplementary Fig. \ref{fig:AD_phewas} highlights the features driving the differences between the fast and slow decline groups in each institution.  The fast decline groups are characterized by higher intensity of mental disorders, including neurological disorders, mood disorders, depression, as well as more prescription of anti-psychotics such as quetiapine and olanzapine \cite{cipriani2011comparative, arvanitis1997multiple} across all institutions. This pattern further suggests that cognitive and psychiatric comorbidities may have compounded the presentation of AD in the fast decline group \cite{ismail2022psychosis}, leading to more rapid cognitive deterioration and a more challenging clinical trajectory. Abnormal movement, essential hypertension, and malaise and fatigue were also found to be associated with fast decline group at all institutions.

\begin{table}[H]
    \centering
    \scriptsize
    \resizebox{\textwidth}{!}{%
    \begin{tabular}{l c c c c c c c c c}
\toprule
& \multicolumn{3}{c}{{UPMC}}&
\multicolumn{3}{c}{{MGB}}&
\multicolumn{3}{c}{{Duke}} \\
\cmidrule(r){2-4}\cmidrule(l){5-7}\cmidrule(l){8-10}
& Overall & Fast (1) & Slow (2) & Overall & Fast (1) & Slow (2) & Overall & Fast (1) & Slow (2)  \\
\bottomrule
{Number of Patients, $N(\%)$} & 16411 & 6726 & 9685 &  17770 & 6089 & 11681 & 10660 & 3972 & 6688 \\
\bottomrule
{Age at AD diagnosis, mean (SD)} & 81.2 & 79.8 & 82.1 & 79.7 & 79.9 & 79.5 & 79.6 & 79.6 & 79.6 \\
 & (9.0) & (9.6) & (8.4) & (9.5) & (9.1) & (9.7) & (9.0) & (8.7) & (9.2)\\
\bottomrule
{Gender, $N(\%)$}\\
Women & 10559  & 4266  & 6293 & 10795 & 3644 & 7151 & 6760 & 2481 & 4279 \\
 & (64.3\%) & (63.4\%) & (65.0\%) & (60.7\%) & (59.8\%) & (61.2\%) & (63.4\%) & (62.5\%) & (64.0\%) \\
Men & 5852 & 2460 & 3392 & 6975 & 2445 & 4530 & 3900 & 1491 & 2409 \\
 & (35.7\%) & (36.6\%) & (35.0\%) & (39.3\%) & (40.2\%) & (38.8\%) & (36.6\%) & (37.5\%) & (36.0\%) \\
\bottomrule
{Race/Ethnicity, $N(\%)$} \\
Non-Hispanic White & 15105 & 6214 & 8891 & 15793 & 5324 & 10469 & 7512 & 2854 & 4658 \\
 & (92.0\%) & (92.4\%) & (91.8\%) & (88.9\%) & (87.4\%) & (89.6\%) & (70.5\%) & (71.9\%) & (69.6\%) \\
Others & 1306 & 512 & 794 & 1977 & 765 & 1212 & 3148 & 1118 & 2030 \\
 & (8.0\%) & (7.6\%) & (8.2\%) & (11.1\%) & (12.6\%) & (10.4\%) & (29.5\%) & (28.1\%) & (30.4\%) \\
\bottomrule
\end{tabular}
}
\caption{Demographics of AD patients in each cluster at UPMC, MGB, and Duke.}
\label{tab:AD_demo}
\end{table}

\begin{figure}[H]
    \centering
    \includegraphics[width=\textwidth]{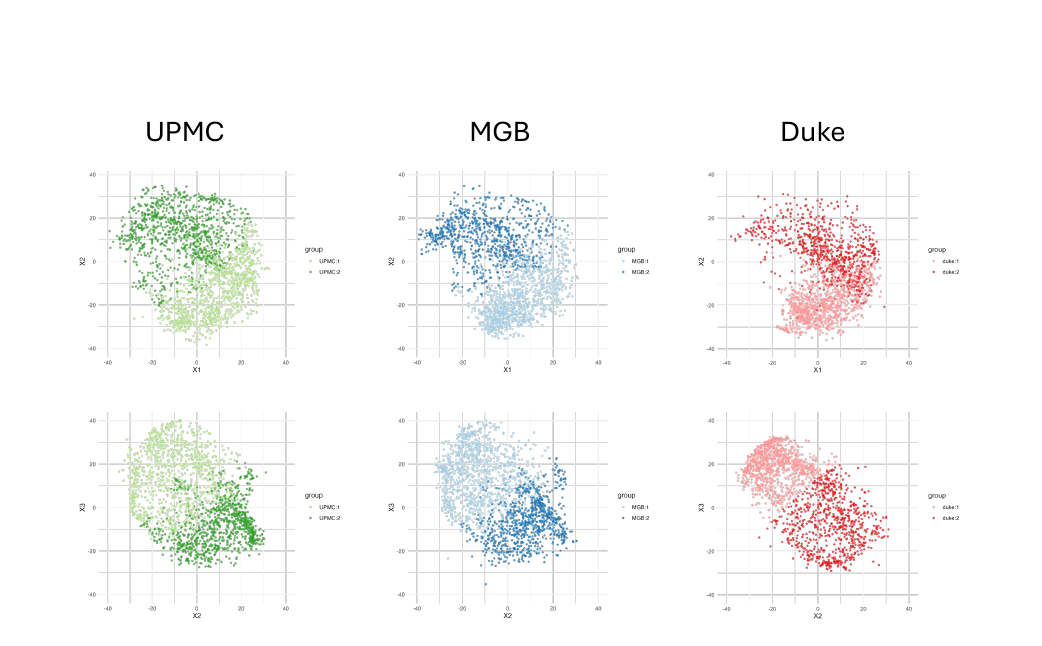}
    \caption{Projected t-SNE visualization of patient embedding for $5000$ randomly sampled AD patients at each insitution (UPMC, MGB and Duke). }
    \label{fig:AD_tsne}
\end{figure}

\begin{figure}[H]
    \centering
    \includegraphics[width=0.3\textwidth]{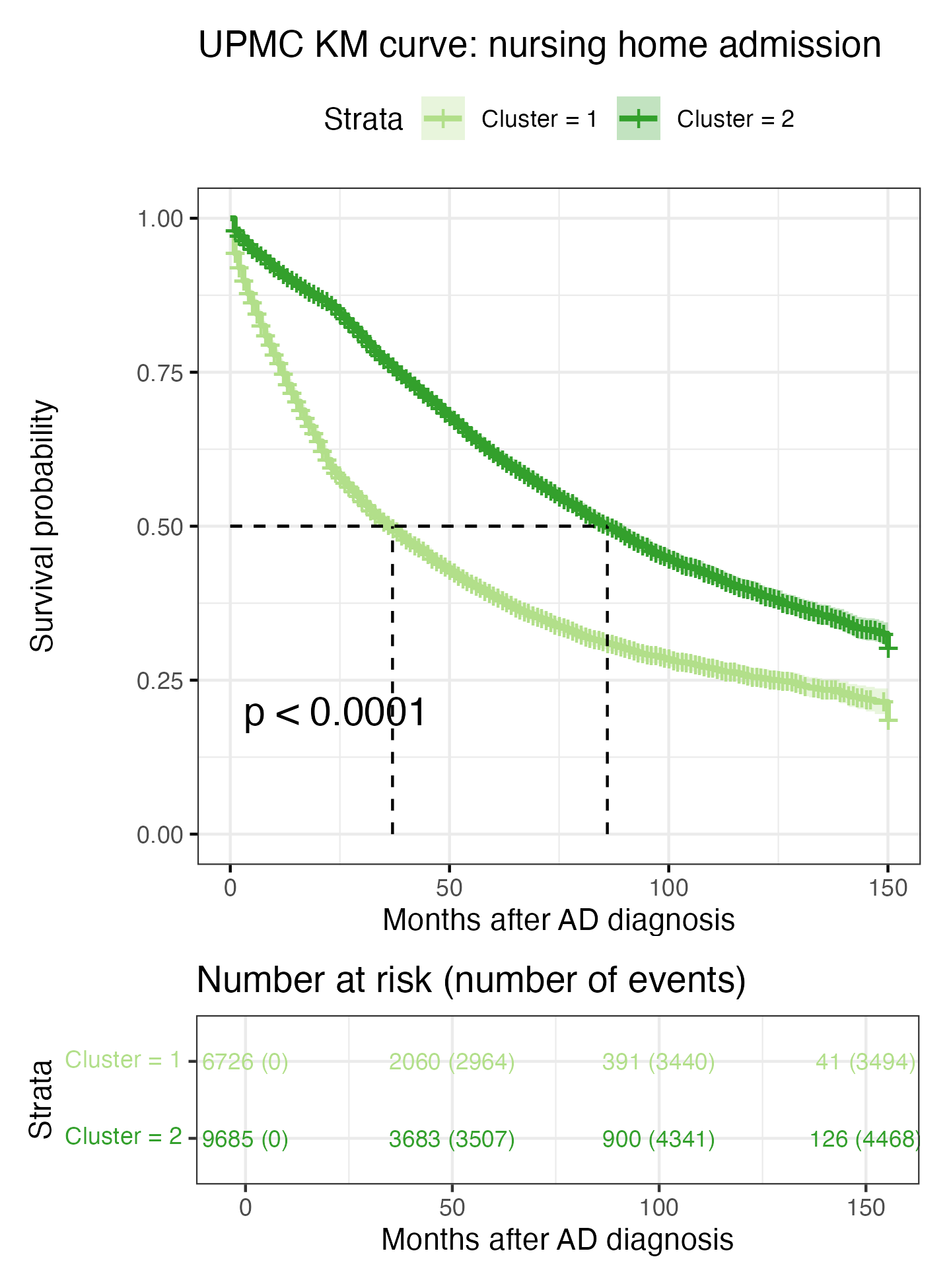}\hfill
    \includegraphics[width=0.3\textwidth]{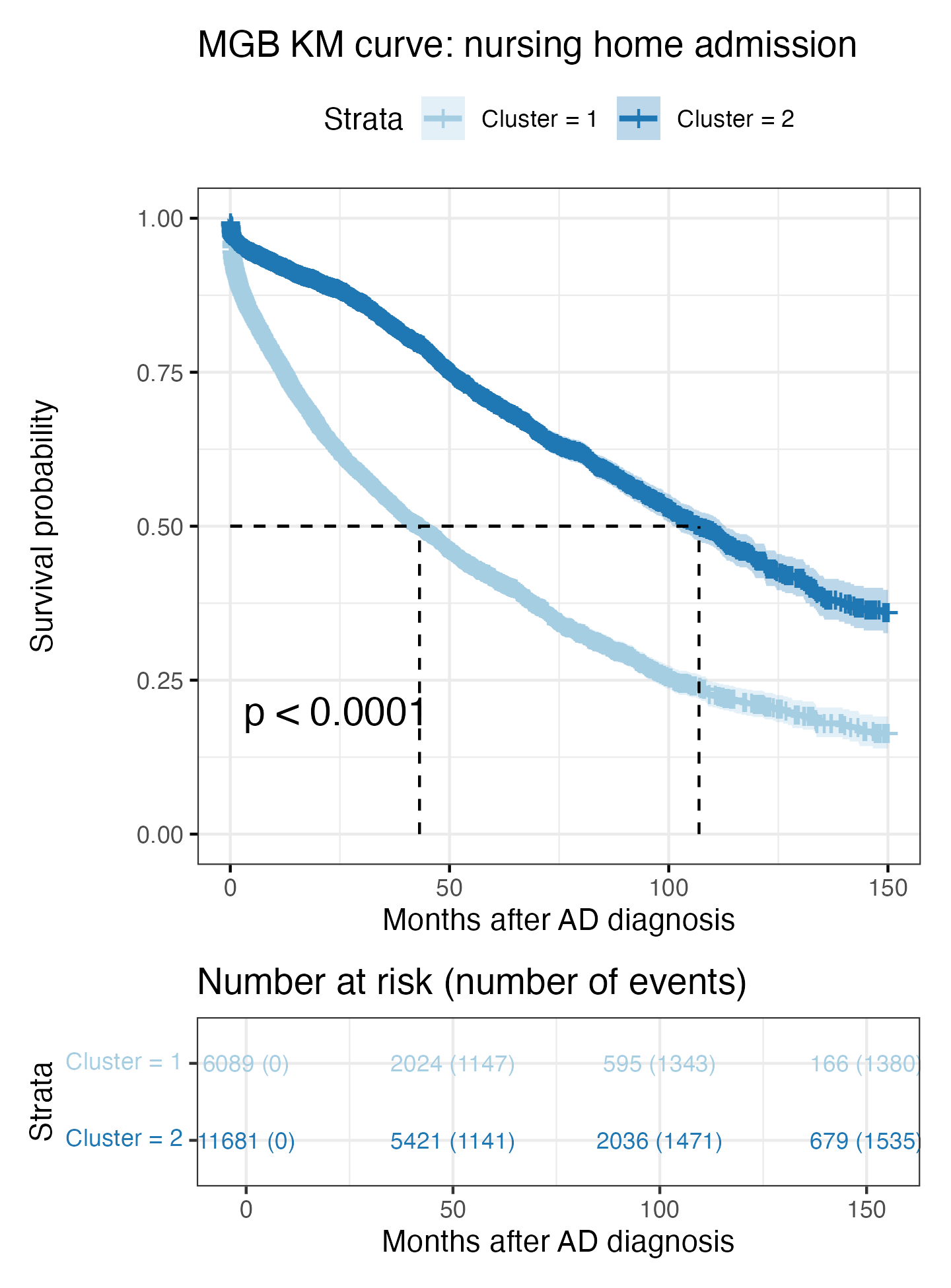}\hfill
    \includegraphics[width=0.3\textwidth]{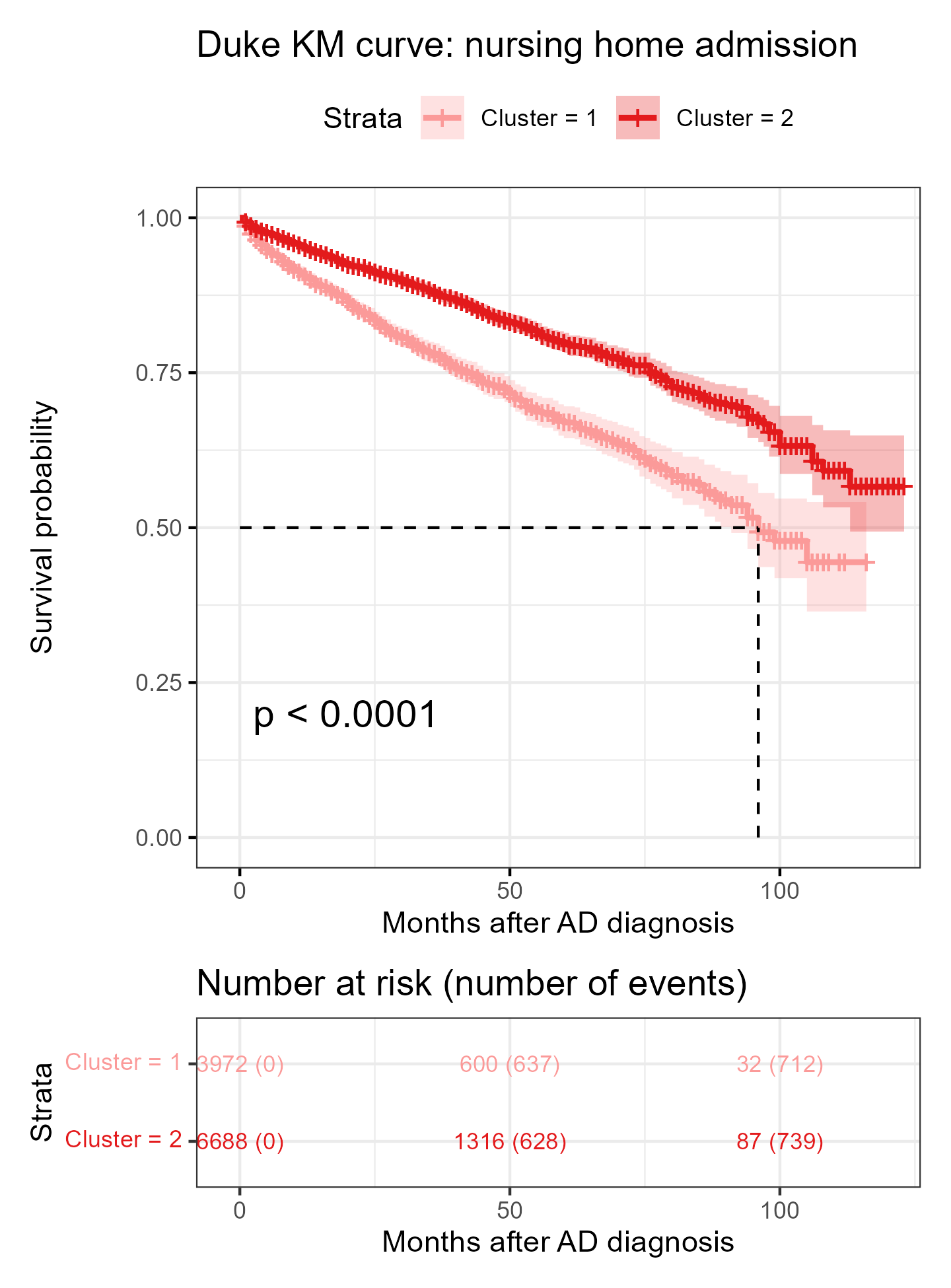}
    \caption{Kaplan Meier curve of time to nursing home admission for each cluster of AD patients in UPMC, MGB and Duke.}
    \label{fig:AD_KM}
\end{figure}

The prediction performance of models, trained using GAME embeddings and other benchmark methods, for nursing home admission is presented in Supplementary Fig.~\ref{fig:AD_AUC}. Compared to models trained on raw EHR features, those using GAME-derived embeddings achieved higher predictive accuracy within the local site and demonstrated greater transportability across institutions. 

\begin{figure}[H]
    \centering
    \includegraphics[width=\textwidth]{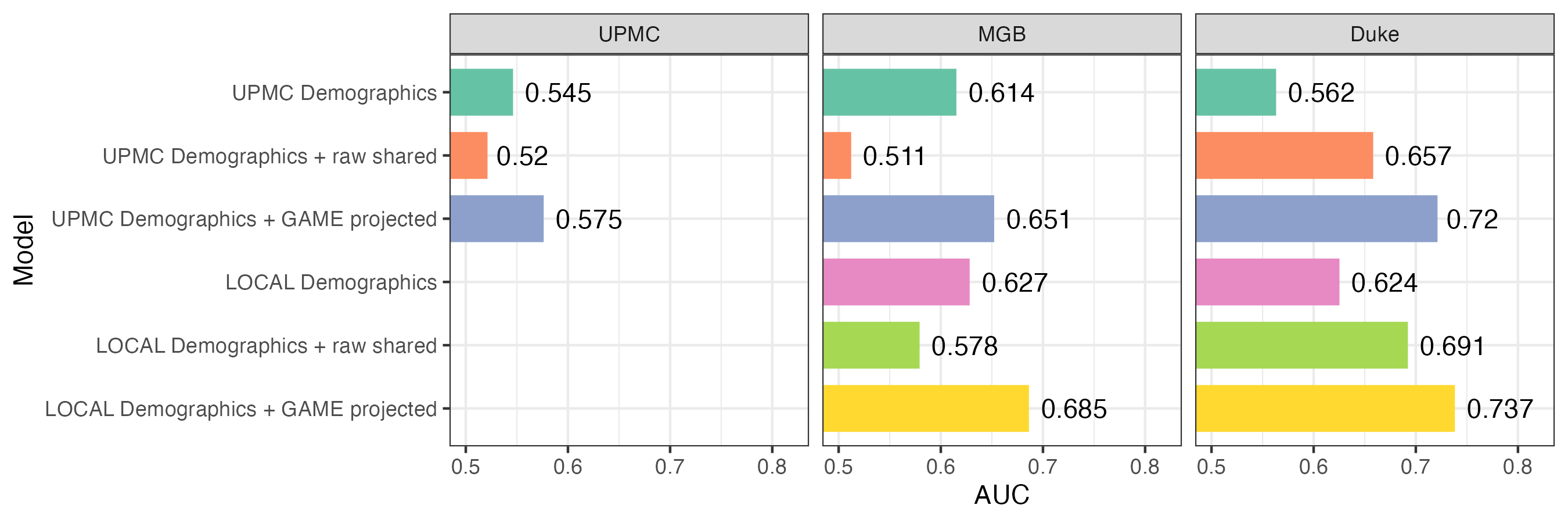}
    \caption{AUC for prediction of 1-year risk of nursing home admission for AD patients.  Source data are provided as a Source Data file.}
    \label{fig:AD_AUC}
\end{figure}

\subsubsection{Suicide risk assessment}

We clustered patients with mental health disorders based on their EHR profiles from the two years following first diagnosis. Phecodes used for selection of mental health patients is shown in Supplementary Table~\ref{tab:mental_phecode}. Given that suicide risk varies significantly by age \cite{fazel2020suicide}, we stratified patients into five age groups: $\text{age} < 18$, $18 \le \text{age} \le 25$, $26 \le \text{age} \le 49$, $50 \le \text{age} \le 65$, and $\text{age} > 65$, and performed clustering and analysis separately for each group. Demographic information of the mental health cohorts at MGB and Duke is detailed in Supplementary Table \ref{tab:mental_demo_consolidated} and patient embeddings from both institutions are visualized in Supplementary Fig. \ref{fig:suicide_tsne}. Patients in the same age group from different institutions were projected to the same embedding space, which enabled cross-institutional co-clustering.

\begin{table}[H]
\centering
\begin{tabular}{ll}
  \toprule
PheCode & Description \\ 
  \midrule
PheCode:295 & schizophrenia and other psychotic disorders \\ 
  PheCode:295.1 & schizophrenia \\ 
  PheCode:295.2 & paranoid disorders \\ 
  PheCode:295.3 & psychosis \\ 
  PheCode:296 & mood disorders \\ 
  PheCode:296.1 & bipolar \\ 
  PheCode:296.2 & depression \\ 
  PheCode:296.22 & major depressive disorder \\ 
  PheCode:300 & anxiety, phobic and dissociative disorders \\ 
  PheCode:300.1 & anxiety disorder \\ 
  PheCode:300.11 & generalized anxiety disorder \\ 
  PheCode:300.12 & agorophobia, social phobia, and panic disorder \\ 
  PheCode:300.13 & phobia \\ 
  PheCode:300.3 & obsessive-compulsive disorders \\ 
  PheCode:300.4 & dysthymic disorder \\ 
  PheCode:300.8 & acute reaction to stress \\ 
  PheCode:300.9 & posttraumatic stress disorder \\ 
  PheCode:301 & personality disorders \\ 
  PheCode:301.1 & schizoid personality disorder \\ 
  PheCode:301.2 & antisocial/borderline personality disorder \\ 
  PheCode:302 & sexual and gender identity disorders \\ 
  PheCode:302.1 & decreased libido \\ 
  PheCode:303 & psychogenic and somatoform disorders \\ 
  PheCode:303.1 & dissociative disorder \\ 
  PheCode:303.3 & psychogenic disorder \\ 
  PheCode:303.31 & gastrointestinal malfunction arising from mental factors \\ 
  PheCode:303.4 & somatoform disorder \\ 
  PheCode:304 & adjustment reaction \\ 
  PheCode:305.2 & eating disorder \\ 
  PheCode:305.21 & anorexia nervosa \\ 
   \bottomrule
\end{tabular}
\caption{Phecodes indicating mental health disorders. }
\label{tab:mental_phecode}
\end{table}

\begin{table}[H]
    \centering
    \scriptsize
    \begin{tabular}{l c c c c c c}
    \toprule
    & \multicolumn{3}{c}{{MGB}} & \multicolumn{3}{c}{{Duke}} \\
    \cmidrule(r){2-4} \cmidrule(l){5-7}
    & Overall & Higher risk & Lower risk & Overall & Higher risk & Lower risk \\
    \midrule
    
    {Age < 18} \\
    Number of Patients & 24,048 & 9,070 & 14,978 & 31,576 & 10,589 & 20,987 \\
    Age at AD diagnosis, mean (SD) & 12.3 (4.0) & 13.0 (3.5) & 11.9 (4.2) & 10.8 (4.6) & 10.9 (4.4) & 10.8 (4.6) \\
    Women (\%) & 59.0 & 57.5 & 60.0 & 52.0 & 47.7 & 54.2 \\
    Non-Hispanic White (\%) & 69.0 & 76.7 & 64.3 & 50.7 & 58.8 & 46.6 \\
    \midrule
    
    {Age 19-25} \\
    Number of Patients & 36,666 & 16,527 & 20,139 & 23,271 & 13,591 & 9,680 \\
    Age at AD diagnosis, mean (SD) & 21.8 (2.2) & 21.7 (2.2) & 21.8 (2.2) & 21.6 (2.3) & 21.7 (2.3) & 21.5 (2.4) \\
    Women (\%) & 70.9 & 67.4 & 73.8 & 66.4 & 69.8 & 61.6 \\
    Non-Hispanic White (\%) & 73.4 & 76.3 & 71.0 & 59.4 & 64.5 & 52.1 \\
    \midrule
    
    {Age 26-49} \\
    Number of Patients & 123,034 & 59,938 & 63,096 & 94,846 & 63,501 & 31,345 \\
    Age at AD diagnosis, mean (SD) & 37.5 (6.9) & 37.6 (7.0) & 37.4 (6.9) & 37.9 (6.9) & 38.2 (6.9) & 37.3 (6.9) \\
    Women (\%) & 69.0 & 65.6 & 72.2 & 66.0 & 68.1 & 61.8 \\
    Non-Hispanic White (\%) & 23.0 & 22.3 & 23.6 & 63.7 & 65.0 & 61.1 \\
    \midrule
    
    {Age 50-65} \\
    Number of Patients & 72,482 & 37,230 & 35,252 & 70,651 & 47,152 & 23,499 \\
    Age at AD diagnosis, mean (SD) & 56.8 (4.5) & 56.9 (4.5) & 56.8 (4.2) & 57.4 (4.6) & 57.6 (4.6) & 57.0 (4.5) \\
    Women (\%) & 65.0 & 62.8 & 67.4 & 64.2 & 64.7 & 63.2 \\
    Non-Hispanic White (\%) & 85.1 & 84.7 & 85.5 & 70.7 & 69.6 & 72.9 \\
    \midrule
    
    {Age $\geq$ 66} \\
    Number of Patients & 39,509 & 19,887 & 19,622 & 51,867 & 43,157 & 8,710 \\
    Age at AD diagnosis, mean (SD) & 74.0 (6.4) & 74.2 (6.5) & 73.8 (6.3) & 73.8 (6.5) & 74.0 (6.5) & 72.7 (6.1) \\
    Women (\%) & 66.4 & 64.9 & 67.9 & 66.0 & 65.9 & 66.8 \\
    Non-Hispanic White (\%) & 89.1 & 89.1 & 89.0 & 79.4 & 78.6 & 83.1 \\
    \bottomrule
    \end{tabular}
\caption{Demographics of mental health patients by age group and risk cluster at MGB and Duke. Percentages are shown for Women and Non-Hispanic White categories. Complementary categories (Men and Other race/ethnicity) are omitted for simplicity as percentages sum to 1.}
    \label{tab:mental_demo_consolidated}
\end{table}
\begin{figure}[H]
    \centering
    \includegraphics[width=\textwidth]{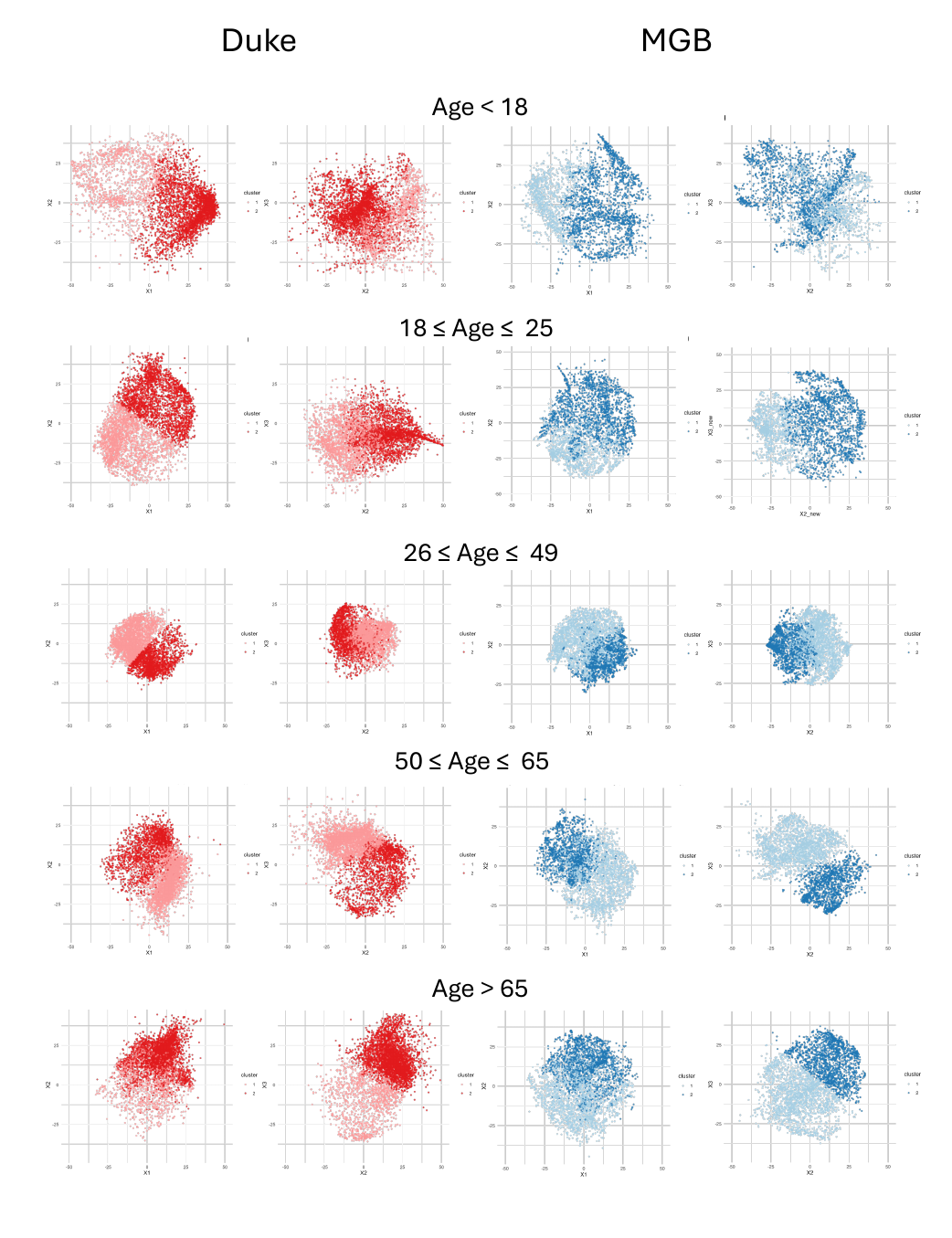}
    \caption{Projected t-SNE plots for $5000$ randomly samples mental health patients in each age group at each institution (Duke and MGB).}
    \label{fig:suicide_tsne}
\end{figure}

The KM curve of time to suicidal ideation and time to suicidal attempt are shown in Supplementary Fig. \ref{fig:KM_rpdr}. The subgroups had differential risk of future suicidal ideation across all age groups after adjusting for age, gender and race/ethnicity.
Key diagnostic and medication/procedure codes driving the difference between the clusters within each age group at MGB and Duke are shown in Supplementary Figs. \ref{fig:suicide_phewas_phe} and \ref{fig:suicide_phewas_oth}. Differences were primarily characterized by mental health conditions, such as mood disorders, major depressive disorder, and generalized anxiety disorder. In older age groups, conditions such as insomnia, essential hypertension, GERD, and symptoms like joint/back pain, nausea, vomiting, and fatigue also emerged as important differentiators between clusters.
\begin{figure}[H]
    \centering

    \begin{tikzpicture}
        \node[anchor=south west, inner sep=0] (img) at (0,0)
            {\includegraphics[width=3cm]{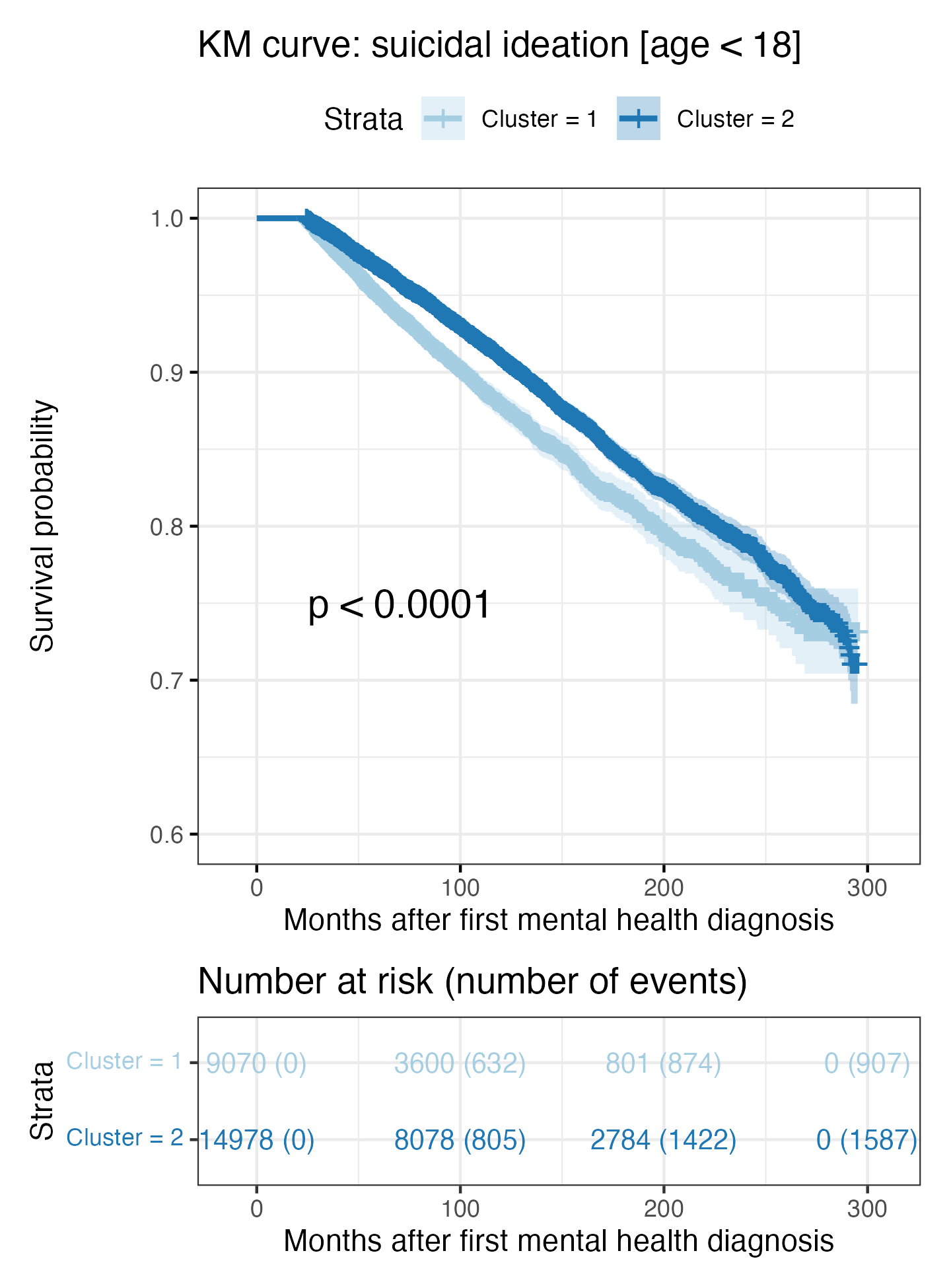}
             \includegraphics[width=3cm]{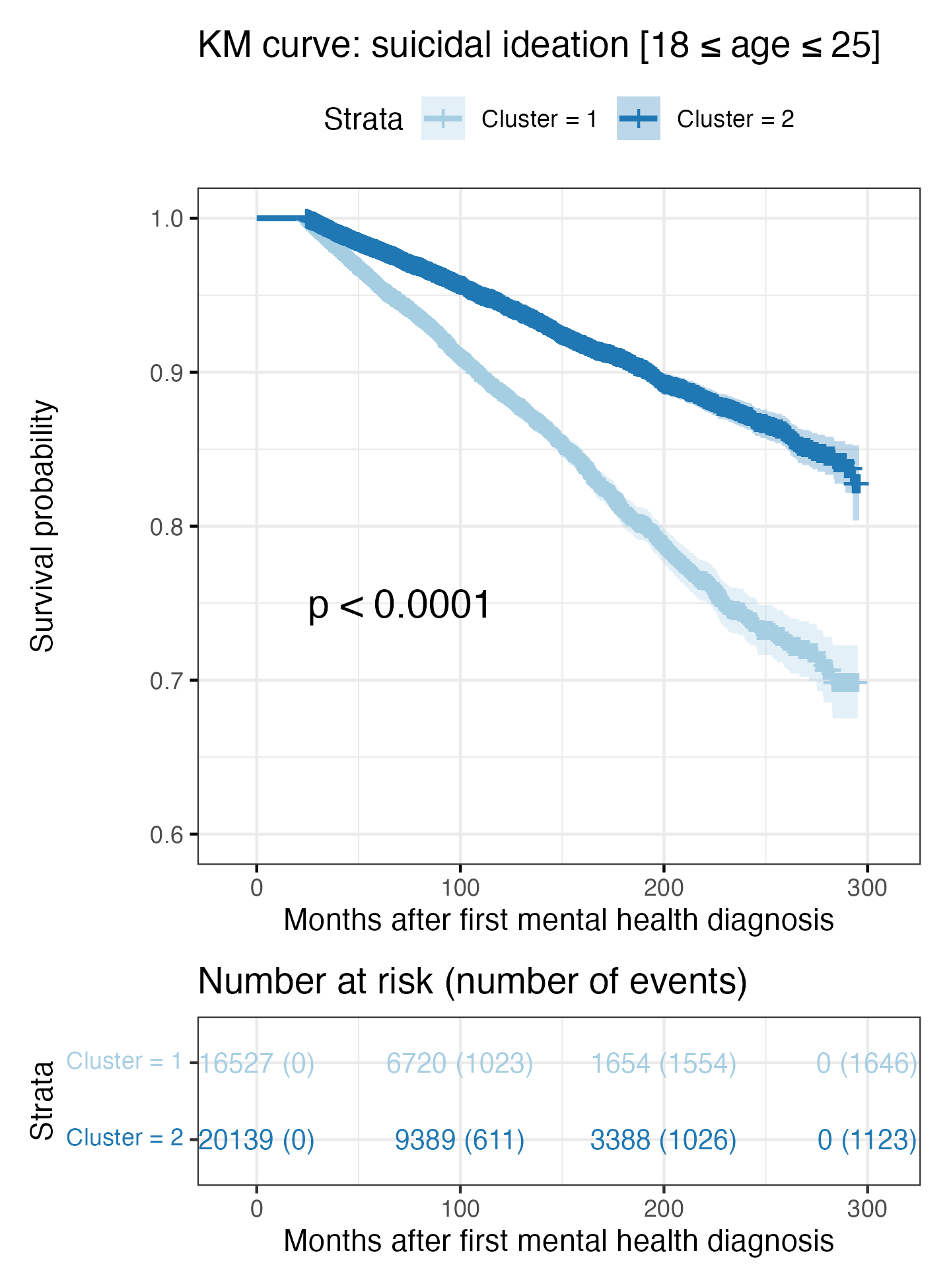}
             \includegraphics[width=3cm]{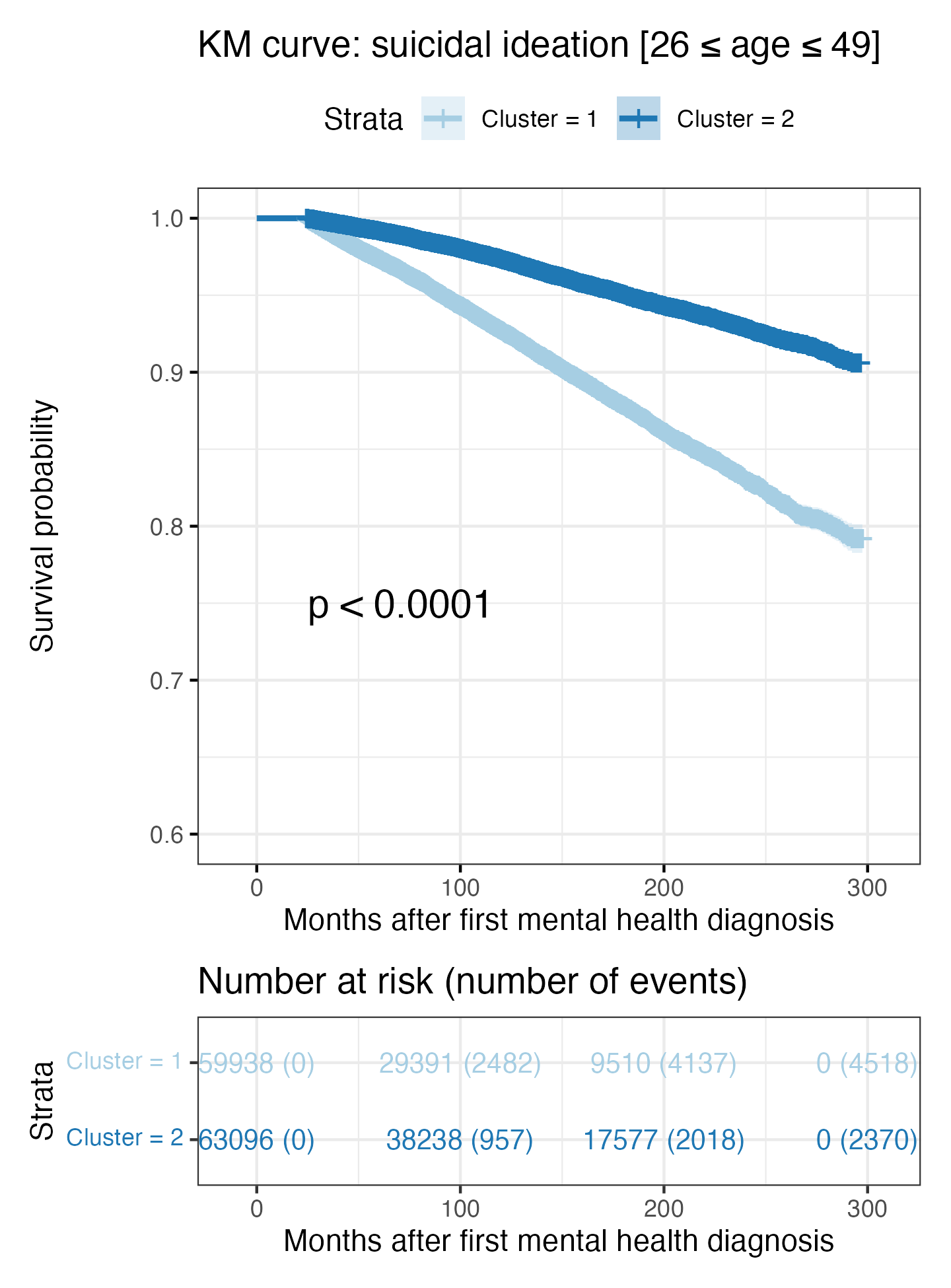}
             \includegraphics[width=3cm]{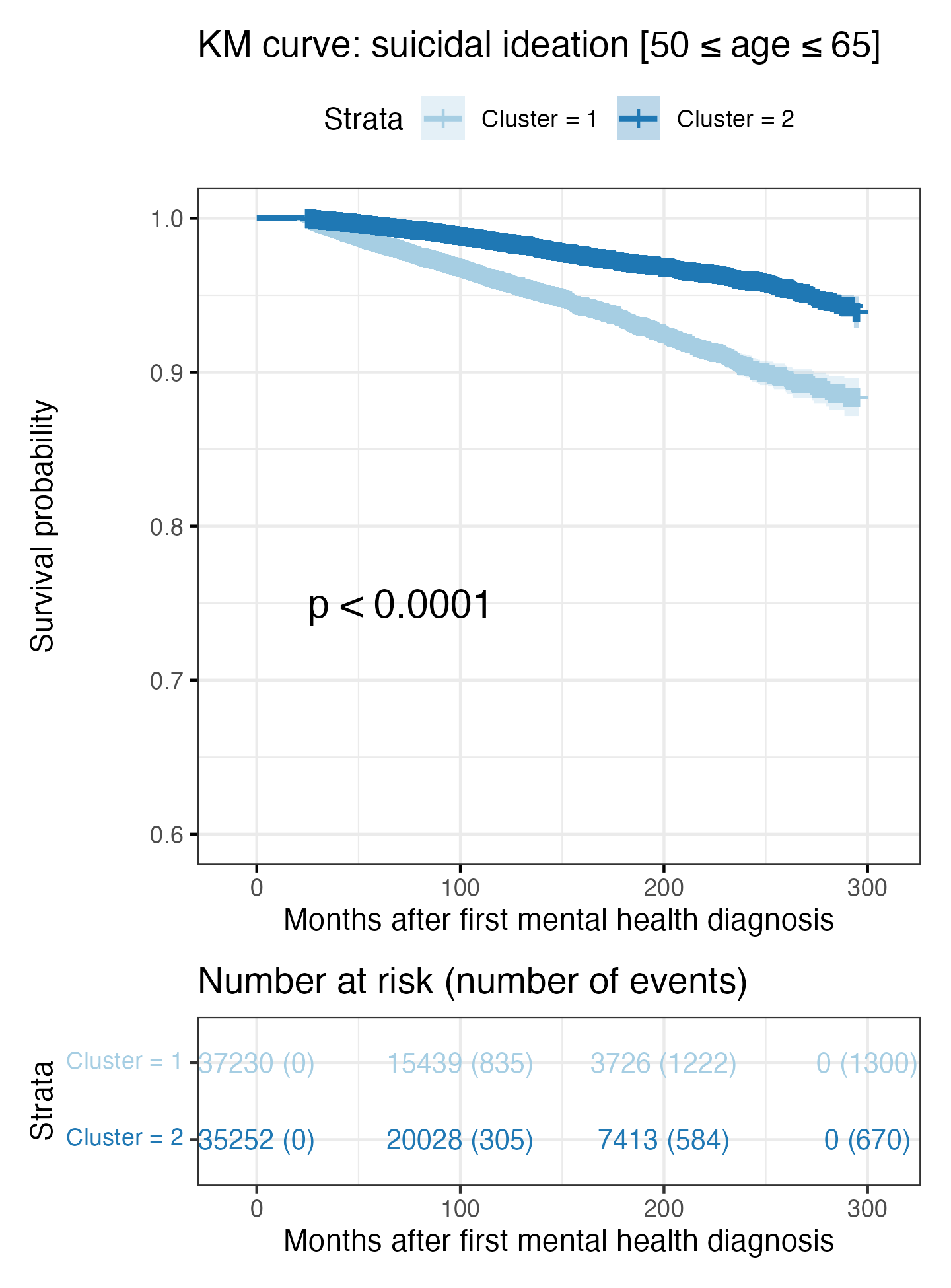}
             \includegraphics[width=3cm]{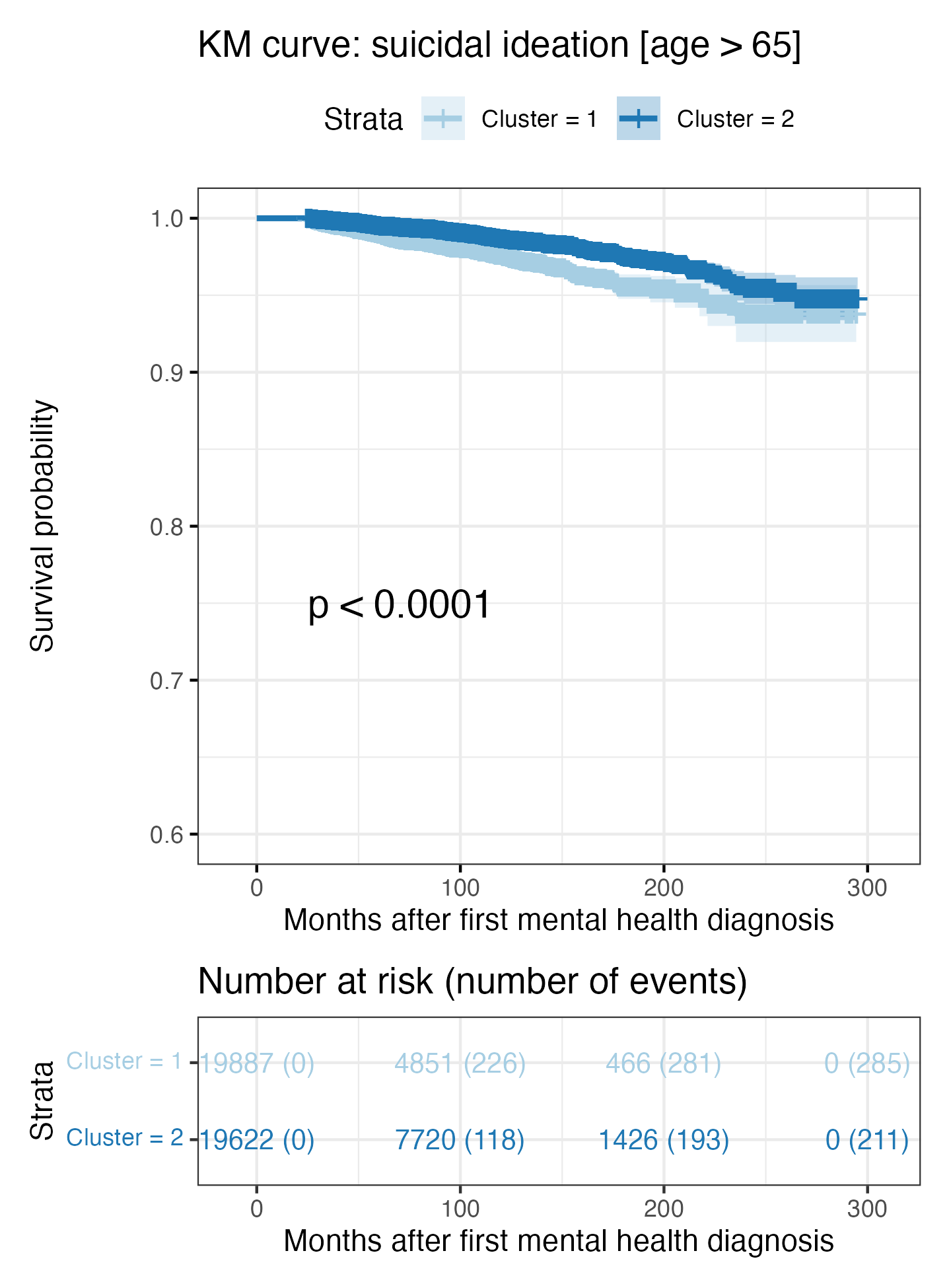}};
        \node[anchor=south west, xshift=3pt, yshift=8pt, fill=white, text opacity=1] 
            at (img.north west) {\textbf{(a)}};
    \end{tikzpicture}

    \vspace{0.5pt}

    \begin{tikzpicture}
        \node[anchor=south west, inner sep=0] (img) at (0,0)
            {\includegraphics[width=3cm]{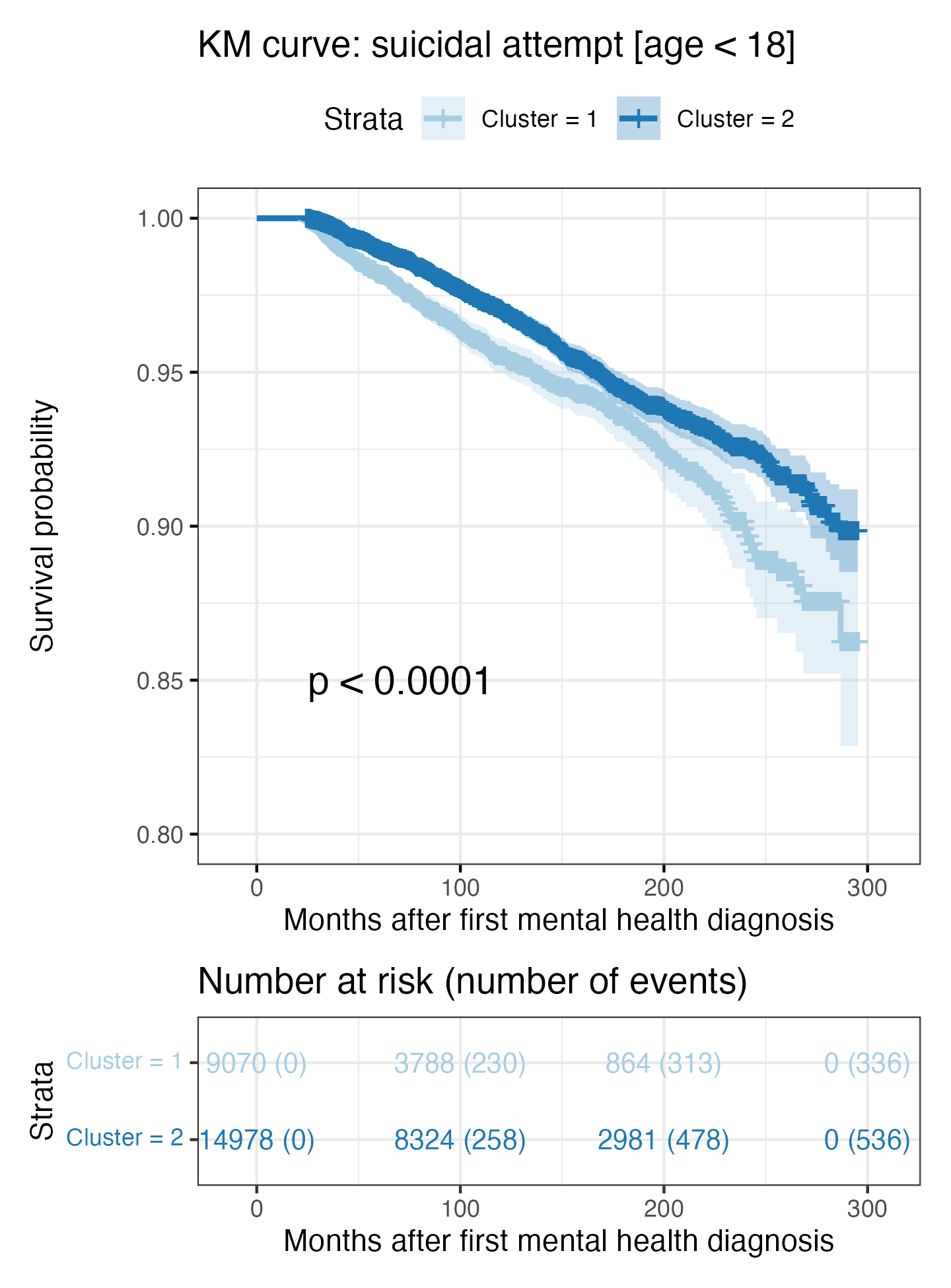}
             \includegraphics[width=3cm]{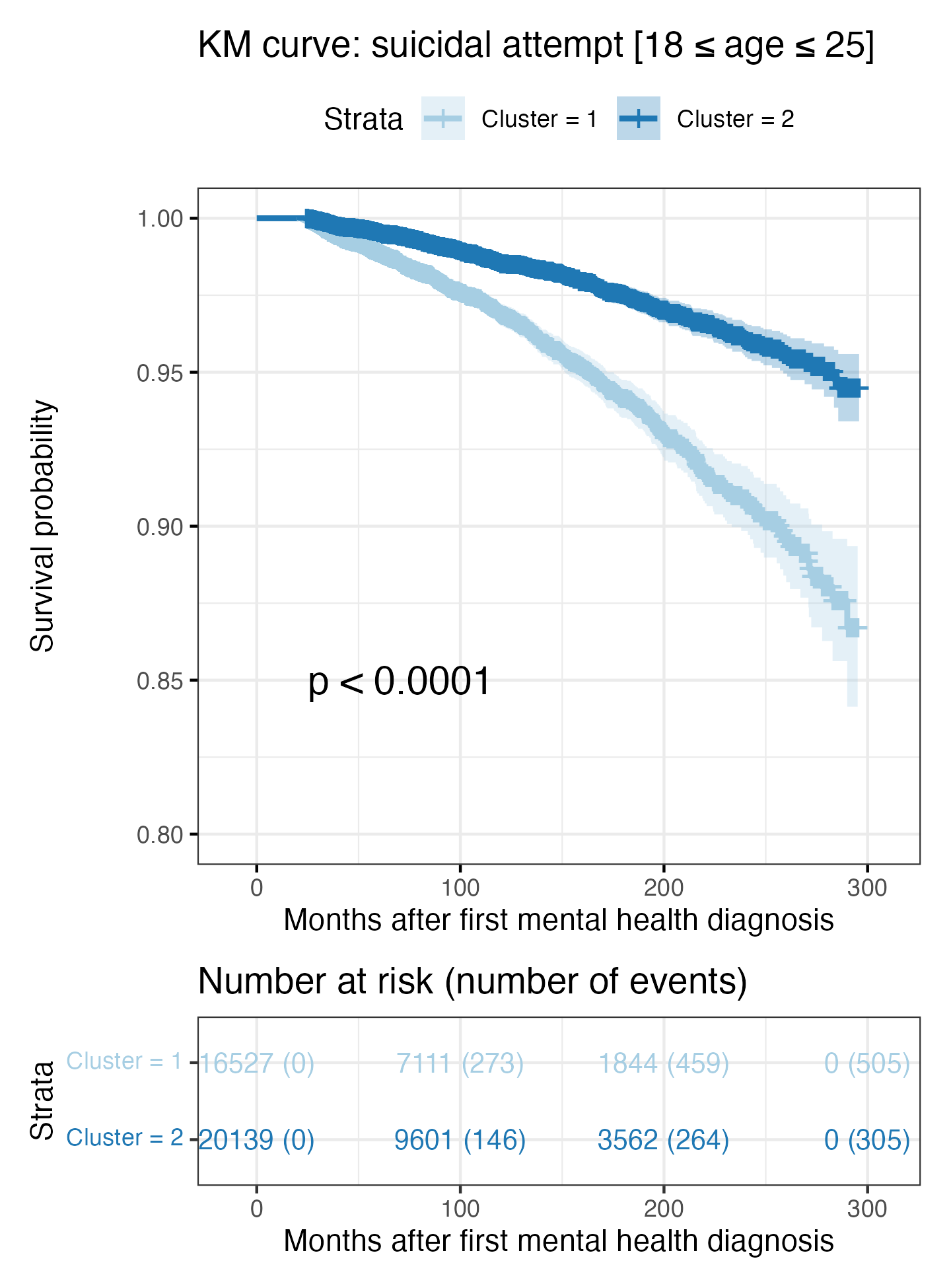}
             \includegraphics[width=3cm]{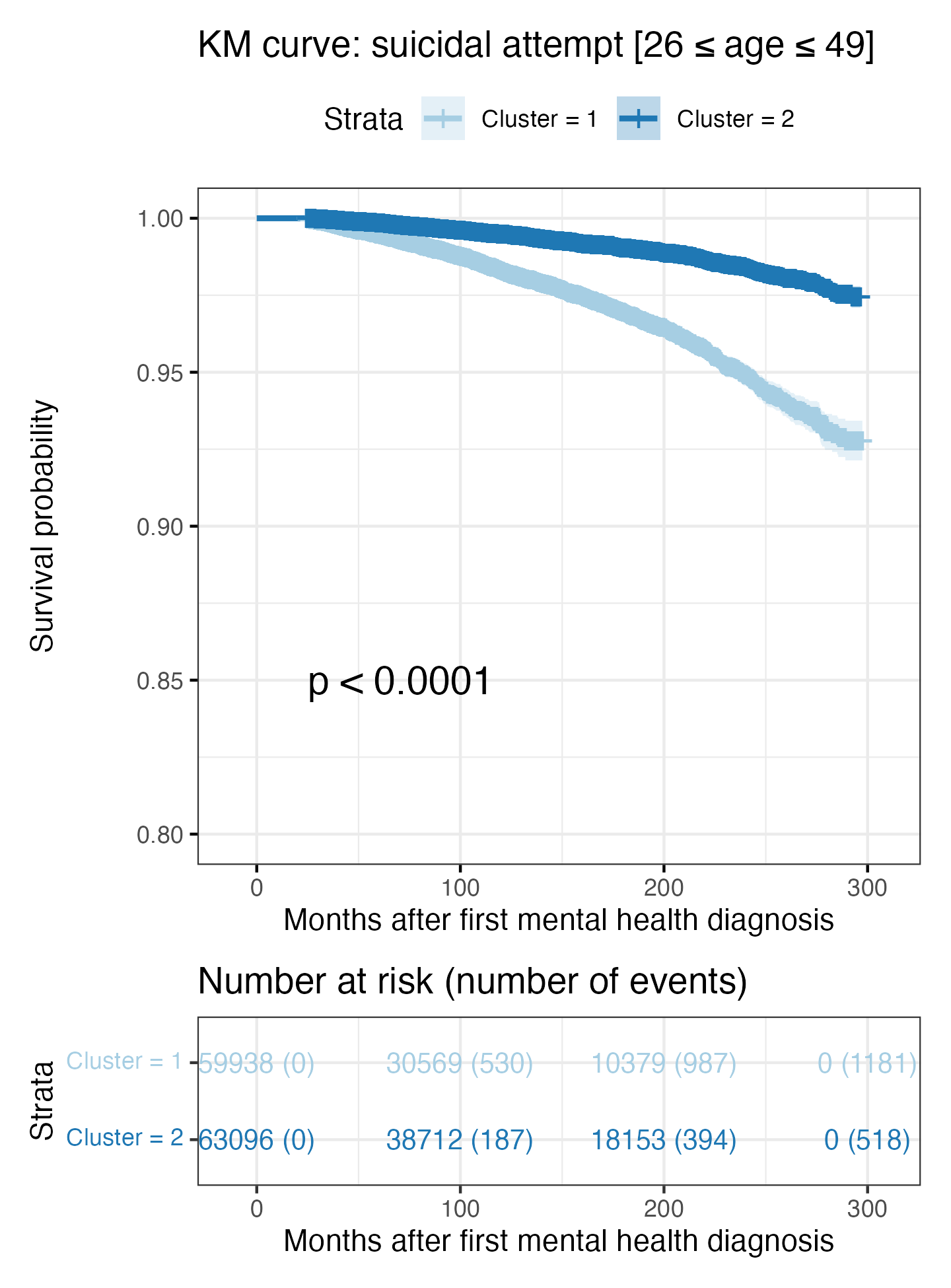}
             \includegraphics[width=3cm]{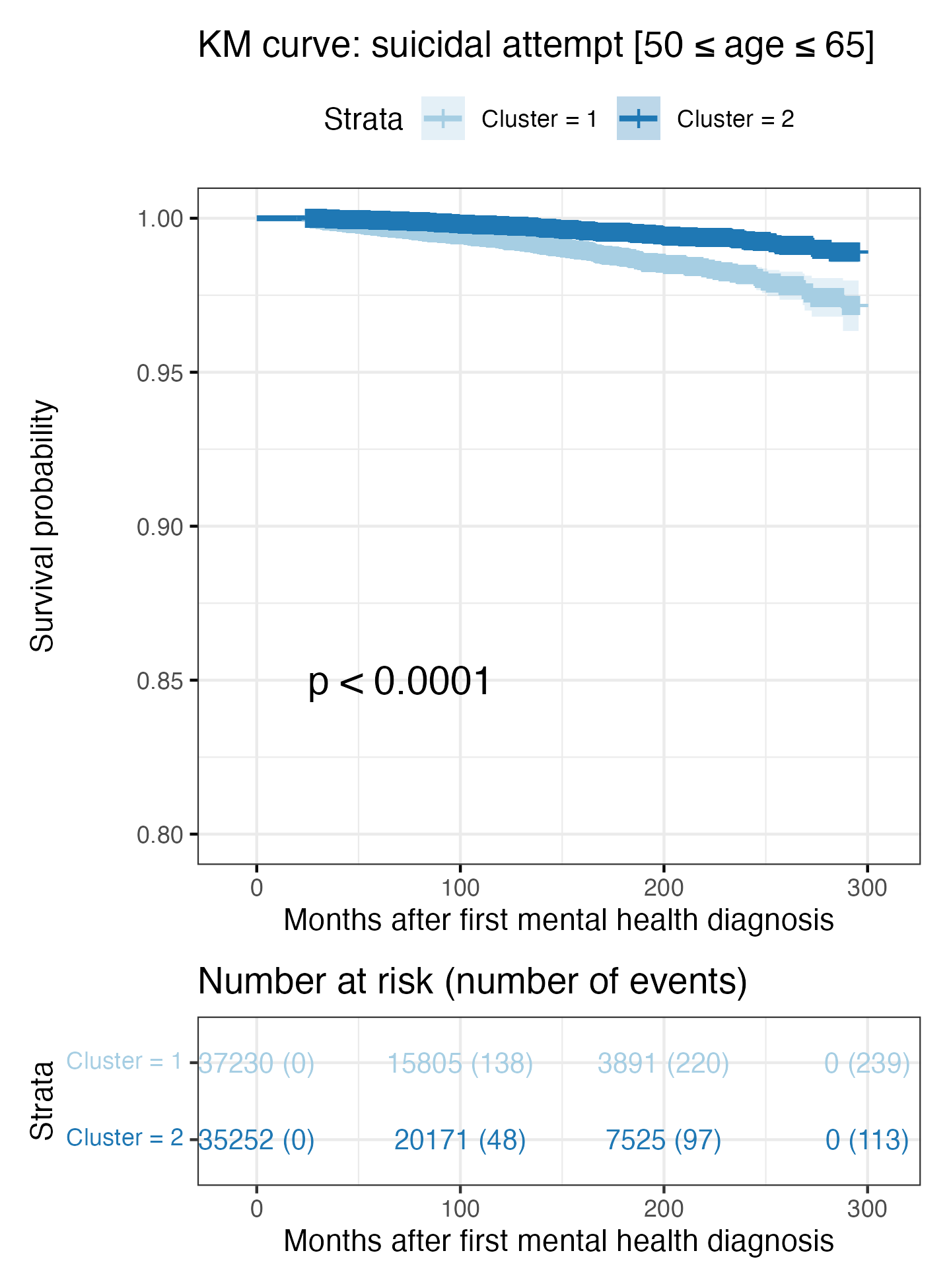}
             \includegraphics[width=3cm]{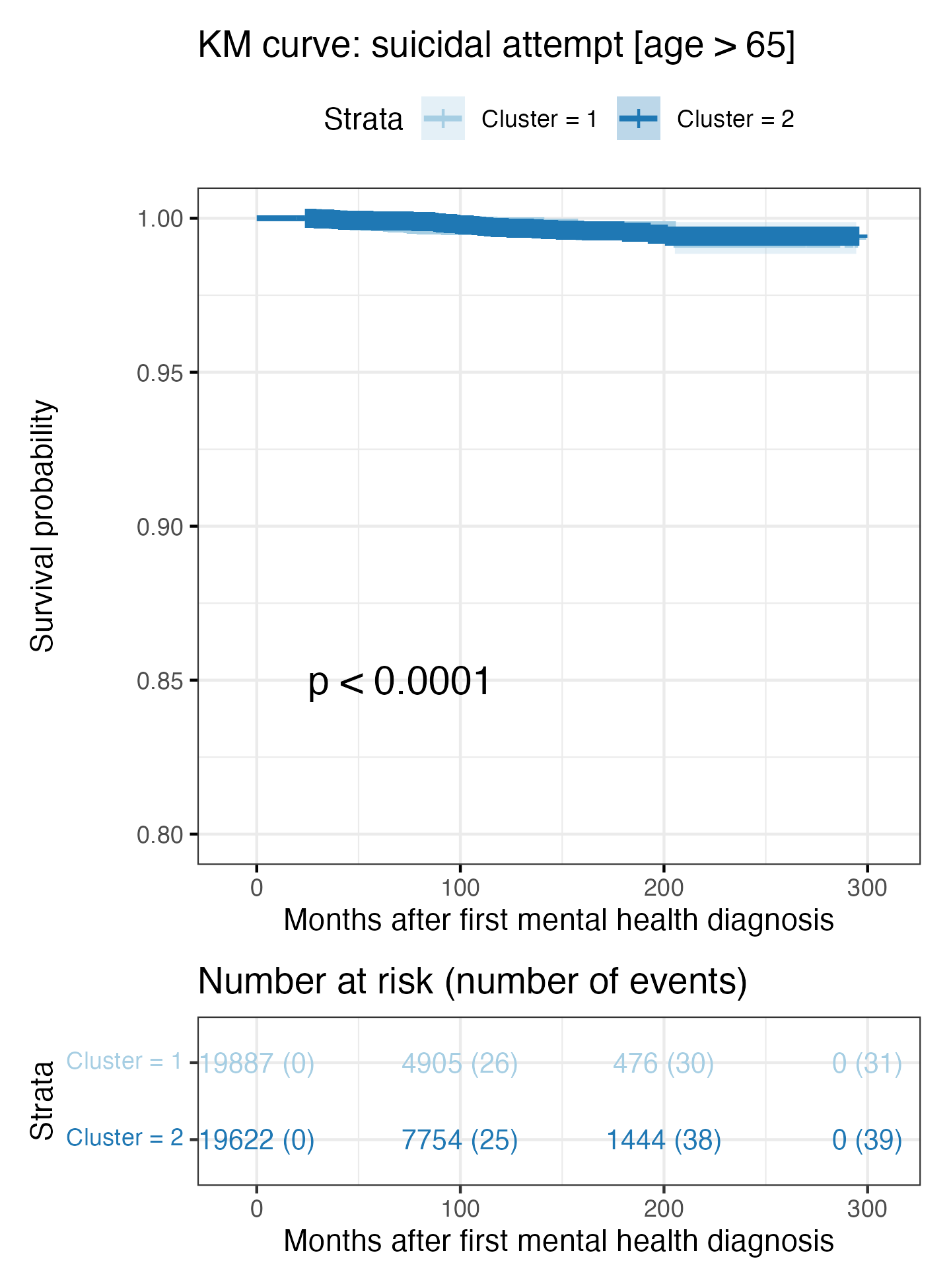}};
        \node[anchor=south west, xshift=3pt, yshift=8pt, fill=white, text opacity=1] 
            at (img.north west) {\textbf{(b)}};
    \end{tikzpicture}

    \vspace{0.5pt}

    \begin{tikzpicture}
        \node[anchor=south west, inner sep=0] (img) at (0,0)
            {\includegraphics[width=3cm]{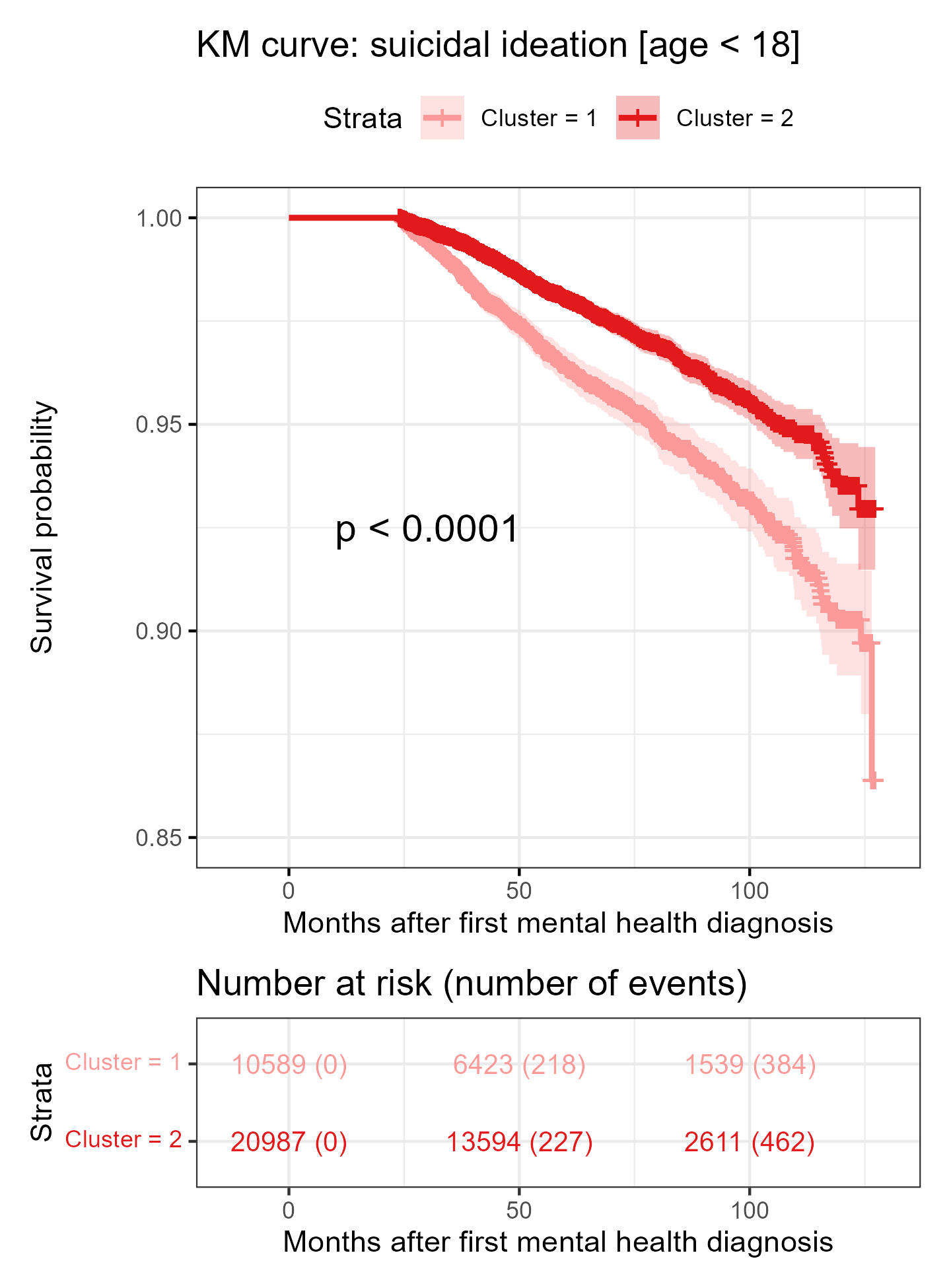}
             \includegraphics[width=3cm]{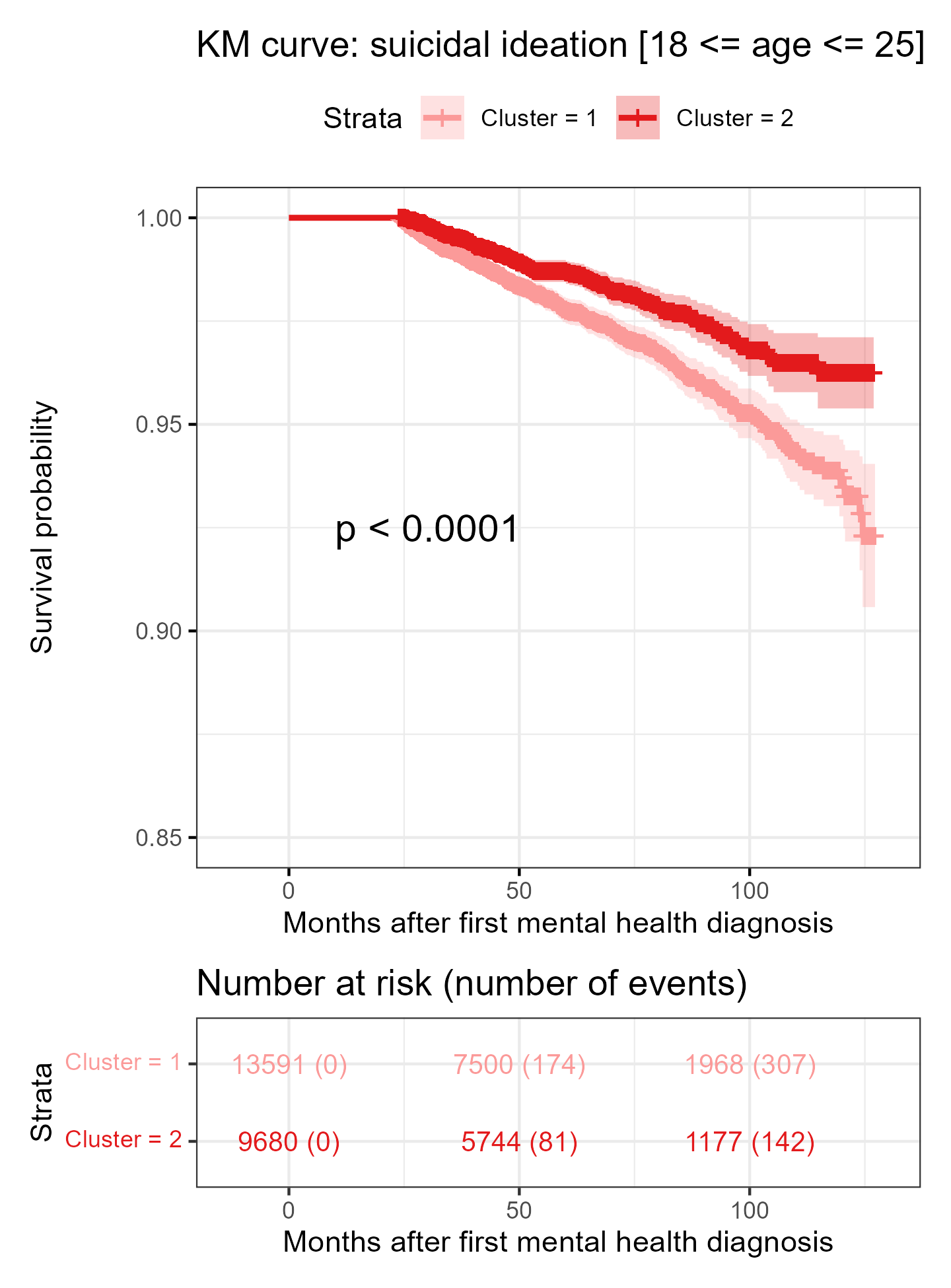}
             \includegraphics[width=3cm]{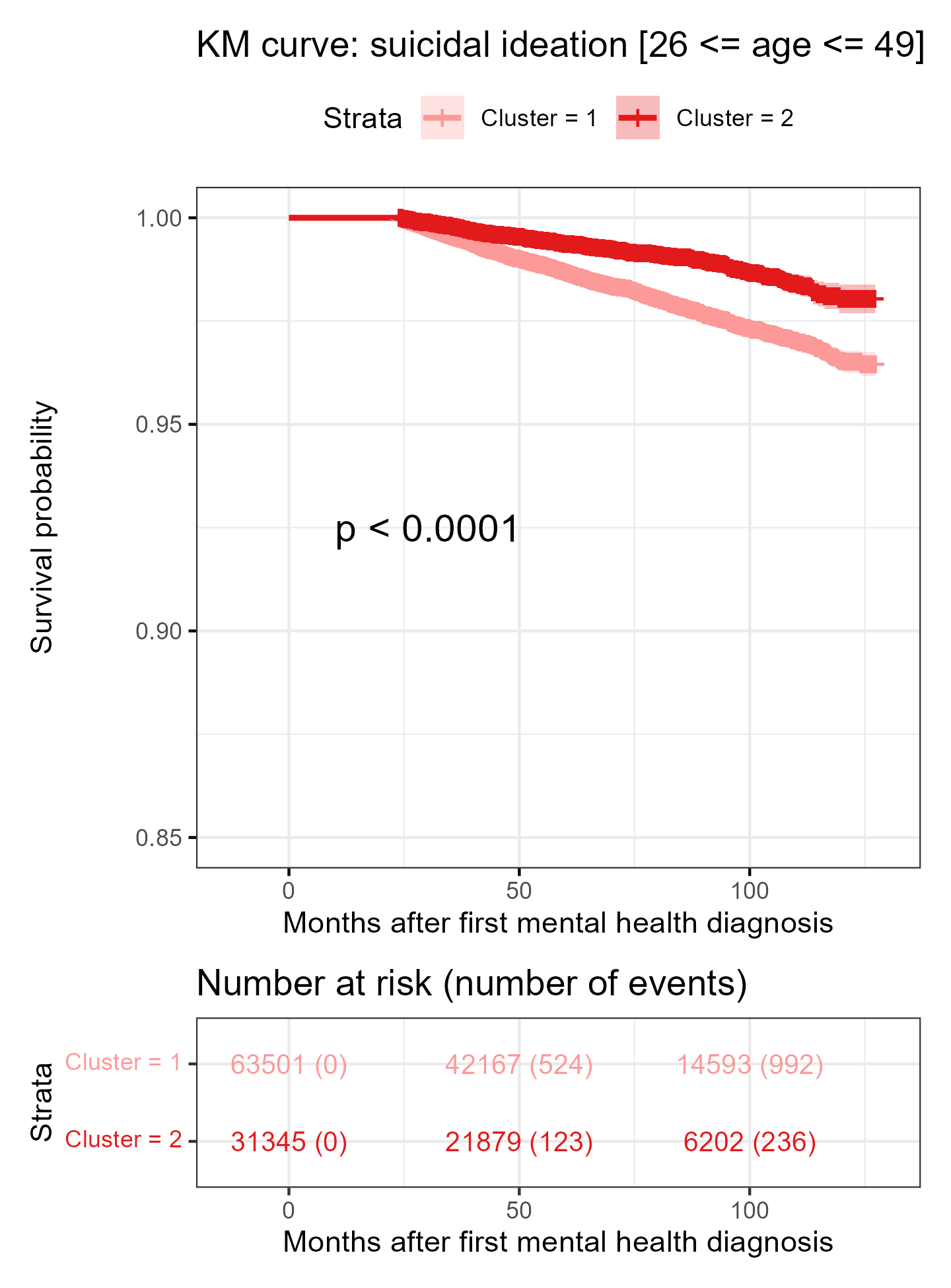}
             \includegraphics[width=3cm]{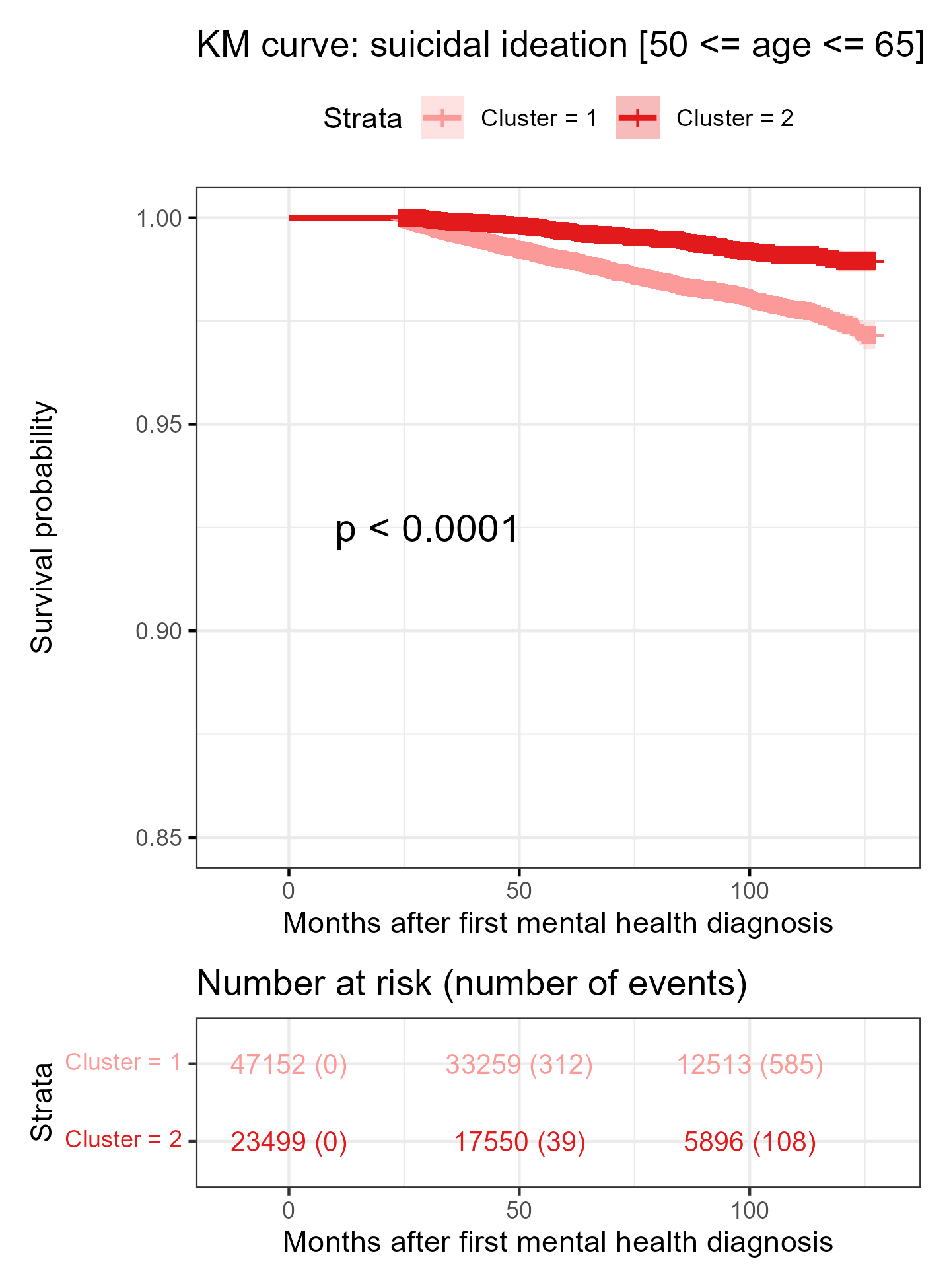}
             \includegraphics[width=3cm]{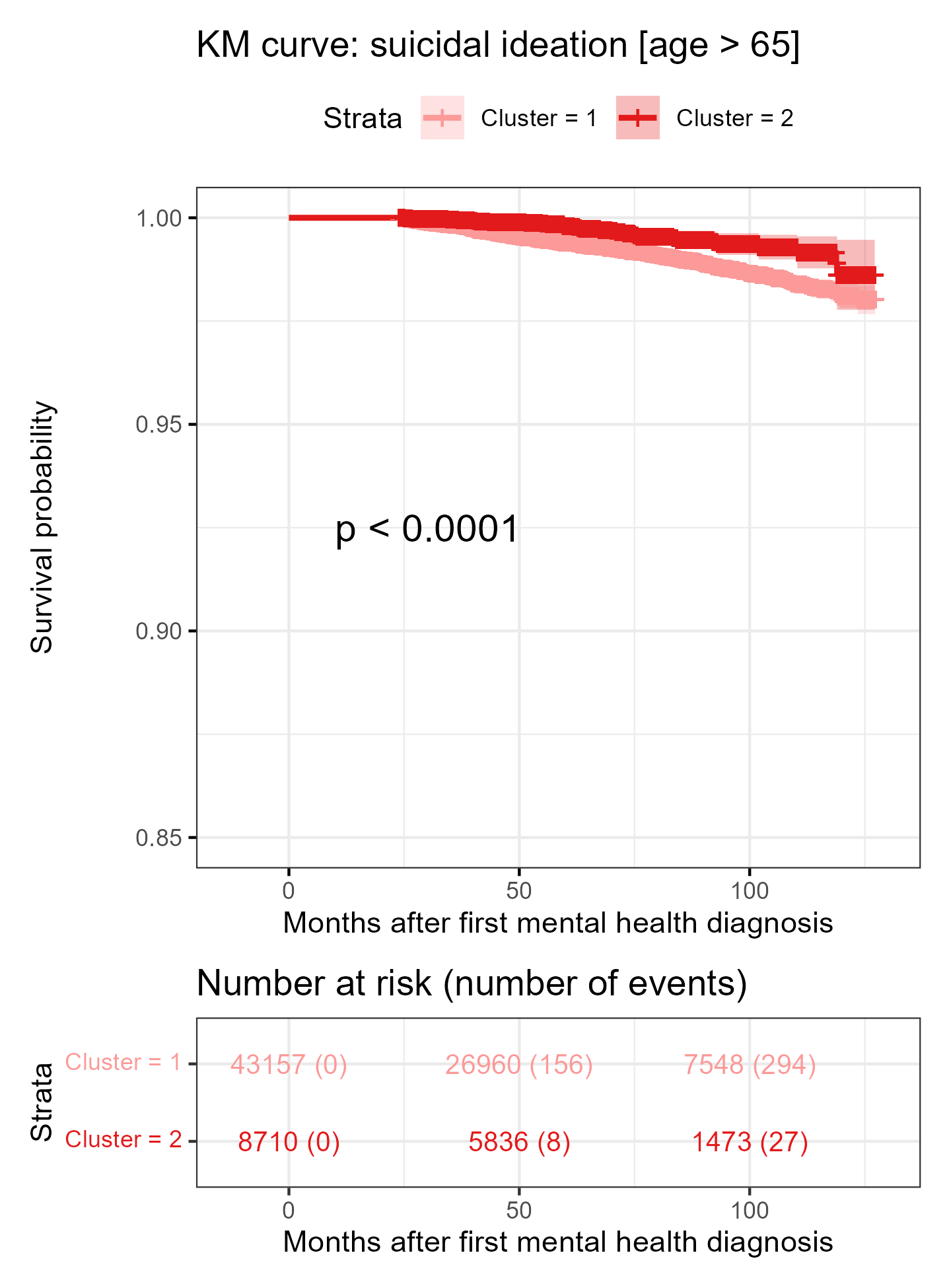}};
        \node[anchor=south west, xshift=3pt, yshift=8pt, fill=white, text opacity=1] 
            at (img.north west) {\textbf{(c)}};
    \end{tikzpicture}

    \vspace{0.5pt}

    \begin{tikzpicture}
        \node[anchor=south west, inner sep=0] (img) at (0,0)
            {\includegraphics[width=3cm]{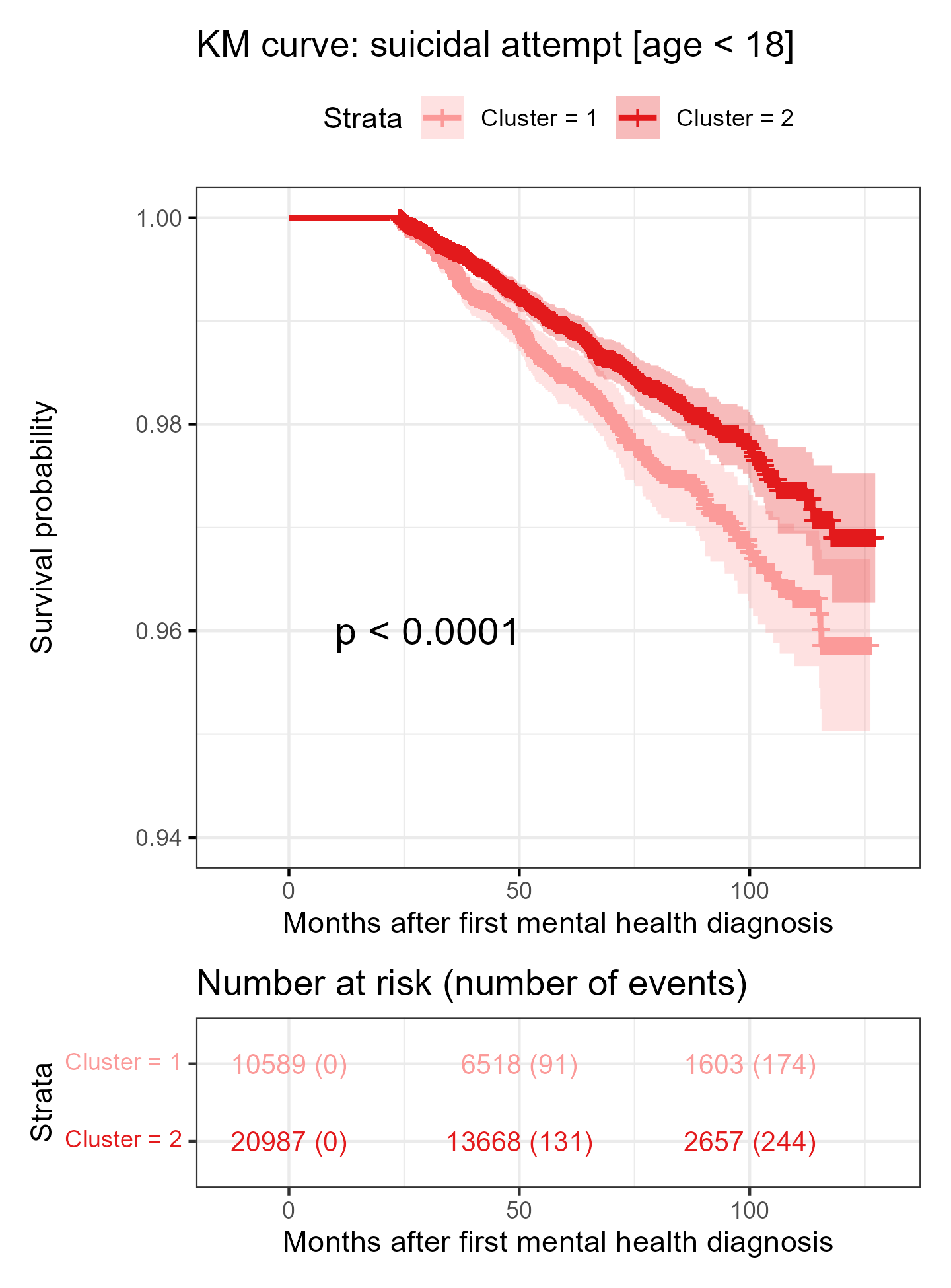}
             \includegraphics[width=3cm]{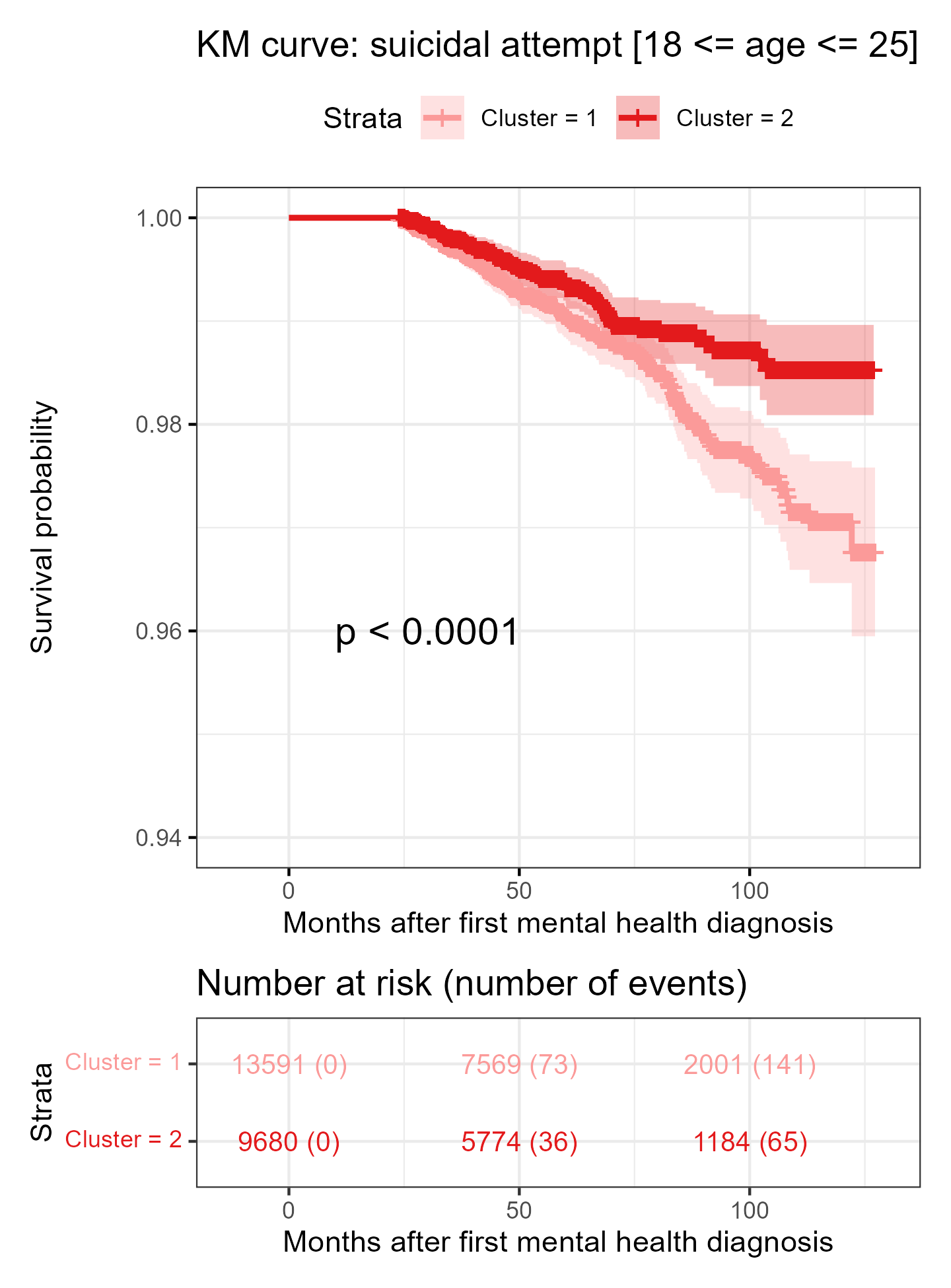}
             \includegraphics[width=3cm]{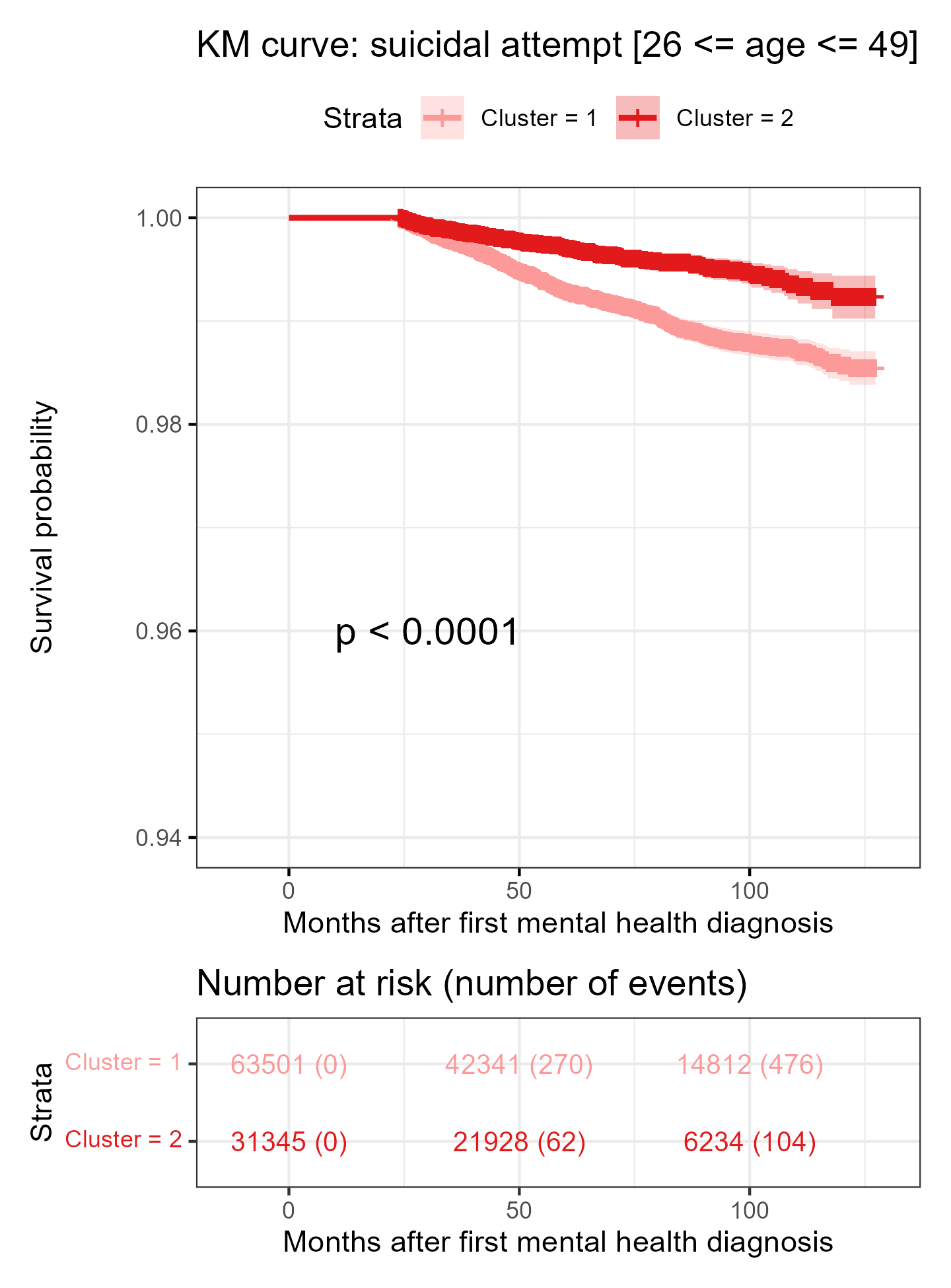}
             \includegraphics[width=3cm]{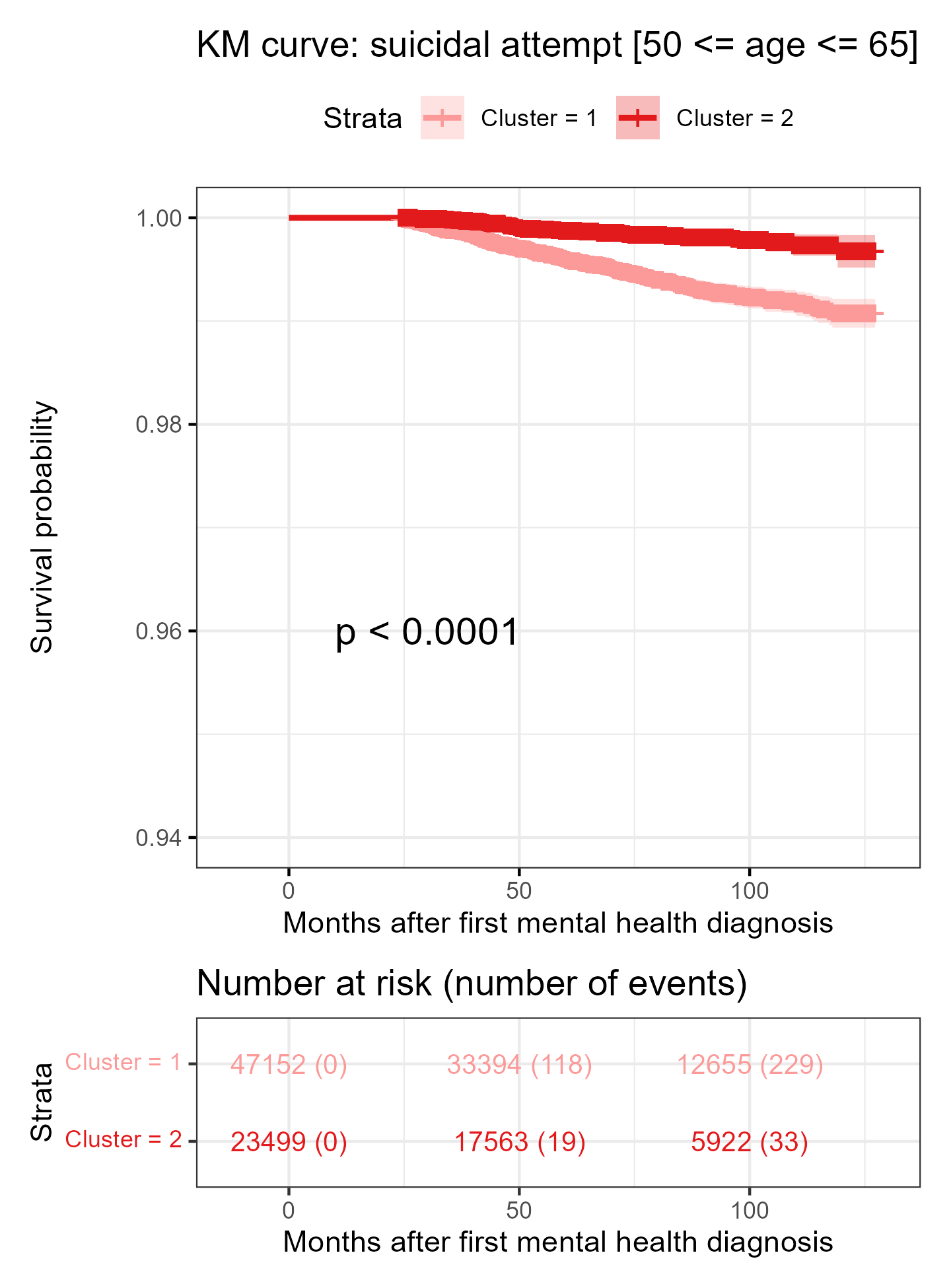}
             \includegraphics[width=3cm]{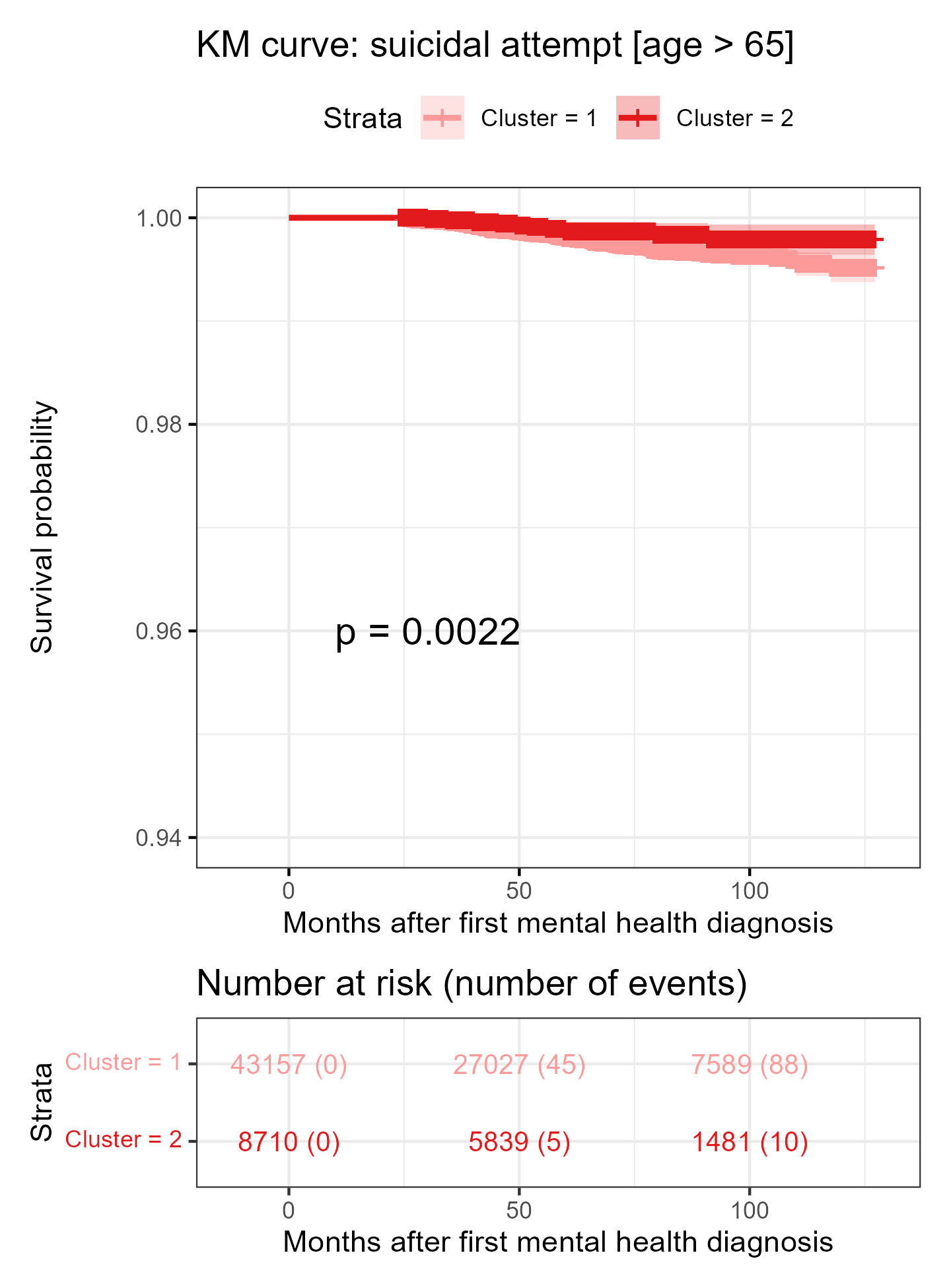}};
        \node[anchor=south west, xshift=3pt, yshift=8pt, fill=white, text opacity=1] 
            at (img.north west) {\textbf{(d)}};
    \end{tikzpicture}

    \caption{Kaplan–Meier curves of time to suicidal ideation or suicide attempt since the first mental health–related EHR code, stratified by age group. 
    Each panel corresponds to a different combination of institution and outcome: 
(a) MGB suicidal ideation, 
(b) MGB suicidal attempt, 
(c) Duke suicidal ideation, and 
(d) Duke suicidal attempt. 
    Blue lines represent MGB; red lines represent Duke.}
    \label{fig:KM_rpdr}
\end{figure}

\begin{figure}[H]
    \centering
    \includegraphics[width=.8\textwidth]{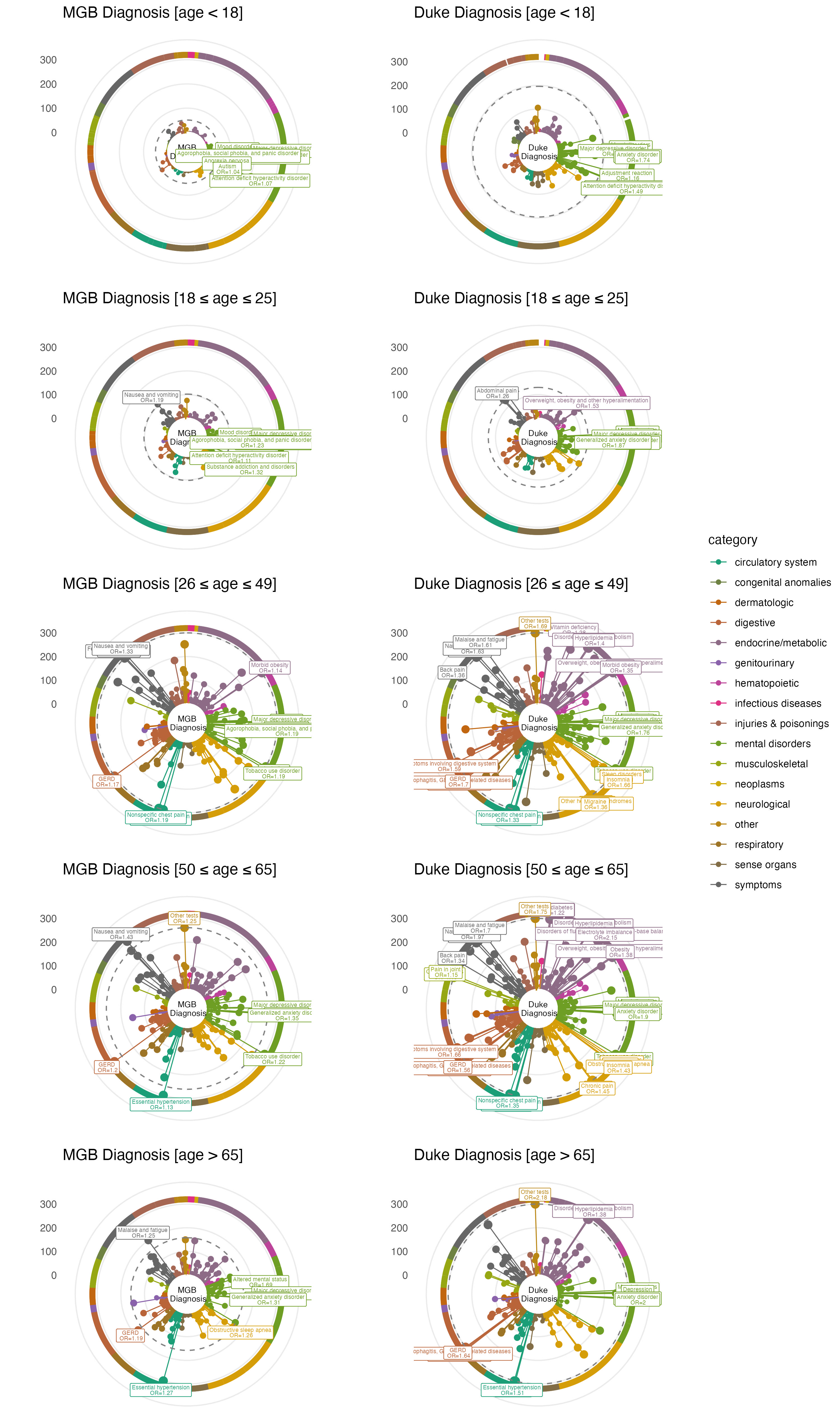}
    \caption{The association between suicide-related diagnosis codes and cluster membership (cluster 1 vs clsuter 2) in each institution (MGB and Duke). Each dot/line represent a unqiue code, and the length shows $-\log_{10}(\text{p-value})$ of the association. Diagnosis codes are grouped into disease categories. Top $12$ diagnosis features are labeled with the code description and odds ratio.  Source data are provided as a Source Data file.}
    \label{fig:suicide_phewas_phe}
\end{figure}

\begin{figure}[H]
    \centering
    \includegraphics[width=.8\textwidth]{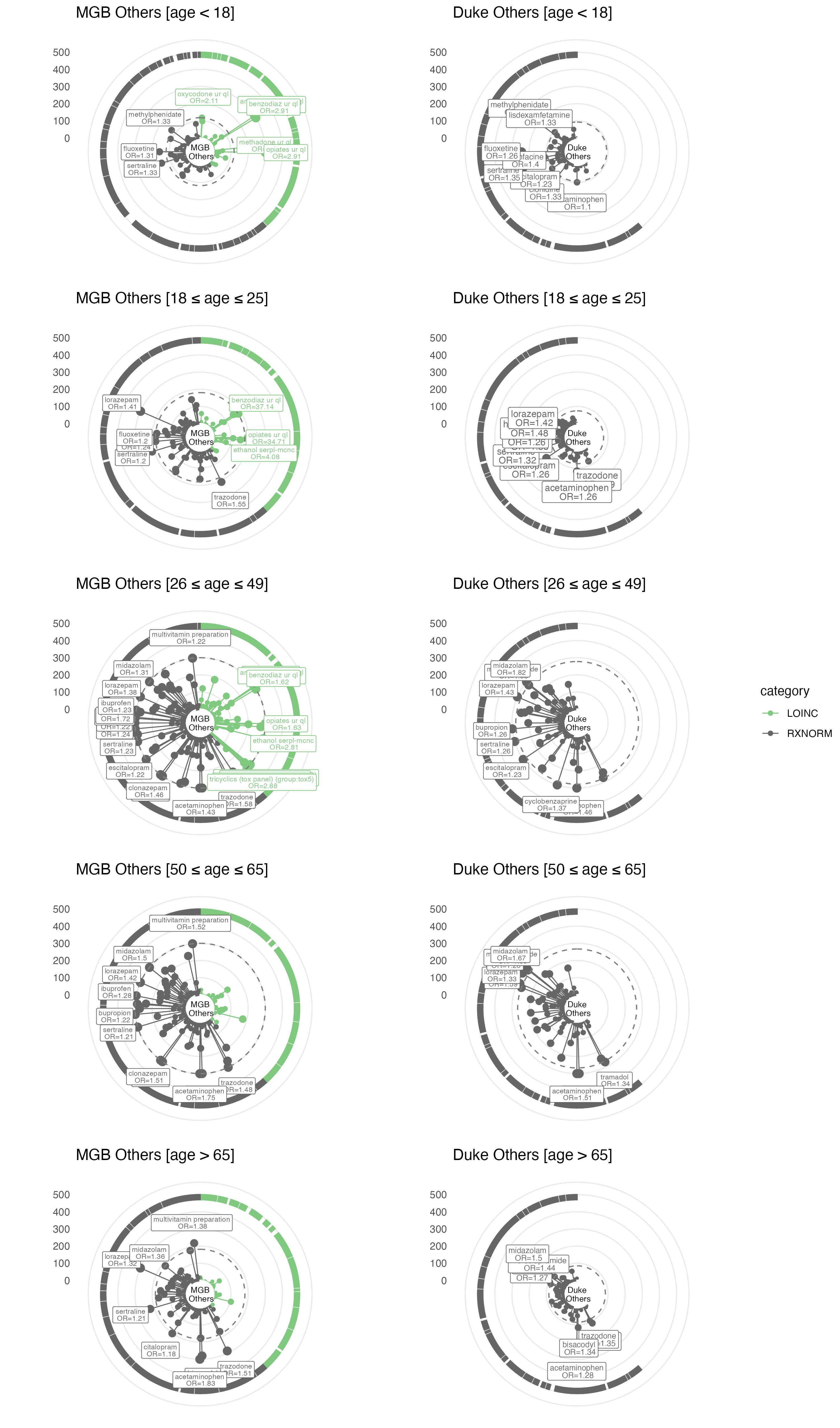}
    \caption{the association between other EHR codes (non-diagnosis codes) and cluster membership (cluster 1 vs clsuter 2) in each institution (MGB and Duke). Each dot/line represent a unqiue code, and the length shows $-\log_{10}(\text{p-value})$ of the association. The codes are grouped into categories. Top 8 features are labeled with the code description and odds ratio.  Source data are provided as a Source Data file.}
    \label{fig:suicide_phewas_oth}
\end{figure}

The prediction performance of suicide attempt models using patient embeddings is shown in Supplementary Fig.~\ref{fig:suicide_AUC}. Compared to the existing clinical model, models trained with GAME-derived embeddings achieved comparable performance at MGB and outperformed it when evaluated at Duke. Additionally, the GAME-based models also outperformed those trained on raw EHR features. The stratified analysis of model performance across different age groups as well as across institutions is shown in Supplementary Table \ref{fig:suicide_AUC_table}. 

\begin{figure}[H]
    \centering\includegraphics[width=\textwidth]{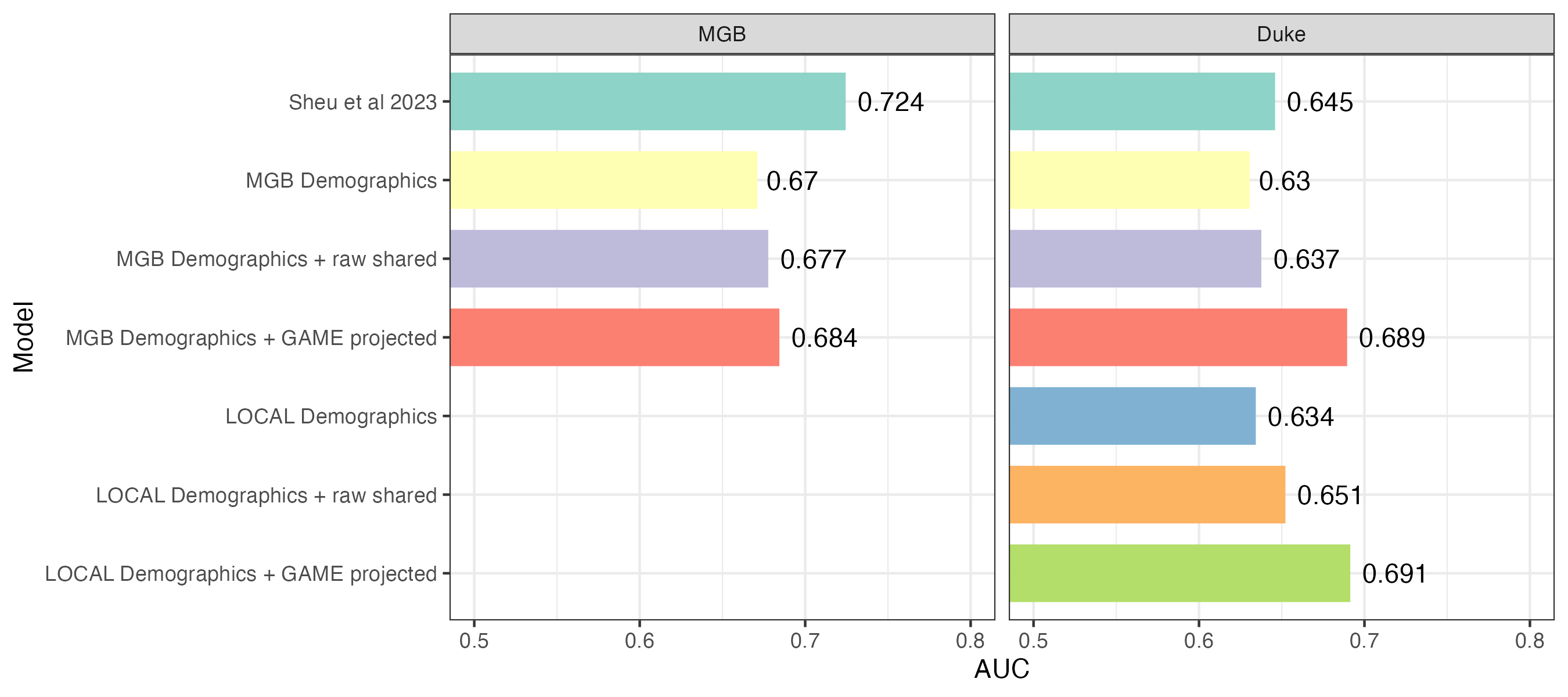}
    \caption{AUC for prediction of 1-year risk of suicidal attempt for mental health patients.  Source data are provided as a Source Data file.}
    \label{fig:suicide_AUC}
\end{figure}

\begin{table}[h]
\centering
\begin{adjustbox}{max width=\textwidth}
\begin{tabularx}{2.0\textwidth}{@{}l*{12}{>{\centering\arraybackslash}X}@{}}
\toprule
\multirow{3}{*}{Model} & \multicolumn{6}{c}{MGB} & \multicolumn{6}{c}{Validation} \\
\cmidrule(lr){2-7} \cmidrule(lr){8-13}
 & \multicolumn{5}{c}{Age Group} & \multirow{2}{*}{Overall} & \multicolumn{5}{c}{Age Group} & \multirow{2}{*}{Overall} \\
\cmidrule(lr){2-6} \cmidrule(lr){8-12}
 & Under 18 & 18-25 & 26-49 & 50-65 & Over 65 & & Under 18 & 18-25 & 26-49 & 50-65 & Over 65 & \\
\midrule
Jiehuan\_ICD & 0.735 & 0.719 & 0.728 & 0.732 & 0.687 & 0.724 & 0.660 & 0.628 & 0.654 & 0.639 & 0.632 & 0.645 \\
MGB Demographics & 0.702 & 0.677 & 0.687 & 0.646 & 0.610 & 0.670 & 0.667 & 0.635 & 0.638 & 0.602 & 0.619 & 0.630 \\
MGB Demographics + raw shared & 0.636 & 0.668 & 0.713 & 0.624 & 0.614 & 0.677 & 0.677 & 0.623 & 0.655 & 0.610 & 0.601 & 0.637 \\
MGB Demographics + raw selected & 0.741 & 0.721 & 0.750 & 0.721 & 0.721 & 0.737 & 0.713 & 0.684 & 0.685 & 0.623 & 0.628 & 0.670 \\
MGB Demographics + GAME projected & 0.710 & 0.698 & 0.691 & 0.684 & 0.626 & 0.684 & 0.723 & 0.682 & 0.705 & 0.665 & 0.648 & 0.689 \\
\midrule
LOCAL Demographics & & & & & & & 0.669 & 0.630 & 0.651 & 0.591 & 0.619 & 0.634 \\
LOCAL Demographics + raw shared & & & & & & & 0.670 & 0.645 & 0.689 & 0.605 & 0.557 & 0.651 \\
LOCAL Demographics + raw selected & & & & & & & 0.731 & 0.670 & 0.721 & 0.671 & 0.612 & 0.693 \\
LOCAL Demographics + GAME projected & & & & & & & 0.721 & 0.698 & 0.706 & 0.667 & 0.635 & 0.691 \\
LOCAL Demographics + GAME projected + prior attempt & & & & & & & 0.669 & 0.643 & 0.677 & 0.669 & 0.620 & 0.664 \\
\bottomrule
\end{tabularx}
\end{adjustbox}
\caption{AUC for prediction of 1-year risk of suicidal attempt for mental health patients}
\label{fig:suicide_AUC_table}
\end{table}

{
\subsection{Evaluating PLM fine-tuning as an alternative backbone to GAME}
\label{sec:PLM}
To further assess the effectiveness of the \textsc{GAME} framework, we compare it against fine-tuning pretrained language models (PLMs) using the same relational supervision. Specifically, we fine-tune PLMs on (i) relatedness and similarity pairs, and (ii) biomedical concept pairs derived by thresholding the PPMI matrices.

We adopt the BAAI General Embedding (BGE) as a representative PLM, given its state-of-the-art performance on the MTEB leaderboard. The model is fine-tuned on (a) the constructed knowledge graph (Ablation~(b)), which integrates hierarchical and UMLS-derived edges, and (b) the 99th-percentile PPMI edges from each institution (Ablation~(c)), without GPT confirmation. We denote the pretrained model as BGE, and the fine-tuned variants as BGE\_KG (knowledge-graph edges), BGE\_PMI (PPMI edges), and BGE\_ALL (combined supervision). All models are evaluated on the same validation and test sets used for \textsc{GAME}, and results are summarized in Supplementary Table~\ref{tab:plm_fine}.

For all fine-tuning experiments, we use the Adam optimizer with a learning rate of $5\times10^{-5}$ and a batch size of $512$. Models are trained for $150$ epochs using a contrastive loss with a margin of $0.1$; negative samples are randomly generated within each batch. Experiments are conducted on an internal GPU cluster (A100-equivalent), and each run completes in $10-24$~hours depending on dataset size.

\begin{table}[h]
\centering
\begin{tabular}{l|c|c|c|c|c|c|c}
\toprule
Model & Relatedness & Similarity & Feature selection & TOP1 & TOP5 & TOP10 & TOP20\\
\midrule
GAME &\textbf{0.914}&0.916&\textbf{0.863}&\textbf{74.2\%}&\textbf{88.0\%}&\textbf{90.7\%}&\textbf{92.7\%}\\
BGE&0.843&\textbf{0.925}&0.724&61.9\%&79.0\%&84.3\%&89.7\%\\
BGE\_KG&0.787&0.739&0.734&7.7\%&12.6\%&18.2\%&29.9\%\\
BGE\_PMI&0.640&0.834&0.730&1.4\%&4.1\%&7.0\%&11.8\%\\
BGE\_ALL&0.680&0.834&0.747&1.4\%&4.8\%&7.2\%&12.9\%\\
\bottomrule
\end{tabular}
\begin{flushleft}
\footnotesize
\textbf{Bold} indicates the highest AUC/C-index/accuracy within each column.
\end{flushleft}
\caption{Comparison of GAME and fine-tuning PLM with knowledge graph edges and PMI pairs. Relatedness and similarity are evaluated on the same test set as in the main experiments; the feature selection and code mapping tasks follow \emph{Methods} section.}

\label{tab:plm_fine}
\end{table}

As shown in Supplementary Table~\ref{tab:plm_fine}, fine-tuning BGE improves performance marginally relative to the pretrained baseline on feature selection, but substantially degrades similarity AUC, relatedness AUC, and code-mapping accuracy. Overall, all fine-tuned PLMs underperform \textsc{GAME} across evaluation tasks. These results indicate that while PLMs can incorporate relational information through contrastive objectives, they struggle to capture the structured dependencies that \textsc{GAME} leverages through its graph-based harmonization and loss design.
}

{
\subsection{Extending GAME to NLP-derived Concept Unique Identifiers (CUIs)}
\label{sec:NLP}

To assess whether \textsc{GAME} generalizes beyond structured EHR codes, we extend the framework to incorporate information extracted from free-text clinical notes. All unstructured notes were processed using the Narrative Information Linear Extraction (NILE) system~\citep{yu2013nile}, which maps clinical expressions to Concept Unique Identifiers (CUIs) defined in the Unified Medical Language System (UMLS). This mapping aligns text-derived entities with established biomedical ontologies and thus expands the relational space beyond institution-specific coding systems.
Given the large universe of CUIs, we focus on the curated subset of $17,779$ CUIs used in~\citep{Xiong2023.09.29.23296239}, which are enriched for clinically meaningful concepts and have documented associations with existing disease categories.

The resulting NLP-derived CUIs were incorporated alongside codified data from each institution. In particular, the MIMIC and VA datasets were augmented with CUIs covering standard semantic categories such as disorders, chemicals, procedures, and physiological processes (Supplementary Table~\ref{stat-CUI}). We then applied the same \textsc{GAME} architecture to these CUI-augmented corpora. This allowed us to test whether the model’s design for code harmonization, relational learning, and privacy-preserving embedding construction extends naturally to text-derived knowledge.

We evaluated three downstream tasks—(i) similarity and relatedness detection, (ii) cross-institutional code mapping, and (iii) feature selection—using the identical experimental procedures as in the main analysis. The inclusion of CUIs provides a stringent test of \textsc{GAME}’s robustness under modality expansion, as CUIs capture linguistic and contextual nuances that are not present in codified EHR data.

\begin{table}[ht!]
\footnotesize
\centering
\resizebox{\textwidth}{!}{
\begin{tabular}{lcc|ccccc|cccccccc} 
\toprule
\multirow{2}{*}{{Institution}} & \multirow{2}{*}{{Location}} & \multirow{2}{*}{{Patients}} & \multicolumn{5}{c}{{Structured EHR codes}} & \multicolumn{7}{c}{{UMLS Concept Unique Identifiers (CUIs)}} &\\
 & & & {PheCode} & {CCS} & {LOINC} & {RxNorm} & {Non-Standard Code} & {ACTI} & {ANAT} & {CHEM} & {DISO} & {PHEN} & {PHYS} & {PROC} & {Total} \\ 
\midrule
BCH   & {\scriptsize NE} & $251$K & 1405 & 199 & 3024 & 1117 & 0     & 0 & 0 & 0 & 0 & 0 & 0 & 0 & 5745 \\ 
BDX   & {\scriptsize France}       & $2.5$M & 1656 & 0   & 2935 & 1103 & 20938 & 0 & 0 & 0 & 0 & 0 & 0 & 0 & 26632 \\ 
Duke  & {\scriptsize SE} & $6.0$M & 1439 & 0   & 554  & 278  & 854   & 0 & 0 & 0 & 0 & 0 & 0 & 0 & 3125 \\ 
MGB   & {\scriptsize NE} & $2.5$M & 1772 & 243 & 3719 & 1235 & 0     & 0 & 0 & 0 & 0 & 0 & 0 & 0 & 6969 \\ 
MIMIC & {\scriptsize NE} & $265$K & 1480 & 203 & 0    & 1141 & 3686  & 3 & 102 & 1238 & 7003 & 155 & 5 & 1567 & 16583 \\ 
UPMC  & {\scriptsize NE} & $95$K  & 1841 & 245 & 5127 & 1987 & 8071  & 0 & 0 & 0 & 0 & 0 & 0 & 0 & 17271 \\  
VA    & {\scriptsize US}           & $12.6$M& 1776 & 224 & 691  & 1561 & 2461  & 114 & 0 & 2590 & 9694 & 274 & 232 & 2740 & 22357 \\ 
\midrule
{Total} & - & {24.2M} & {1869} & {248} & {9410} & {3792} & {36010} & {114} & {102} & {2774} & {11258} & {311} & {235} & {2985} & {69108} \\ 
\bottomrule
\end{tabular}}
\caption{
Summary of structured EHR codes and NLP-derived CUIs across the seven participating institutions. 
Abbreviations of regions: NE = Northeast; SE = Southeast. 
Institutional abbreviations: 
BCH = Boston Children's Hospital; 
BDX = Bordeaux University Hospital; 
Duke = Duke Clinical Research Datamart; 
MGB = Mass General Brigham; 
MIMIC = Medical Information Mart for Intensive Care IV; 
UPMC = University of Pittsburgh Medical Center; 
VA = U.S. Veterans Health Administration.
\\[4pt]
\textit{Non-standard codes} refer to locally defined or non-ontology identifiers used within each institution:  
BDX (CCAM, DXC\_RESULTAT\_NUM, GLI, PMSI, PRESC, SYN\_ANA);  
Duke (Local Lab);  
MIMIC (ICD-10, ICD-9, Local Codes);  
UPMC (ICD-10-CM, ICD-9-CM, Local Lab, Local PX, NDC);  
VA (Other Lab, ShortName).
\\[4pt]
\textit{CUI semantic types}:  
ACTI = Activities \& Behaviors;  
ANAT = Anatomy;  
CHEM = Chemicals \& Drugs;  
DISO = Disorders;  
PHEN = Phenomena;  
PHYS = Physiology;  
PROC = Procedures.
}
\label{stat-CUI}
\end{table}

\subsubsection{Evaluation of similarity and relatedness detection}

\begin{table}[H]
\centering
\resizebox{1\textwidth}{!}{%
\begin{tabular}{cc|cccc|cccc}
\toprule
\multicolumn{2}{c|}{{Method}} 
 & {Similarity} & {Similarity CI} 
 & {Relatedness} & {Relatedness CI}
 & {Sim (CUI)} & {Sim (CUI) CI}
 & {Rel (CUI)} & {Rel (CUI) CI} \\ 
\midrule

\multirow{7}{*}{PPMI-SVD} 
 & BCH   & 0.838 & [0.818, 0.846] & 0.753 & [0.735, 0.768] &  &  &  &  \\ 
 & BDX   & 0.873 & [0.863, 0.883] & 0.825 & [0.816, 0.848] &  &  &  &  \\ 
 & Duke  & 0.832 & [0.812, 0.843] & 0.795 & [0.760, 0.808] &  &  &  &  \\
 & MGB   & 0.942 & [0.938, 0.955] & 0.831 & [0.818, 0.842] &  &  &  &  \\ 
 & MIMIC & 0.744 & [0.739, 0.773] & 0.768 & [0.741, 0.773] & 0.812 & [0.798, 0.829] & 0.757 & [0.752, 0.784]\\ 
 & UPMC  & 0.875 & [0.865, 0.883] & 0.765 & [0.744, 0.773] &  &  &  &  \\ 
 & VA    & 0.930 & [0.926, 0.941] & 0.825 & [0.818, 0.841] & 0.956 & [0.944,0.958] & 0.882 & [0.880,0.900] \\
\multicolumn{2}{c|}{AVE} & 0.862 & [0.858, 0.868] & 0.795 & [0.784, 0.796] & 0.884   & [0.872,0.891] & 0.820  & [0.819,0.839] \\ \hline

\multicolumn{2}{c|}{BBERT} & 0.682 & [0.671, 0.689] & 0.627 & [0.613, 0.637] & 0.633 & [0.615, 0.645] & 0.587 & [0.576, 0.602] \\ 
\multicolumn{2}{c|}{PBERT} & 0.644 & [0.634, 0.656] & 0.607 & [0.596, 0.620] & 0.616 & [0.604, 0.637] & 0.590 & [0.576, 0.605] \\ 
\multicolumn{2}{c|}{SBERT} & 0.829 & [0.820, 0.835] & 0.733 & [0.725,0.751] & 0.897 & [0.891, 0.907] & 0.847 & [0.838, 0.856] \\ 
\multicolumn{2}{c|}{CODER} & 0.888 & [0.882, 0.894] & 0.644 & [0.636, 0.665] & 0.969 & [0.964, 0.971] & 0.831 & [0.828, 0.839] \\  
\multicolumn{2}{c|}{BGE} & 0.935 & [0.927, 0.936] & 0.830 & [0.819, 0.835] & 0.969 & [0.963, 0.970] & 0.934 & [0.929, 0.940] \\ 
\multicolumn{2}{c|}{OpenAI} & \textbf{0.951} & [0.944, 0.951] & 0.808 & [0.799, 0.817] & \textbf{0.983} & [0.980, 0.986] & 0.902 & [0.892, 0.903] \\ 
\multicolumn{2}{c|}{GAME} & 0.936 & [0.932, 0.941] & \textbf{0.894} & [0.893, 0.907] &  0.980 & [0.978,0.983]  & \textbf{0.956} & [0.951,0.958] \\ 
\bottomrule
\end{tabular}
}
\begin{flushleft}
\footnotesize
\textbf{Bold} indicates the highest AUC within each column.
\end{flushleft}
\caption{AUC for detecting concept similarity and relatedness, including CUI-level extensions. Values in brackets indicate $95\%$ confidence intervals (CIs), estimated via bootstrapping; the CI for the average (AVE) is based on a normal approximation.  BBERT = BioBERT; PBERT = PubMedBERT; SBERT = SAPBERT. Boldface indicates the best-performing method within each column. For simplicity, the notations below remain the same and are omitted.}
\label{R2-CUI}
\end{table}

As shown in Supplementary Table~\ref{R2-CUI}, the performance trends are consistent with those observed in Supplementary Table~\ref{R2}. 
\textsc{GAME} achieves the highest AUC in relatedness detection and performs comparably to OpenAI on similarity detection. 
After incorporating NLP-derived CUIs, \textsc{GAME} maintains or slightly improves performance relative to the codified-only setting, demonstrating that the model generalizes robustly across both structured and text-derived features.

\subsubsection{Cross-institutional code mapping}

\begin{table}[H]
\centering
\setlength{\tabcolsep}{4pt}
\begin{tabular}{c|lllllll}
\toprule
{Measure (\%)} & PBERT & BBERT & SBERT & CODER & OpenAI & BGE & GAME \\ \hline
TOP1
& $11.2^{(0.7)}_{***}$ 
& $20.8^{(0.9)}_{***}$ 
& $59.6^{(1.1)}_{***}$ 
& $55.6^{(1.1)}_{***}$ 
& $59.8^{(1.1)}_{***}$ 
& $61.9^{(1.1)}_{***}$ 
& $\mathbf{74.4^{(1.0)}}$ \\
TOP5
& $18.3^{(0.9)}_{***}$ 
& $31.4^{(1.0)}_{***}$ 
& $79.0^{(0.9)}_{***}$ 
& $75.9^{(0.9)}_{***}$ 
& $85.0^{(0.8)}_{**}$ 
& $79.0^{(0.9)}_{***}$ 
& $\mathbf{87.4^{(0.7)}}$ \\
TOP10
& $21.9^{(0.9)}_{***}$ 
& $36.5^{(1.1)}_{***}$ 
& $82.9^{(0.8)}_{***}$ 
& $85.0^{(0.8)}_{***}$ 
& $89.0^{(0.7)}_{*}$ 
& $84.3^{(0.8)}_{***}$ 
& $\mathbf{90.4^{(0.7)}}$ \\
TOP20
& $27.0^{(1.0)}_{***}$ 
& $42.4^{(1.1)}_{***}$ 
& $85.7^{(0.8)}_{***}$ 
& $89.7^{(0.7)}_{***}$ 
& $92.0^{(0.6)}$ 
& $89.7^{(0.7)}_{***}$ 
& $\mathbf{92.4^{(0.6)}}$ \\
\bottomrule
\end{tabular}
\begin{flushleft}
\footnotesize
\textbf{Bold} indicates the best performance in each row.
Values are reported as mean$^{(\mathrm{standard\ deviation})}$, with standard deviation measured in percentage points. Statistical significance relative to GAME was assessed using a two-sided McNemar's test ($\mathrm{df}=1$) and is indicated by subscripts:
* $p<0.05$, ** $p<0.005$, *** $p<0.0005$.
\end{flushleft}
\caption{Accuracy (\%) of mapping VA local lab codes to LOINC/LP codes using different methods.}
\label{R1-CUI}
\end{table}

\begin{table}[H]
\centering
\begin{tabular}{cc|c|c|c}
\toprule
\multicolumn{2}{c|}{} & {UPMC PX-CCS} & {BDX CCAM-CCS} & {UPMC LAB-LOINC} \\ \midrule
\multicolumn{2}{c|}{ BBERT}        & 0.059 & 0.138 & 0.102  \\ 
\multicolumn{2}{c|}{ PBERT}         & 0.001 & 0.317 & 0.112  \\ 
\multicolumn{2}{c|}{ SBERT}         & 0.313 & 0.423 & 0.529  \\ 
\multicolumn{2}{c|}{ CODER}         & 0.418 & 0.540 & 0.554  \\ 
\multicolumn{2}{c|}{ BGE}          & 0.409 & 0.615 & \textbf{0.610}  \\ 
\multicolumn{2}{c|}{ OpenAI}       & 0.484 & 0.616 & 0.583  \\ 
\multicolumn{2}{c|}{{GAME}} & \textbf{0.596} & \textbf{0.675} & {0.602} \\ 
\midrule
\multicolumn{2}{c|}{{\# Pairs}} & 199 & 537 & 1809 \\
\multicolumn{2}{c|}{{\#  Standard Code}} & 10 & 29 & 84 \\ \bottomrule
\end{tabular}
\begin{flushleft}
\footnotesize
\textbf{Bold} indicates the highest correlation within each column.
\end{flushleft}
\caption{
Spearman's rank correlation between cosine similarity scores and human annotations. A higher rank correlation indicates that cosine similarity better reflects the relationship between local codes and standard codes. \# Pairs denotes the number of evaluated pairs; \# Standard Codes is the number of unique standard codes. \# Pairs denotes the number of evaluated pairs, while \#  Standard Code denotes the total number of unique standard codes.
}
\label{R2_code_map_2-CUI}
\end{table}

As shown in Supplementary Tables~\ref{R1-CUI} and~\ref{R2_code_map_2-CUI}, the results closely mirror those in Supplementary Tables~\ref{R1} and~\ref{R2_code_map_2}. \textsc{GAME} consistently achieves the highest accuracy in code mapping, indicating that incorporating CUI-derived information does not compromise performance on other tasks. This robustness underscores \textsc{GAME}’s ability to support multi-task learning across both structured and text-enriched data representations.

\subsubsection{Feature selection}

Supplementary Tables~\ref{tab:fea_sel_human-CUI}, \ref{R3_CUI}, and \ref{R3_-CUI} summarize the evaluation of \textsc{GAME}’s feature selection performance after incorporating CUI data.
Using the same expert annotations as in Supplementary Table~\ref{tab:fea_sel_human}, Supplementary Table~\ref{tab:fea_sel_human-CUI} shows that CUI integration does not degrade \textsc{GAME}’s alignment with human-labeled feature relevance.
Supplementary Table~\ref{R3_CUI} extends this analysis to the six primary PheCodes in Supplementary Table~\ref{R3_}, where \textsc{GAME} again achieves consistently superior concordance indices. Finally, Supplementary Table~\ref{R3_-CUI} evaluates the primary CUI associated with each of the six diseases, showing that \textsc{GAME} continues to achieve strong concordance with GPT-4 relevance assessments. Together, these results confirm that \textsc{GAME} maintains stable and effective feature selection across both structured and NLP-derived data sources.

\begin{table}[H]
\centering
\setlength{\tabcolsep}{2pt}
\begin{tabular}{cc|cccccc|c}
\toprule
\multicolumn{2}{c|}{{Method}} & {T1D} & {Epilepsy} & {PH} & {Asthma} & {CD} & {UC} & {AVE} \\
\midrule
\multirow{7}{*}{{PPMI-SVD}} 

& {BCH}   & 0.922 & 0.897 & --    & 0.830 & 0.903 & 0.852 & 0.881 \\
& {BDX}   & 0.799 & 0.938 & 0.707 & 0.793 & 0.885 & 0.780 & 0.817 \\
& {Duke}  & 0.619 & 0.860 & --    & 0.725 & 0.826 & 0.885 & 0.783 \\
& {MGB}   & 0.697 & 0.882 & 0.781 & 0.768 & 0.891 & 0.816 & 0.806 \\
& {MIMIC} & 0.577 & 0.763 & 0.512 & 0.632 & 0.776 & 0.641 & 0.650 \\
& {UPMC}  & 0.866 & 0.885 & 0.713 & 0.729 & 0.849 & 0.831 & 0.812 \\
& {VA}    & 0.832 & 0.873 & 0.789 & 0.808 & 0.908 & 0.838 & 0.841 \\
\multicolumn{2}{c|}{{AVE}} & 0.759 & 0.871 & 0.700 & 0.755 & 0.863 & 0.806 & 0.792 \\
\midrule
\multicolumn{2}{c|}{{BBERT}} & 0.431 & 0.607 & 0.497 & 0.639 & 0.665 & 0.486 & 0.554 \\
\multicolumn{2}{c|}{{PBERT}} & 0.524 & 0.526 & 0.417 & 0.492 & 0.477 & \textbf{0.839} & 0.546 \\
\multicolumn{2}{c|}{{SBERT}} & 0.758 & 0.804 & 0.454 & 0.662 & 0.682 & 0.764 & 0.687 \\
\multicolumn{2}{c|}{{CODER}} & 0.803 & 0.808 & 0.595 & 0.676 & 0.694 & 0.592 & 0.695 \\
\multicolumn{2}{c|}{{BGE}}   & 0.721 & 0.728 & 0.608 & 0.643 & 0.840 & 0.801 & 0.724 \\
\multicolumn{2}{c|}{{OpenAI}} & 0.856 & 0.861 & 0.628 & 0.831 & 0.586 & 0.771 & 0.755 \\
\multicolumn{2}{c|}{{GAME}}  & \textbf{0.873} & \textbf{0.932} & \textbf{0.724} & \textbf{0.875} & \textbf{0.911} & {0.773} & \textbf{0.848} \\
\midrule
\multicolumn{2}{c|}{{GPT-4}} &  0.885 & 0.937 & 0.828 & 0.845 & 0.962 & 0.964 &  0.903\\ 
\multicolumn{2}{c|}{{\# Pairs}} & 60 & 305 & 160 & 180 & 40 & 40 \\ 
\bottomrule
\end{tabular}
\begin{flushleft}
\footnotesize
\textbf{Bold} indicates the highest C-index among all methods for each disease.
\end{flushleft}
\caption{C-index values between cosine similarity scores and expert-labeled feature relevance for six diseases across different embedding methods. Higher values indicate better alignment with expert annotations. The ``GPT-4'' row reports agreement between GPT-4-assigned scores and expert annotations, measured by the C-index. \# Pairs denotes the number of expert-labeled features evaluated for each disease. Results for PH (PheCode 415.21) are unavailable for BCH, Duke because this disease is not present in those institutions, consistent with Supplementary Tables~\ref{R3_CUI} and \ref{R3_-CUI}.}
\label{tab:fea_sel_human-CUI}
\end{table}

\begin{table}[H]
\centering
\setlength{\tabcolsep}{2pt}
\begin{tabular}{cc|cccccc|c}
\toprule
\multicolumn{2}{c|}{{Method}} & {T1D} & {Epilepsy} & {PH} & {Asthma} & {CD} & {UC} & {AVE} \\
\midrule
\multirow{7}{*}{{PPMI-SVD}} 
& {BCH}   & 0.532 & 0.626 & -- & 0.602 & 0.608 & 0.647 & 0.579 \\
& {Bor}   & 0.672 & 0.750 & 0.667 & 0.629 & 0.612 & 0.665 & 0.666 \\
& {Duke}  & 0.470 & 0.536 & -- & 0.453 & 0.528 & 0.558 & 0.500 \\
& {MGB}   & 0.708 & 0.716 & 0.750 & 0.721 & 0.729 & 0.758 & 0.730 \\
& {MIMIC} & 0.598 & 0.621 & 0.543 & 0.408 & 0.569 & 0.557 & 0.549 \\
& {UPMC}  & 0.642 & 0.614 & 0.639 & 0.610 & 0.536 & 0.573 & 0.602 \\
& {VA}    & 0.758 & 0.752 & 0.760 & 0.721 & 0.752 & 0.744 & 0.748 \\
\multicolumn{2}{c|}{{AVE}} & 0.626 & 0.659 & 0.610 & 0.592 & 0.619 & 0.643 & 0.625 \\
\midrule
\multicolumn{2}{c|}{{BBERT}} & 0.421 & 0.429 & 0.503 & 0.479 & 0.461 & 0.486 & 0.463 \\
\multicolumn{2}{c|}{{PBERT}} & 0.441 & 0.431 & 0.507 & 0.466 & 0.477 & 0.472 & 0.466 \\
\multicolumn{2}{c|}{{SBERT}} & 0.602 & 0.517 & 0.619 & 0.608 & 0.592 & 0.578 & 0.586 \\
\multicolumn{2}{c|}{{CODER}} & 0.650 & 0.597 & 0.653 & 0.632 & 0.579 & 0.597 & 0.618 \\
\multicolumn{2}{c|}{{BGE}}   & 0.645 & 0.591 & 0.658 & 0.612 & 0.571 & 0.581 & 0.610 \\
\multicolumn{2}{c|}{{OpenAI}} & 0.684 & 0.695 & \textbf{0.676} & 0.690 & 0.583 & 0.656 & 0.664 \\
\multicolumn{2}{c|}{{GAME}}  & \textbf{0.767} & \textbf{0.803} & 0.600 & \textbf{0.775} & \textbf{0.709} & \textbf{0.694} & \textbf{0.725} \\
\bottomrule
\end{tabular}
\begin{flushleft}
\footnotesize
\textbf{Bold} indicates the highest C-index among all methods for each disease.
\end{flushleft}
\caption{C-index between cosine similarity scores and GPT-4-assigned relevance scores across diseases and methods. }
\label{R3_CUI} 
\end{table}

\begin{table}[H]
\centering
\setlength{\tabcolsep}{2pt}
\begin{tabular}{cc|cccccc|c}
\toprule
\multicolumn{2}{c|}{Method} & T1D & Epilepsy & PH & Asthma & CD & UC & AVE \\
\midrule
\multirow{2}{*}{{PPMI-SVD}} 
& {MIMIC} & 0.609 & 0.585 & 0.592 & 0.512 & 0.608 & 0.601 & 0.584 \\
& {VA}    & 0.752 & 0.758 & 0.774 & 0.670 & 0.740 & 0.753 & 0.741 \\
\multicolumn{2}{c|}{{AVE}} & 0.681 & 0.672 & 0.683 & 0.591 & 0.674 & 0.677 & 0.663 \\
\midrule
\multicolumn{2}{c|}{{BBERT}} & 0.365 & 0.477 & 0.489 & 0.505 & 0.490 & 0.518 & 0.474 \\
\multicolumn{2}{c|}{{PBERT}} & 0.433 & 0.491 & 0.511 & 0.494 & 0.494 & 0.497 & 0.486 \\
\multicolumn{2}{c|}{{SBERT}} & 0.714 & 0.603 & 0.621 & 0.692 & 0.578 & 0.652 & 0.643 \\
\multicolumn{2}{c|}{{CODER}} & 0.674 & 0.644 & 0.652 & 0.680 & 0.624 & 0.654 & 0.655 \\
\multicolumn{2}{c|}{{BGE}}   & 0.710 & 0.691 & 0.666 & 0.669 & 0.577 & 0.650 & 0.661 \\
\multicolumn{2}{c|}{{OpenAI}} & 0.709 & 0.740 & 0.704 & 0.735 & 0.632 & 0.684 & 0.701 \\
\multicolumn{2}{c|}{{GAME}}  & \textbf{0.761} & \textbf{0.809} & \textbf{0.784} & \textbf{0.800} & \textbf{0.735} & \textbf{0.726} & \textbf{0.769} \\
\bottomrule
\end{tabular}
\begin{flushleft}
\footnotesize
\textbf{Bold} indicates the highest C-index among all methods for each disease.
\end{flushleft}
\caption{Concordance index (C-index) comparing cosine similarity scores with GPT-4–assigned relevance scores across diseases and embedding methods. The evaluation is performed on six CUIs (C0011854, C0014544, C0020542, C0004096, C0010346, C0009324) that represent the principal clinical concepts associated with the selected diseases.}
\label{R3_-CUI}
\end{table}
}


\end{document}